\documentclass[11pt]{article}

\usepackage[compress]{natbib}

\usepackage[margin=1.1in]{geometry} 

\input{packages_and_commands.tex}

\begin{document}

\title{Instance-Optimality for Private KL Distribution Estimation}
\author{Jiayuan Ye\footnote{Research done while at Apple}\\National University of Singapore \and Vitaly Feldman\\Apple \and Kunal Talwar\\Apple}
\date{}
\maketitle

\begin{abstract}
    We study the fundamental problem of estimating an unknown discrete distribution $p$ over $d$ symbols, given $n$ i.i.d. samples from the distribution. We are interested in minimizing the KL divergence between the true distribution and the algorithm's estimate. We first construct minimax optimal private estimators. Minimax optimality however fails to shed light on an algorithm's performance on individual (non-worst-case) instances $p$ and simple minimax-optimal DP estimators can have poor empirical performance on real distributions. We then study this problem from an instance-optimality viewpoint, where the algorithm's error on $p$ is compared to the minimum achievable estimation error over a small local neighborhood of $p$. Under natural notions of local neighborhood, we propose algorithms that achieve instance-optimality up to constant factors, with and without a differential privacy constraint. Our upper bounds rely on (private) variants of the Good-Turing estimator. Our lower bounds use additive local neighborhoods that more precisely captures the hardness of distribution estimation in KL divergence, compared to ones considered in prior works.
\end{abstract}

\section{Introduction}

\label{sec:intro}

Accurately estimating a discrete distribution (over $d$ symbols) from the empirical samples is a fundamental task in statistical machine learning. An especially important distribution estimation objective is the Kullback-Leibler (KL) divergence error, as it is crucial in promoting diversity and smoothness by penalizing zero-mass assignment to unseen symbols. Noticeably in speech-recognition and language modeling communities, negative log likelihood on a test set (which up to translation is the KL divergence) is the measure that has been found to best correlate with the performance of the model~\cite{chen1999empirical,chen1998evaluation}, and is therefore the standard loss being optimized for. Additionally, KL error is well-established in the coding community~\cite{cover1972broadcast,krichevsky1981performance} in the context of data compression (i.e., identifying encoding that represents the information with fewer bits). Due to this reason,  KL distribution estimation has been extensively studied~\cite{xie1997minimax,braess2002achieve,braess2004bernstein,paninski2003estimation,paninski2004variational,kamath2015learning,orlitsky2015competitive}. 

However, overly fine-grained release of statistics make it possible to infer the memberships~\cite{homer2008resolving,sankararaman2009genomic,dwork2015robust,shokri2017membership} or even reconstruct the values~\cite{dinur2003revealing,carlini2021extracting} of input data records. To this end, differential privacy~\cite{dwork2006calibrating} offers a powerful mathematical definition to provably control the associated privacy risk.
\begin{definition}[Differential Privacy (DP)~\cite{dwork2006calibrating}] A (randomized) algorithm $\mathcal{A}$ is $(\varepsilon, \delta)$-differentially private ($(\varepsilon, \delta)$-DP) if for all neighboring datasets $x, x'$ that differ in at most one sample, and any measurable output set $T$, 
$\Pr[\mathcal{A}(x)\in T]\leq e^\varepsilon\cdot \Pr[\mathcal{A}(x')\in T] + \delta$.
    \label{def:dp}
\end{definition}
Our work aims to provide an understanding of the KL error of private distribution estimation. Below we present the problem setting and summarize known results and our contributions.

\paragraph{Problem Setting} Let $p\in \Delta(d)\coloneqq \{p\in\mathbb{R}^d: p_1, \cdots, p_d\geq 0\text{ and }\sum_{i=1}^dp_i=1\}$ be an unknown distribution over $d$ symbols. Let $x\in \mathbb{N}^d$ be the histogram representation of a dataset consisting of empirical samples drawn from distribution $p$, where $x_i$ denotes the count of symbol $i$ in the dataset. Following prior works~\cite{acharya2013optimal,orlitsky2015competitive}, we assume that the count of each symbol $i$ is independently drawn from a Poisson distribution with mean $np_i$, i.e., $(x_1, \cdots, x_d)\sim \text{Poi}(np) \coloneqq \text{Poi}(np_1)\times \cdots \times \text{Poi}(np_d)$. This assumption is a convenient choice for analysis as it ensures independent sampling of each symbol, while enjoying the benefit of being equivalent to sampling from multinomial distribution when conditioned on a fixed dataset size $n$. Our goal is to design an algorithm that accurately estimates the unknown distribution $p$ in KL divergence given a sampled dataset $x\sim\text{Poi}(np)$. 

\paragraph{Minimax Optimality Results (and Their Limitations)} We start by looking at the (private) KL distribution estimation problem from the standard minimax optimality objective, where the goal is to design estimators $\mathcal{A}$ that minimizes the KL error on the \textit{worst-case} distribution instance. 
\begin{align}
    \min_{\mathcal{A}}\max_{p\in \Delta(d)}\mathbb{E}\left[\text{KL}(p, \mathcal{A}(x))\right] \text{, }\text{KL}(p\lVert \mathcal{A}(x)) =  \sum_i p_i\log\frac{p_i}{\mathcal{A}(x)_i} 
    \label{eqn:min_max_objective}
\end{align}
where $\text{KL}(p\lVert \mathcal{A}(x)) =  \sum_i p_i\log\frac{p_i}{\mathcal{A}(x)_i}$, and the expectation is over the randomness of estimator $\mathcal{A}$ and over the sampling of $x\sim \text{Poi}(np)$. This minimax objective~\eqref{eqn:min_max_objective} is well-studied for non-DP algorithms, where simple add-constant estimators are proved to be minimax optimal with $O\left(\ln\left(1 + \frac{d}{n}\right)\right)$ rates~\cite{braess2004bernstein,paninski2004variational} (see \cref{app:nondp_minimax} for a complete discussion of the existing results). For the DP setting, to the best of our knowledge, there are no known results for the KL minimax rates. Nevertheless, in \cref{ssec:dp_minmax}, we show that a similarly simple algorithm (that truncates the Laplace perturbations of empirical counts) achieves the minimax optimal $O\left(\ln\left(1 + \frac{d}{\varepsilon n}\right)\right)$ rates. 

However,  such simple estimators achieve poor performance in experiments (\cref{sec:experiment}) on commonly occurring distributions such as power-law distributions. Similar observations, i.e., the poor performance of the minimax-optimal add-constant estimator (compared to the practical Good-Turing~\cite{good1953population} estimator), have long existed in the non-DP setting. This is intuitively because minimax optimality only captures the \textit{worst-case} error over all possible distributions, thus failing to indicate whether an algorithm performs well for each \textit{non-worst-case} distribution $p$.

\paragraph{Instance-optimality} Instance-optimality is a promising framework to address the above limitation of minimax optimality and shed light on the per-instance provable performance of an estimator. In the non-DP setting, many works have studied instance-optimality in different contexts, such as local minimax estimation~\cite{donoho1994ideal,mcmillan2022instance,duchi2024right}, competitive distribution estimation~\cite{acharya2013optimal,orlitsky2015competitive} or instance-by-instance analysis~\cite{valiant2017automatic}. Remarkably, the seminal work of \cite{orlitsky2015competitive} proved that a simple variant of Good-Turing estimators is nearly instance-optimal, in that it estimates every distribution nearly as well as the best
estimator designed with prior knowledge of the distribution up to a permutation. The recent work by Feldman, McMillan, Sivakumar nd Talwar~\cite{feldman2024instance}, further generalize this instance-optimality definition to other natural definitions of prior knowledge in the language of per-instance neighborhood. Formally, we follow Feldman, McMillan, Sivakumar nd Talwar~\cite{feldman2024instance} and say an estimator $\mathcal{A}$ is instance-optimal with respect to a neighborhood map $N$ if:
\begin{align}
 \forall\,&p\in \Delta(d):  \mathbb{E}[\text{KL}(p, \mathcal{A}(x))]\leq O\left(\text{lower}(p, n, N)\right)\nonumber\\
    &\text{where }\text{lower}(p, n, N) \stackrel{def}{=} \min_{\mathcal{A'}}\max_{q\in N(p)} \mathbb{E}\left[\text{KL}(q, \mathcal{A'}(x))\right]  \text{for a neighborhood }N(p)\text{ of }p.
    \label{eqn:instance_optimality_objective}
\end{align}
That is, instance-optimality says that the algorithm $\mathcal{A}$ is competitive with any hypothetical algorithm $\mathcal{A}'$ that has auxiliary knowledge about the neighborhood $N(p)$ of the input distribution $p$. 
We can further constrain the algorithm to be $(\varepsilon, \delta)$-DP in establishing the per-instance lower bound as follows. 
\begin{align}
    \text{lower}_{\varepsilon, \delta}(p, n, N) = \min_{\mathcal{A}\text{ is }(\varepsilon, \delta)\text{-DP}}\max_{q\in N(p)}\mathbb{E}\left[\text{KL}(q,  \mathcal{A}(x))\right]  \label{eqn:per_instance_lower_dp}
\end{align}
This is the lower bound that we will use for establishing instance-optimality of private algorithms. 

\paragraph{Limitations of Prior Instance-Optimality Objectives for KL Error} In instance-optimality results, the smaller the neighborhood, the stronger the result. This is because we are proving that our algorithm is competitive with a hypothetical algorithm that already knows that the input distribution is in the neighborhood. However, if the neighborhood is too small, then no algorithm can be instance-optimal, as the baseline estimator's knowledge of the instance is overly precise. (As an extreme example, when $N(p)=\{p\}$ contains only the target distribution $p$, the per-instance lower bounds \eqref{eqn:instance_optimality_objective} and \eqref{eqn:per_instance_lower_dp} trivially become zero (as achieved by an estimator that always outputs $\mathcal{A}(x) = p$ regardless of the input dataset).) Below we review existing choices of neighborhood $N(p)$ in the literature, and show they are either too large or too small to accurately capture the per-instance estimation hardness in KL.

\begin{itemize}
    \item Permutation neighborhood~\cite{acharya2013optimal,orlitsky2015competitive,valiant2017automatic}: $N_\pi(p) = \{q: \exists\text{ permutation }\pi\text{ s.t. }q_i = p_{\pi\circ i}, \forall i\in[d]\}$, i.e., all distributions obtained by permuting the probability values. On the one hand, prior work~\cite{feldman2024instance} has argued that this neighborhood is too large, as practical estimators often have stronger knowledge about which symbols are more frequent than others (e.g., the word ``and'' is often more frequent than ``differential'', thus assuming the two words are permutable is not reasonable). On the other hand, permutation neighborhood can be too small, in that it provably does not allow for multiplicative instance optimality guarantees, as discussed in \cref{sec:prior_nondp_estimators}. Indeed, permutation neighborhood assumes precise knowledge of the set of true probability values, which can be overly strong for any realistic estimator to satisfy.
    \item Two-point neighborhood~\cite{duchi2024right,mcmillan2022instance,asi2024universally}: $N_2(p) = \{p, q\}$ contains the target instance $p$ and another alternative distribution $q$ that is ``close'' to $p$. This neighborhood reduces the per-instance hardness to binary hypothesis testing of whether the observed samples are from $p$ or $q$. Under appropriate notions of ``close'', this neighborhood allows matching per-instance upper and lower bounds for estimating one-parameter (exponential) families distribution under central DP~\cite{mcmillan2022instance} and local DP~\cite{duchi2024right}, as well as for general one-dimensional statistical estimation problems~\cite{asi2024universally}. However, for high-dimensional problems, this neighborhood is provably too small, as the per-instance lower bound remain constant under growing data dimension, despite the growing hardness for distribution estimation under higher dimension. Indeed testing can be provably easier than learning in simple settings.
    \item Multiplicative neighborhood~\cite{feldman2024instance}: $N_{\times}(p) = \{q: \frac{1}{2}p_i\leq q_i \leq 2p_i, \forall i \in [d]\}$. It allows matching per-instance upper and lower bounds for distribution estimation in Wasserstein distance. However, this neighborhood is too small to capture the KL error: the per-instance lower bound is at most a constant (since an algorithm that is tailored to $N(p)$ can achieve $O(1)$ error), whereas such an error bound is provably unachievable for some distributions.
\end{itemize}
These limitations call for new definitions of local neighborhood to precisely capture the per-instance hardness of distribution estimation in KL divergence, which we study in this paper.

\paragraph{Our Main Contributions: Instance-Optimal Results}
\begin{itemize}
    \item We define instance-optimality objectives that more precisely capture the hardness of private KL distribution estimation. This is by constructing new neighborhoods \eqref{eqn:add_neighborhood_plus} and \eqref{eqn:add_neighborhood_dp_small_symbols} that additively perturb the probability values $p_i$ of individual symbols. We use small perturbation scales -- $1/n$, $1/n\epsilon$, and $\sqrt{p_i/n}$ -- to ensure the dataset (sampled by $p$) could plausibly have come from other distribution in the neighborhood. (See \cref{sec:def_instance_optimal_nondp} and \ref{sec:def_instance_optimal_dp} for more rationales on the neighborhood sizes.) This is stronger than permutation neighborhood~\cite{acharya2013optimal,orlitsky2015competitive,valiant2017automatic} in allowing auxiliary knowledge of frequency order among symbols, and is thus a very strong notion of instance-optimality. Furthermore, we show in \cref{ssec:per_instance_alg_dp} that our additive neighborhoods are (up to constants) the smallest that still allow for instance-optimal algorithms.
    \item Under such additive neighborhoods, we propose a new non-DP algorithm and prove it to be instance-optimal (\cref{ssec:per_instance_alg_non_dp}). Our algorithm resembles the idea of the Good-Turing estimator (which is known to be instance-optimal in prior works), while ensuring significantly smaller sensitivity to adjacent datasets (in the DP sense). This reduced sensitivity is the key that makes our algorithm easier to privatize. We then propose a DP version of this algorithm and show that it is instance-optimal amongst DP algorithms (\cref{ssec:per_instance_alg_dp}).
    \item We validate the performance of our instance-optimal estimators via experiments (\cref{sec:experiment}), and show the reward from studying instance-optimality: while the Add-constant (DP) algorithm is already minimax-optimal, our instance-optimal algorithms achieve significantly better performance on many instances of practical interest, such as power-law distributions and real-word token distributions. 
\end{itemize}

\subsection{Technical Contributions}
\label{ssec:technical_innovations}

\paragraph{Generalized Assouad's method for decomposable statistical distance}  Standard tools for proving lower bounds, such as (DP) Assouad's method~\cite{yu1997assouad,acharya2021differentially,feldman2024instance}, only apply to symmetric statistical distance, which is not satisfied by KL divergence. To prove KL per-instance lower bound, we propose generalized (DP) Assouad's method (\cref{thm:thm_conditional_packing_nondp} and \ref{thm:thm_conditional_packing}) that applies to general \textit{decomposable statistical distance} (\cref{def:decomposable_dist}). This allows us to prove strong per-instance lower bounds. 

\paragraph{Reducing the Sensitivity of Good-Turing Estimator via ``Sampling Twice''} The challenge of privatizing the prior (near) instance-optimal Good-Turing estimators lie in its excessively high sensitivity to neighboring datasets. At the core, Good-Turing estimator is motivated by observing that estimating ``unseen'' symbols as zero-probability is biased, as the symbols that appeared exactly once would intuitively have similar probability values (which are non-zero). To correct this bias, it recursively uses the counts of symbols with frequency $t + 1$ to estimate the probability of symbols that appeared $t$ times, for all low-frequency symbols (e.g., $t=1, \cdots, n^{1/3}$ in Orlitsky and Suresh~\cite{orlitsky2015competitive}). Such computations suffer from high sensitivity, as one record could change the combined counts of symbols with frequency $t$ by $t=n^{1/3}$ in the worst-case. To address this limitation, we perform bias-correction via an alternative "sampling twice" approach: partitioning the dataset into two halves, using one for identifying ``unseen'' symbols and the other for estimating their combined mass. The resulting \cref{alg:nondp_per_instance_upper_alt} is conceptually simpler than prior Good-Turing estimators, while achieving tight instance-optimality guarantees (up to constants) and being empirically competitive in experiments. Crucially, this ``Sampling Twice'' design reduces sensitivity to just one (making the estimator easy to privatize), and is the key to achieving instance-optimality in the DP setting.

\paragraph{Effectively Privatizing Good-Turing Estimator via Calibrated Thresholding} The Good-Turing estimator is based on the intuition that the probability values of symbols that appeared zero times (in the dataset)  are similar to those of symbols that appeared exactly once. However, this simple zero-or-one thresholding does not remain effective for private algorithms, because DP estimates cannot differ significantly on neighboring input datasets. This forces us to use a larger threshold. To identify the best threshold for DP distribution estimation, given every possible threshold, we separately analyze (\cref{thm:non_dp_per_instance_upper_GT_add_constant_simplified_alt} and \ref{thm:dp_instance_LM_noisy_threshold_alt}) the DP estimation error due to False Negatives (FN) (i.e., high-probability symbols being below-threshold) versus False Positives (FP) (i.e., low-probability symbols being above-threshold). This analysis allows us to choose a threshold that balances the FN and FP errors, and achieves DP instance-optimality up to a constant factor (\cref{ssec:per_instance_alg_dp}).

\subsection{Other Related Works}

The Good-Turing estimator, originally developed by~\citep{good1953population} and simplified by \citep{gale1995good,orlitsky2015competitive}, has been widely observed to yield empirically accurate distribution estimates  (especially for language modeling tasks~\cite{katz1987estimation,chen1999empirical} under large vocabulary size). Several later works prove it to be minimax optimal~\cite{mcallester2000convergence,drukh2005concentration,orlitsky2003always} as well as instance-optimal \cite{acharya2013optimal,orlitsky2015competitive} for discrete distribution estimation in various metrics. However, to our knowledge, no prior works have designed or analyzed DP variants of Good-Turing estimator, which is the main technical innovation of this paper.

For DP discrete distribution estimation, minimax optimality is well-studied across a variety of models for DP, including central DP~\cite{diakonikolas2015differentially,acharya2021differentially}, local DP~\cite{duchi2013local,kairouz2016discrete,ye2018optimal,duchi2018minimax,acharya2019communication} and user-level DP~\cite{liu2020learning}. Existing works also focus on different error metrics, such as total variation distance (i.e., $\ell_1$ error)~\cite{diakonikolas2015differentially} and $\ell_2$ error~\cite{acharya2021differentially} (see a summary of results in~\citep[Table 1-2]{acharya2021differentially}). Our work falls into the central DP setting, and is the first to study the KL divergence error. Additionally, we establish the stronger instance-optimality (rather than minimax optimality). The closest work to our paper in the literature is \citep{feldman2024instance}, where the authors study instance-optimal density estimation in Wasserstein distance (that covers discrete distribution estimation in $\ell_1$ error as a special case). The results in \cite{feldman2024instance} are largely incomparable to our analysis,  due to the lack of (tight) conversions between total variation distance error and KL divergence error. (See \cref{ssec:dp_minmax} for a nuanced comparison between prior TV distance lower bounds and our KL lower bound for DP minimax distribution estimation.) Indeed, achieving optimality under KL often requires non-trivial change to the algorithm and analysis compared to $\ell_1$ error, as evidenced by the abundant literature on distribution estimation in KL~\cite{braess2004bernstein,paninski2003estimation,paninski2004variational,acharya2013optimal,orlitsky2015competitive}.

A closely related problem is  frequency estimation (a.k.a. histogram estimation), where the goal is to privately and accurately estimate the empirical distribution. DP histogram estimation is extensively studied in a variety of models for DP, including central DP~\cite{hay2009boosting,xu2013differentially}, local DP~\cite{bassily2015local,wang2017locally}, and shuffle DP~\cite{erlingsson2019amplification,cheu2019distributed,ghazi2021power}. Due to sampling error, algorithms for empirical frequency estimation typically do not directly yield good utility for distribution estimation.
\section{Tighter Instance-Optimality for Non-DP Estimation}

\label{sec:def_instance_optimal_nondp}

We first define instance-optimality under additive neighborhoods, that get around the limitations of previously studied neighborhood notions as discussed in \cref{sec:intro}.

\paragraph{Neighborhood Choices} As discussed in \cref{sec:intro}, a good neighborhood should be as small as possible and contain distributions that are similarly hard to estimate compared to the target distribution instance. Additive neighborhood is thus a natural choice as intuitively, the hardness of distribution estimation does not change a lot when perturbing each symbol’s probability by a small amount. The key question is how small should the scale be. To this end, our main design choices are

\begin{enumerate}
    \item For every symbol, we allow up to $t/n$ perturbation, for a small $t\geq 1$. This only changes its expected count up to $t$, and thus from the the empirically sampled count, it is hard to distinguish the perturbed distribution from the target distribution.
    \item For symbol with large $p_i>t/n$, we allow a larger $\sqrt{p_i/n}$ perturbation, for a small $t\geq 1$. This captures the fact that with constant probability, counts sampled from $\text{Poi}(np_i)$ and $\text{Poi}(np_i + \sqrt{np_i})$ are ``indistinguishable'' due to the statistical variance in sampling. Indeed these are related to the standard confidence intervals for binomial estimation~\cite{clopper1934use,agresti1998approximate}.
\end{enumerate}
On top of these design choices, we try to reduce the size of the neighborhood as much as possible. Thus we add a constraint that the \textit{combined mass} of small symbols (with $p_i\leq \frac{t}{n}$ given a small $t\geq 1$) should not change by more than $\frac{t}{n}$ – this added condition only makes the results stronger (as the resulting neighborhood is smaller).  As a result, we obtain the following additive neighborhood.
\begin{align}
    &N_{+}(p) = \Bigg\{q: \forall i\in[d],\, |q_i-p_i| \leq \min\left\{\frac{t}{n}, \sqrt{\frac{p_i}{n}}\right\}  \text{ and }\sum\limits_{i: p_i\leq \frac{t}{n}}q_i\leq \max\Big\{\frac{t}{n}, \sum\limits_{i: p_i\leq \frac{t}{n}}p_i\Big\}\Bigg\} \label{eqn:add_neighborhood_plus}
\end{align}
\paragraph{Per-Instance Lower Bounds} We next prove per-instance lower bound under this neighborhood $N_{+}(p)$. Our analysis requires generalized variants of Assouad's method (discussed in \cref{ssec:technical_innovations}). We defer all statements and proofs to \cref{app:nondp_per_instance_lower}, and only present the per-instance lower bound below: if $d\geq 2$ and $n\geq 4$, then we have
\begin{align}
    \text{lower}(p, n, N_{+}) \geq \Omega\left(\frac{\ln(1 + d_{small}(L'))}{n} + p_{small}(L')  \ln\left(1 + \frac{d_{small}(L')}{np_{small}(L')}\right) + \sum\limits_{i}\min\left\{p_i, \frac{1}{n}\right\}\right) \label{eqn:non_dp_lower_neighborhood_plus}
\end{align}
for any $t\geq 1$ and any set $L'\subseteq [d]$, where $p_{small}(L') = \sum\limits_{i\in L', p_i\leq \frac{t}{n}}p_i$ and $d_{small}(L') = \sum\limits_{i\in L', p_i\leq \frac{t}{n}} 1$. 

To interpret this lower bound, first observe that \eqref{eqn:non_dp_lower_neighborhood_plus} is always smaller than the $\Omega\left(1 + \frac{d}{n}\right)$ minimax lower bound, especially when there are many small symbols (with $p_i<\frac{1}{n}$) that jointly takes a small combined mass $p_{small}(L')$. As an extreme example, imagine a highly concentrated distribution $p=(1/3, 2/3, 0, \cdots, 0)$, then our per-instance lower bound \eqref{eqn:non_dp_lower_neighborhood_plus} is as small as $\frac{\ln(1 + d)}{n}$. This is significantly smaller than the $\ln(1 + \frac{d}{n})$ minimax lower bound  for high-dimensional setting ($d\rightarrow\infty$), thus correctly indicating that $p$ is an extremely easy-to-estimate distribution. Achieving error upper bound that matches this small per-instance lower bound \eqref{eqn:non_dp_lower_neighborhood_plus}, then serves as a strong requirement for instance-optimal algorithm to fulfill (which we prove in the next section).

\begin{table*}[t]
    \centering
    {\small\begin{tabular}{c|c|c}
    \toprule
         & KL Error Bound  & Reference \\\midrule
        $\text{lower}(p, n, N_{\pi}) $ & $\Omega\left(\sum_{t=0}^\infty\underset{x\sim \text{Poi}(np)}{\mathbb{E}}\left[\sum\limits_{i:x_i=t} p_i\ln\left(\frac{p_i \sum\limits_{j:x_j=t}1}{\sum_{j:x_j=t}p_j}\right)\right]\right)$  & \citep[Lemma 4 \& 5]{orlitsky2015competitive}\\
        & & $N_{\pi}$: permutation neighborhood \\\hline
        Add-constant & $\text{lower}(p, n, N_{\pi}) + \Omega\left(\min\left\{n^{-1/3}, \frac{d}{n}\right\}\right)$ & ~\citep[Lemma 1]{acharya2013optimal}\\\hline
        Good-Turing & $\text{lower}(p, n, N_{\pi}) +  O\left(\min\left\{n^{-1/3}, \frac{d}{n}\right\}\right)$ & \citep[Theorem 1]{orlitsky2015competitive}\\\hline
        Smoothed & $\text{lower} (p, n, N_{\pi}) +  O\left(\min\left\{n^{-1/2}, \frac{d}{n}\right\} \right)$ & \citep[Theorem 2]{orlitsky2015competitive}\\
        Good-Turing & & \citep[Theorem 2]{acharya2013optimal}\\\hline
        Any estimator & $\text{lower} (p, n, N_{\pi}) + \Omega\left(\min\left\{n^{-2/3}, \frac{d}{n}\right\}\right)$ & \citep[Theorem 3]{orlitsky2015competitive}\\
        \bottomrule
    \end{tabular}}
    \caption{Prior instance-optimality results of Non-DP estimators}
    \label{tab:nondp_kl_hist_prior_competitive}
\end{table*}

\subsection{An Instance-Optimal ``Sampling Twice'' Estimator}
\label{ssec:per_instance_alg_non_dp}

We now present a simple ``sampling twice'' \cref{alg:nondp_per_instance_upper_alt}, and prove that it is instance-optimal up to constants. Compared to the minimax optimal add-constant estimator, this algorithm observes that simply estimating ``unseen'' symbols as zero-probability (or constant-probability) is heavily biased, and use a conceptual ``sampling-twice'' procedure to correct the bias of low-frequency symbols. 

As discussed in \cref{ssec:technical_innovations}, this bias-correction idea originates from prior (near) instance-optimal Good-Turing algorithm~\cite{good1953population,gale1995good,orlitsky2015competitive} and maintains their utility benefit. The key benefit of our ``sampling-twice'' design lies in significantly reducing the estimator's sensitivity to neighboring dataset (making it easier to privatize), enabling instance-optimality under DP (as we will show in \cref{sec:def_instance_optimal_dp}).

\begin{algorithm}[t!]
	\caption{Non-DP  ``Sampling Twice''}
	\begin{algorithmic}
		\STATE {\bfseries Input:} Data partition ratio $\alpha = 0.5$. Independently sampled datasets $x \sim \text{Poi}(\alpha \cdot np)$ and $x'\sim  \text{Poi}((1 - \alpha) \cdot np)$ s.t. $x + x'\sim\text{Poi}(np)$. Threshold $\tau = 0$.   
		\STATE {{\bfseries Thresholding: }$L = \{i\in[d]: x_i\leq \tau\}$} 
		\STATE {{\bfseries Estimate combined mass for symbols in $L$: } $\tilde{c} = \max\left\{\sum_{i\in L}x'_i, 1 \right\}$}
        \STATE {{\bfseries Truncate individual estimates: } let $\tilde{x}_i = \max\left\{x'_i, 1\right\}$ for $i=1, \cdots, d$}
        \STATE {{\bfseries Return} $\mathcal{A}(x)$ with $\mathcal{A}(x)_i = \begin{cases}
            \frac{1}{N}\cdot \tilde{c} \cdot \frac{\tilde{x}_i}{\sum_{i\in L}\tilde{x}_i} & i\in L\\
            \frac{1}{N} \cdot  \tilde{x}_{i} & i \notin L 
        \end{cases}$ where $N = \tilde{c} + \sum_{i\notin L}\tilde{x}_i$}
	\end{algorithmic}
 \label{alg:nondp_per_instance_upper_alt}
\end{algorithm}

\paragraph{Instance-optimality Guarantee} In \cref{thm:non_dp_per_instance_upper_GT_add_constant_simplified_alt}, we prove a per-instance KL error upper bounds for the ``sampling twice'' estimator (\cref{alg:nondp_per_instance_upper_alt}). In  \cref{cor:nondp_per_instance_upper_additive}, we further prove that this upper bound can be rewritten as $\underset{x\sim\text{Poi}(np)}{\mathbb{E}}\left[\text{KL}(p, \mathcal{A}(x))\right] \leq O\left(\text{lower}(p,n, N_{+}) \right)$  for any choice of neighborhood size $t\geq 1$ s.t. $t \cdot e^{-t}\leq 1/\ln d$. Specifically, we can choose $t = \min\left\{1, 2\ln \ln d\right\}$. 
\subsection{Comparison to Prior Instance-Optimality Results}
\label{sec:prior_nondp_estimators}
In this section, we discuss and compare with prior instance-optimality results, which mainly cover two representative non-DP estimators -- the add-constant estimator and the Good-Turing (GT) estimator. 

\paragraph{Add-constant Estimator} Given by $\mathcal{A}(x) = x_i + c, \forall i\in[d]$ for a constant $c>0$, the add-constant estimator is one of the oldest and simplest distribution estimator. Its variants cover several known estimators, such as Laplace smoothing ($c=1$) and  Krichevsky-Trofimov estimator~\cite{krichevsky1981performance} ($c = 1/2$).

\paragraph{Good-Turing Estimator~\cite{good1953population}} The Good-Turing estimator~\cite{good1953population}, when combined with add-constant estimator, has long been observed to yield strong empirical performance. Many variants of GT estimator exist in the literature~\cite{good1953population,gale1995good,orlitsky2015competitive}. In the comparison, we use the simplified variant of Good-Turing estimators proposed in Orlitsky and Suresh~\cite{orlitsky2015competitive}, that is also provably (near) instance-optimal under the “permutation neighborhood”, i.e. all distributions obtained by permuting the distribution. 

We summarize these prior instance-optimality results under permutation neighborhood in \cref{tab:nondp_kl_hist_prior_competitive}. Apart from the fact that this neighborhood may be too large to be appropriate for some applications (as discussed in \cref{sec:intro}), the last row of \cref{tab:nondp_kl_hist_prior_competitive} shows a lower bound for the suboptimality – it shows that no algorithm can do better than an additive $\min\left\{n^{-2/3}, \frac{d}{n}\right\}$ compared to a hypothetical algorithm that already knows the neighborhood. Observe that this additive gap is a lot larger than the per-instance lower bounds for certain distributions, e.g., highly concentrated distributions. Thus, multiplicative instance optimality is not achievable with permutation neighborhood. By contrast, tight instance-optimality (up to constant) is achievable under our additive neighborhood, as proved in \cref{ssec:per_instance_alg_non_dp}. We will also discuss in \cref{ssec:per_instance_alg_dp} that our additive neighborhoods are the smallest (up to constants) that can still allow for the possibility of any DP instance-optimal algorithms.

\section{DP Instance-Optimality}

\label{sec:def_instance_optimal_dp}

In this section, we present the first results for DP instance-optimality of KL distribution estimation. 

\paragraph{Neighborhood Choices} To define appropriate DP instance-optimality objective, we modify the $N_{+}$ neighborhood~\eqref{eqn:add_neighborhood_plus} with perturbations calibrated to the privacy parameters. For a small $t\geq 1$, let
\begin{align}
    & N_{\leq \frac{t}{n\varepsilon}}\left(p, n\right) =  \Bigg\{q: |q_i - p_i|\leq \frac{t}{n\varepsilon} \text{ for any }i\in [d] \text{ and }\sum\limits_{i: p_i\leq \frac{t}{n\varepsilon}}q_i\leq \max\Big\{\frac{t}{n\varepsilon}, \sum\limits_{i: p_i\leq \frac{t}{n\varepsilon}}p_i\Big\}\Bigg\}
    \label{eqn:add_neighborhood_dp_small_symbols}
\end{align}
Compared to \eqref{eqn:add_neighborhood_plus}, the main change is that we allow the perturbation scale for individual symbols to be proportional to $1/\varepsilon$. This corresponds to the obliviousness of $(\varepsilon, \delta)$-DP algorithm to dataset changes up to $O\left(\frac{1}{\varepsilon}\right)$ hamming distance (which allows an additional $\frac{1}{n\varepsilon}$ change in probability).

\paragraph{Per-Instance Lower Bounds} Under the above neighborhood $N_{\leq \frac{t}{n\varepsilon}}\left(p, n\right) $ that is calibrated to $(\varepsilon, \delta)$-DP guarantee, we prove the below per-instance lower bound: if $\delta\leq \varepsilon$, $d\geq 2$, and $n\varepsilon\geq 1$, then
\begin{align}
    \text{lower}_{\varepsilon, \delta}\left(p, n, N_{\leq \frac{t}{n\varepsilon}}\right) \geq \,& \Omega\Bigg(\sum\limits_{i} \min\left\{p_i, \frac{1}{p_i} \cdot \frac{1}{n^2\varepsilon^2} \right\} + \frac{\ln\left(1 + d_{small}(L')\right)}{n\varepsilon} \nonumber\\
    & + p_{small}(L') \cdot \ln\left(1 + \frac{d_{small}(L')}{n\varepsilon \cdot  p_{small}(L') }\right) \Bigg) \label{eqn:dp_lower_main_body}
\end{align}
for any $t\geq 1$ and any set $L'\subseteq [d]$, where $p_{small}(L') = \sum\limits_{i\in L': p_i\leq \frac{t}{n\varepsilon}}p_i$ and $d_{small}(L') = \sum\limits_{i\in L': p_i\leq \frac{t}{n\varepsilon}}1$.

The proofs are deferred to \cref{app:dp_per_instance_lower}. The lower bound \eqref{eqn:dp_lower_main_body} consists of three terms:
\begin{enumerate}
    \item Cost of privacy for \textbf{large symbols $p_i\geq \frac{1}{n\varepsilon}$} is  $\frac{1}{n^2\varepsilon^2p_i}$, which is significantly smaller than the corresponding non-DP lower bound $\Omega(\frac{1}{n})$ in \eqref{eqn:non_dp_lower_neighborhood_plus} when $p_i$ is large enough, i.e., privacy is free for sufficiently large $p_i$.
    \item Cost of privacy for \textbf{combined mass estimate of small symbols $p_i\leq \frac{t}{n\varepsilon}$} is $\frac{\ln\left(1 + d_{small}(L')\right)}{n\varepsilon}$, which is non-zero whenever there exist small symbols in $L'$.
    \item Cost of privacy for \textbf{small symbols $p_i< \frac{t}{n\varepsilon}$} is $p_{small}(L') \cdot \ln\left(1 + \frac{d_{small}(L')}{n\varepsilon\cdot p_{small}(L')}\right) + \sum\limits_{i<\frac{1}{n\varepsilon}}p_i$, which is larger than the corresponding non-DP lower bound in \eqref{eqn:non_dp_lower_neighborhood_plus} for small symbols whenever $\varepsilon\ll 1$, i.e., in the high privacy regime.
\end{enumerate} 

\subsection{Privatized Instance-Optimal ``Sampling Twice'' Estimator}
\label{ssec:per_instance_alg_dp}

We then propose a privatized variant of ``sampling-twice'' \cref{alg:dp_per_instance_upper_alt} by applying the Laplace Mechanism~\cite{dwork2006calibrating}. The $\varepsilon$-DP guarantee follows from standard results in DP by observing that the $\ell_1$-sensitivity of $(x_1, \cdots, x_d, x'_1, \cdots, x'_d)$ is one.  The main novel ingredient (as discussed in \cref{ssec:technical_innovations}) is to choose a calibrated threshold for privately selecting a set of small symbols. This selected set is then used for estimating the combined mass of small symbols on fresh samples. An accurate combined mass estimate is the key to reducing KL error. 

\begin{algorithm}[t!]
	\caption{$\varepsilon$-DP ``Sampling Twice'' (Instance-Optimal)}
    \label{alg:dp_per_instance_upper_alt}
	\begin{algorithmic}
		\STATE {{\bfseries Inputs:} Data partition ratio $\alpha = 0.5$. Independently sampled datasets $x \sim \text{Poi}(\alpha \cdot np)$ and $x'\sim  \text{Poi}((1 - \alpha) \cdot np)$ s.t. $x + x'\sim\text{Poi}(np)$. Threshold $\tau = 4\ln d$. }
        \STATE {$L=\emptyset$}\\
        \FOR {symbol $i = 1, \cdots, d$}
        \STATE {{\bfseries Private Thresholding: }  If 
        $\tilde{x}_i \coloneqq x_i + z_i \leq  \frac{\tau}{\min\{\varepsilon, 1\}}$ for $z_i\sim\text{Lap}\left(0, \frac{1}{\varepsilon}\right)$, add $i$ to $L$}
		\ENDFOR
		\STATE {{\bfseries Estimate small symbols' combined mass: } $\tilde{c} = \max\left\{\sum_{i\in L}x'_i +  Lap\left(0, \frac{1}{\varepsilon}\right), \frac{1}{\min\{\varepsilon, 1\}} \right\}$}
        \STATE {{\bfseries Estimate individual large symbols: } for $i\in[d]\setminus L$, $\tilde{x}'_i = x'_i + z'_i$ for $z'_i\sim\text{Lap}\left(0, \frac{1}{\varepsilon}\right)$}
        \STATE {{\bfseries Truncation: } $\bar{x}_i = \max\left\{\tilde{x}_i, \frac{1}{\min\{\varepsilon, 1\}}\right\}$ for $i\in L$; $\bar{x}_i = (1-\alpha)\left(\max\left\{\tilde{x}_i, \frac{1}{\min\{\varepsilon, 1\}}\right\} + \max\left\{\tilde{x}'_i, \frac{1}{\min\{\varepsilon, 1\}}\right\}\right)$ for $i\in [d]\setminus L$}
        \STATE {{\bfseries Return} $\mathcal{A}(x)$ with $\mathcal{A}(x)_i = \begin{cases}
            \frac{1}{N}\cdot \tilde{c} \cdot \frac{\bar{x}'_i}{\sum_{i\in L}\bar{x}'_i} & i\in L\\
            \frac{1}{N} \cdot  \bar{x}'_{i} & i \notin L 
        \end{cases}$ where $N = \tilde{c} + \sum_{i\notin L}\bar{x}_i$}
	\end{algorithmic}
\end{algorithm}

\paragraph{Instance-optimality Guarantee} In \cref{thm:dp_instance_LM_noisy_threshold_alt}, we prove a per-instance  upper bound for  \cref{alg:dp_per_instance_upper_alt}. In \cref{cor:dp_per_instance_upper_additive}, we further prove that this upper bound   can be controlled by the per-instance lower bound under the combination of our Non-DP addditive neighborhood $N_+(p)$ in \eqref{eqn:add_neighborhood_plus} and our $(\varepsilon, \delta)$-DP calibrated neighborhood in \eqref{eqn:add_neighborhood_dp_small_symbols}. Specifically, under neighborhood size $t=24\ln d$, we prove that $ \underset{x\sim\text{Poi}(np)}{\mathbb{E}}\left[\text{KL}(p, \mathcal{A}(x))\right]  \leq  O\left(\text{lower}\left(p, n, N_{+}\cup N_{\leq \frac{t}{n\varepsilon}}\right)\right)$. 

\paragraph{Discussions on the Neighborhood Size} We have established instance-optimality under local neighborhood $N_{\frac{24\ln d}{n\varepsilon}}$. It is a natural question to ask whether the size of such neighborhoods could be further reduced to enable stronger instance-optimality notions. In \cref{thm:necessity_neighborhood_size}, we prove that this neighborhood size is necessary up to constants---no DP estimator can be instance-optimal for a neighborhood $N_{\leq \frac{\gamma\cdot \ln d}{n\varepsilon}}$ with $\gamma\leq o(1)$.
\section{Experiments}
\label{sec:experiment}

We now evaluate the performance of our algorithms and compare them with baselines. All experiments are performed on a MacOS intergrated CPU ($\leq$ 30 minutes) with 18GB RAM.  All reported performance numbers are averaged across five random trials of data sampling and estimators. 





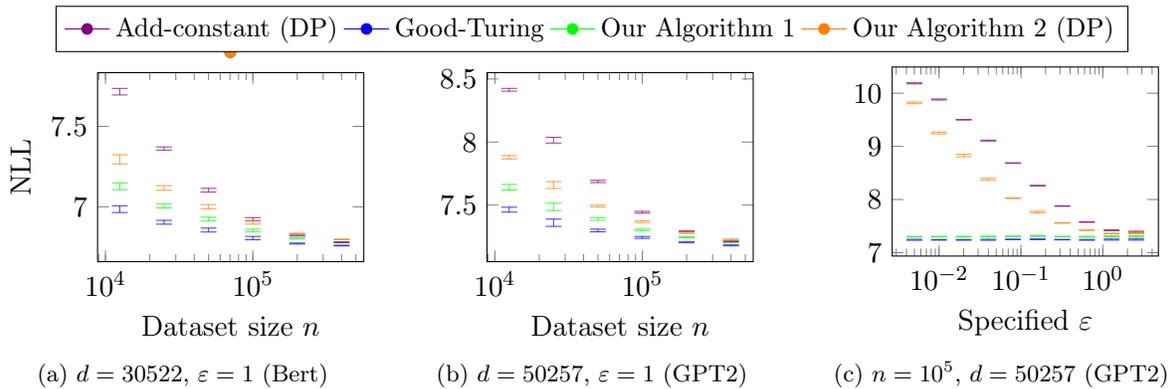
\begin{figure*}[h!]
    \centering
    
    \begin{tikzpicture}
        \begin{axis}[
            width=0.7\linewidth, 
            height=0.12\linewidth, 
            hide axis, 
            xmin=0, xmax=1, ymin=0, ymax=1,
            legend columns=4, 
            legend style={at={(0.5,0)}, anchor=south}, 
            legend cell align=left
        ]
            \addplot[violet, mark=*, mark size=2pt] coordinates {(0,0)};
            \addlegendentry{\small Add-constant (DP)}

            \addplot[blue, mark=*, mark size=2pt] coordinates {(0,0)};
            \addlegendentry{\small Good-Turing}

            \addplot[green, mark=*, mark size=2pt] coordinates {(0,0)};
            \addlegendentry{\small Our \cref{alg:nondp_per_instance_upper_alt}}

            \addplot[orange, mark=*, mark size=2pt] coordinates {(0,0)};
            \addlegendentry{\small Our \cref{alg:dp_per_instance_upper_alt} (DP)}
        \end{axis}
    \end{tikzpicture}

    \begin{subfigure}{0.32\textwidth}
        \centering
        \begin{tikzpicture}
            \begin{axis}[
            width=1\linewidth,
            height=0.8\linewidth,
            xlabel={Dataset size $n$},
            ylabel={NLL},
            xmode=log,
            only marks,
            mark size=3pt,
            ]
            \addplot[blue, error bars/.cd, y dir = both, y explicit, error bar style={opacity=0.75}]
            table[x=n, y=kl_gt_nondp_mean, y error=kl_gt_nondp_std, col sep=comma] {./data_lnd/reddit_bert/d_30522_eps_1_different_n.csv};
            \addplot[green, error bars/.cd, y dir = both, y explicit, error bar style={opacity=0.75}]
            table[x=n, y=kl_our_nondp_mean, y error=kl_our_nondp_std, col sep=comma] {./data_lnd/reddit_bert/d_30522_eps_1_different_n.csv};
            \addplot[violet, error bars/.cd, y dir = both, y explicit, error bar style={opacity=0.75}]
            table[x=n, y=kl_naive_mean, y error=kl_naive_std, col sep=comma] {./data_lnd/reddit_bert/d_30522_eps_1_different_n.csv};
            \addplot[orange, error bars/.cd, y dir = both, y explicit, error bar style={opacity=0.75}]
            table[x=n, y=kl_our_alg2_mean, y error=kl_our_alg2_std, col sep=comma] {./data_lnd/reddit_bert/d_30522_eps_1_different_n.csv};
            \end{axis}
        \end{tikzpicture}
        \caption{$d=30522$, $\varepsilon = 1$ (Bert)}
        \label{fig:gain_vs_samples_reddit_sample_twice_bert}
    \end{subfigure}
    \hfill
    \begin{subfigure}{0.32\textwidth}
        \centering
        \begin{tikzpicture}
            \begin{axis}[
            width=1\linewidth,
            height=0.8\linewidth,
            xlabel={Dataset size $n$},
            xmode=log,
            only marks,
            mark size=3pt,
            ]
            \addplot[blue, error bars/.cd, y dir = both, y explicit, error bar style={opacity=0.75}]
            table[x=n, y=kl_gt_nondp_mean, y error=kl_gt_nondp_std, col sep=comma] {./data_lnd/reddit_gpt2/d_50257_eps_1_different_n.csv};
            \addplot[green, error bars/.cd, y dir = both, y explicit, error bar style={opacity=0.75}]
            table[x=n, y=kl_our_nondp_mean, y error=kl_our_nondp_std, col sep=comma] {./data_lnd/reddit_gpt2/d_50257_eps_1_different_n.csv};
            \addplot[violet, error bars/.cd, y dir = both, y explicit, error bar style={opacity=0.75}]
            table[x=n, y=kl_naive_mean, y error=kl_naive_std, col sep=comma] {./data_lnd/reddit_gpt2/d_50257_eps_1_different_n.csv};
            \addplot[orange, error bars/.cd, y dir = both, y explicit, error bar style={opacity=0.75}]
            table[x=n, y=kl_our_alg2_mean, y error=kl_our_alg2_std, col sep=comma] {./data_lnd/reddit_gpt2/d_50257_eps_1_different_n.csv};
            \end{axis}
        \end{tikzpicture}
        \caption{$d=50257$, $\varepsilon = 1$ (GPT2)}
        \label{fig:gain_vs_reddit_sample_twice_gpt}
    \end{subfigure}
    \hfill
    \begin{subfigure}{0.32\textwidth}
        \centering
        \begin{tikzpicture}
            \begin{axis}[
            width=1\linewidth,
            height=0.8\linewidth,
            xlabel={Specified $\varepsilon$},
            xmode=log,
            only marks,
            mark size=3pt,
            ]
            \addplot[blue, error bars/.cd, y dir = both, y explicit, error bar style={opacity=0.75}]
            table[x=eps, y=kl_gt_nondp_mean, y error=kl_gt_nondp_std, col sep=comma] {./data_lnd/reddit_gpt2/n_100000_d_50257_different_eps.csv};
            \addplot[green, error bars/.cd, y dir = both, y explicit, error bar style={opacity=0.75}]
            table[x=eps, y=kl_our_nondp_mean, y error=kl_our_nondp_std, col sep=comma] {./data_lnd/reddit_gpt2/n_100000_d_50257_different_eps.csv};
            \addplot[violet, error bars/.cd, y dir = both, y explicit, error bar style={opacity=0.75}]
            table[x=eps, y=kl_naive_mean, y error=kl_naive_std, col sep=comma] {./data_lnd/reddit_gpt2/n_100000_d_50257_different_eps.csv};
            \addplot[orange, error bars/.cd, y dir = both, y explicit, error bar style={opacity=0.75}]
            table[x=eps, y=kl_our_alg2_mean, y error=kl_our_alg2_std, col sep=comma] {./data_lnd/reddit_gpt2/n_100000_d_50257_different_eps.csv};
            \end{axis}
        \end{tikzpicture}
        \caption{$n=10^5$, $d = 50257$ (GPT2)}
        \label{fig:gain_vs_eps_reddit_sample_twice}
    \end{subfigure}

    \caption{(Reddit Token Distribution Estimation) KL error versus dataset size $n$, distribution dimension $d$, and DP guarantee $\varepsilon$ for our methods compared with the simple minimax optimal Add-constant (DP) baseline, and the strongest non-DP baseline of prior (near) instance-optimal Good-Turing estimator.}
    \label{fig:sampling_twice_reddit}
    \vspace{-0.5em}
\end{figure*}

\begin{figure*}[h!]
    \centering
    \begin{tikzpicture}
        \begin{axis}[
            width=0.7\linewidth, 
            height=0.12\linewidth, 
            hide axis, 
            xmin=0, xmax=1, ymin=0, ymax=1,
            legend columns=4, 
            legend style={at={(0.5,0)}, anchor=south}, 
            legend cell align=left
        ]
            \addplot[violet, mark=*, mark size=2pt] coordinates {(0,0)};
            \addlegendentry{\small Add-constant (DP)}

            \addplot[blue, mark=*, mark size=2pt] coordinates {(0,0)};
            \addlegendentry{\small Good-Turing}

            \addplot[green, mark=*, mark size=2pt] coordinates {(0,0)};
            \addlegendentry{\small Our \cref{alg:nondp_per_instance_upper_alt}}

            \addplot[orange, mark=*, mark size=2pt] coordinates {(0,0)};
            \addlegendentry{\small Our \cref{alg:dp_per_instance_upper_alt} (DP)}
        \end{axis}
    \end{tikzpicture}

    \begin{subfigure}{0.32\textwidth}
        \centering
        \begin{tikzpicture}
            \begin{axis}[
            width=1\linewidth,
            height=0.8\linewidth,
            xlabel={Dataset size $n$},
            ylabel={KL Error},
            xmode=log,
            only marks,
            mark size=3pt,
            ]
            \addplot[green, error bars/.cd, y dir = both, y explicit, error bar style={opacity=0.75}]
            table[x=n, y=kl_our_nondp_mean, y error=kl_our_nondp_std, col sep=comma] {./data_lnd/power_law_1/d_50000_eps_1_different_n.csv};
            \addplot[blue, error bars/.cd, y dir = both, y explicit, error bar style={opacity=0.75}]
            table[x=n, y=kl_gt_nondp_mean, y error=kl_gt_nondp_std, col sep=comma] {./data_lnd/power_law_1/d_50000_eps_1_different_n.csv};
            \addplot[violet, error bars/.cd, y dir = both, y explicit, error bar style={opacity=0.75}]
            table[x=n, y=kl_naive_mean, y error=kl_naive_std, col sep=comma] {./data_lnd/power_law_1/d_50000_eps_1_different_n.csv};
            \addplot[orange, error bars/.cd, y dir = both, y explicit, error bar style={opacity=0.75}]
            table[x=n, y=kl_our_alg2_mean, y error=kl_our_alg2_std, col sep=comma] {./data_lnd/power_law_1/d_50000_eps_1_different_n.csv};
            \end{axis}
        \end{tikzpicture}
        \caption{$d=50000$, $\varepsilon = 1$}
        \label{fig:gain_vs_samples_sample_twice_power_law_1}
    \end{subfigure}
    \hfill
    \begin{subfigure}{0.32\textwidth}
        \centering
        \begin{tikzpicture}
            \begin{axis}[
            width=1\linewidth,
            height=0.8\linewidth,
            xlabel={Data dimension $d$},
            xmode=log,
            only marks,
            mark size=3pt,
            ]
            \addplot[blue, error bars/.cd, y dir = both, y explicit, error bar style={opacity=0.75}]
            table[x=d, y=kl_gt_nondp_mean, y error=kl_gt_nondp_std, col sep=comma] {./data_lnd/power_law_1/n_2000_eps_1_different_d.csv};
            \addplot[green, error bars/.cd, y dir = both, y explicit, error bar style={opacity=0.75}]
            table[x=d, y=kl_our_nondp_mean, y error=kl_our_nondp_std, col sep=comma] {./data_lnd/power_law_1/n_2000_eps_1_different_d.csv};
            \addplot[violet, error bars/.cd, y dir = both, y explicit, error bar style={opacity=0.75}]
            table[x=d, y=kl_naive_mean, y error=kl_naive_std, col sep=comma] {./data_lnd/power_law_1/n_2000_eps_1_different_d.csv};
            \addplot[orange, error bars/.cd, y dir = both, y explicit, error bar style={opacity=0.75}]
            table[x=d, y=kl_our_alg2_mean, y error=kl_our_alg2_std, col sep=comma] {./data_lnd/power_law_1/n_2000_eps_1_different_d.csv};
            \end{axis}
        \end{tikzpicture}
        \caption{$n=2000$, $\varepsilon = 1$}
        \label{fig:gain_vs_dimension_sample_twice_power_law_1}
    \end{subfigure}
    \hfill
    \begin{subfigure}{0.32\textwidth}
        \centering
        \begin{tikzpicture}
            \begin{axis}[
            width=1\linewidth,
            height=0.8\linewidth,
            xlabel={Specified $\varepsilon$},
            xmode=log,
            only marks,
            mark size=3pt,
            ]
            \addplot[blue, error bars/.cd, y dir = both, y explicit, error bar style={opacity=0.75}]
            table[x=eps, y=kl_gt_nondp_mean, y error=kl_gt_nondp_std, col sep=comma] {./data_lnd/power_law_1/n_1000_d_10000_different_eps.csv};
            \addplot[green, error bars/.cd, y dir = both, y explicit, error bar style={opacity=0.75}]
            table[x=eps, y=kl_our_nondp_mean, y error=kl_our_nondp_std, col sep=comma] {./data_lnd/power_law_1/n_1000_d_10000_different_eps.csv};
            \addplot[violet, error bars/.cd, y dir = both, y explicit, error bar style={opacity=0.75}]
            table[x=eps, y=kl_naive_mean, y error=kl_naive_std, col sep=comma] {./data_lnd/power_law_1/n_1000_d_10000_different_eps.csv};
            \addplot[orange, error bars/.cd, y dir = both, y explicit, error bar style={opacity=0.75}]
            table[x=eps, y=kl_our_alg2_mean, y error=kl_our_alg2_std, col sep=comma] {./data_lnd/power_law_1/n_1000_d_10000_different_eps.csv};
            \end{axis}
        \end{tikzpicture}
        \caption{$n=1000$, $d = 10000$}
        \label{fig:gain_vs_eps_sample_twice_power_law_1}
    \end{subfigure}

    \caption{(Power law distribution $p_i\propto \frac{1}{i}$) KL error versus dataset size $n$, distribution dimension $d$, and DP guarantee $\varepsilon$ for our methods compared with the simple minimax optimal Add-constant (DP) baseline, and the strongest non-DP baseline of prior (near) instance-optimal Good-Turing estimator.}
    \vspace{-0.5em}
    \label{fig:power_law_1}
\end{figure*}
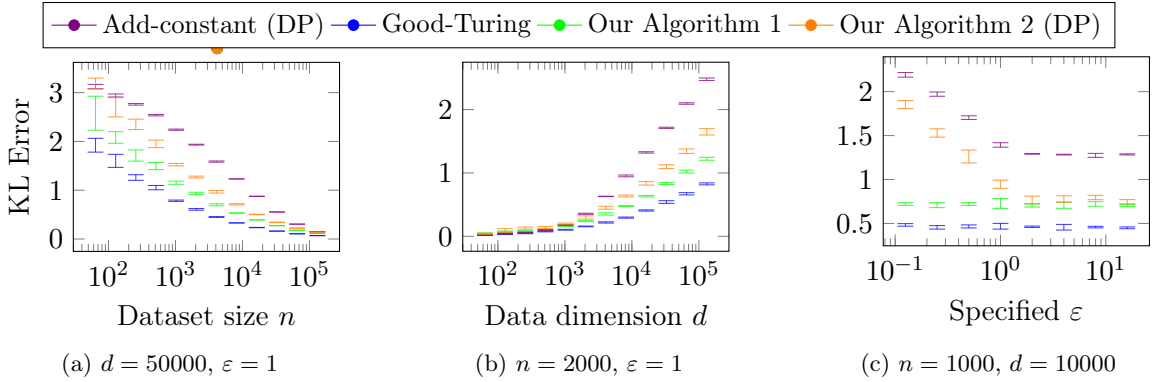

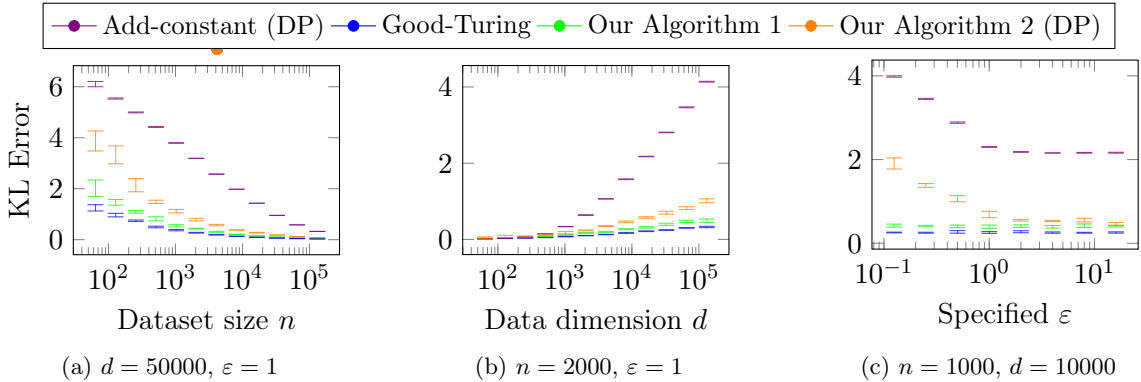
\begin{figure*}[h!]
    \centering
    
    \begin{tikzpicture}
        \begin{axis}[
            width=0.7\linewidth, 
            height=0.12\linewidth, 
            hide axis, 
            xmin=0, xmax=1, ymin=0, ymax=1,
            legend columns=4, 
            legend style={at={(0.5,0)}, anchor=south}, 
            legend cell align=left
        ]
            \addplot[violet, mark=*, mark size=2pt] coordinates {(0,0)};
            \addlegendentry{\small Add-constant (DP)}

            \addplot[blue, mark=*, mark size=2pt] coordinates {(0,0)};
            \addlegendentry{\small Good-Turing}

            \addplot[green, mark=*, mark size=2pt] coordinates {(0,0)};
            \addlegendentry{\small Our \cref{alg:nondp_per_instance_upper_alt}}

            \addplot[orange, mark=*, mark size=2pt] coordinates {(0,0)};
            \addlegendentry{\small Our \cref{alg:dp_per_instance_upper_alt} (DP)}
        \end{axis}
    \end{tikzpicture}

    \begin{subfigure}{0.32\textwidth}
        \centering
        \begin{tikzpicture}
            \begin{axis}[
            width=1\linewidth,
            height=0.8\linewidth,
            xlabel={Dataset size $n$},
            ylabel={KL Error},
            xmode=log,
            only marks,
            mark size=3pt,
            ]
            \addplot[green, error bars/.cd, y dir = both, y explicit, error bar style={opacity=0.75}]
            table[x=n, y=kl_our_nondp_mean, y error=kl_our_nondp_std, col sep=comma] {./data_lnd/power_law_1.5/d_50000_eps_1_different_n.csv};
            \addplot[blue, error bars/.cd, y dir = both, y explicit, error bar style={opacity=0.75}]
            table[x=n, y=kl_gt_nondp_mean, y error=kl_gt_nondp_std, col sep=comma] {./data_lnd/power_law_1.5/d_50000_eps_1_different_n.csv};
            \addplot[violet, error bars/.cd, y dir = both, y explicit, error bar style={opacity=0.75}]
            table[x=n, y=kl_naive_mean, y error=kl_naive_std, col sep=comma] {./data_lnd/power_law_1.5/d_50000_eps_1_different_n.csv};
            \addplot[orange, error bars/.cd, y dir = both, y explicit, error bar style={opacity=0.75}]
            table[x=n, y=kl_our_alg2_mean, y error=kl_our_alg2_std, col sep=comma] {./data_lnd/power_law_1.5/d_50000_eps_1_different_n.csv};
            \end{axis}
        \end{tikzpicture}
        \caption{$d=50000$, $\varepsilon = 1$}
        \label{fig:gain_vs_samples_sample_twice_power_law_1.5}
    \end{subfigure}
    \hfill
    \begin{subfigure}{0.32\textwidth}
        \centering
        \begin{tikzpicture}
            \begin{axis}[
            width=1\linewidth,
            height=0.8\linewidth,
            xlabel={Data dimension $d$},
            xmode=log,
            only marks,
            mark size=3pt,
            ]
            \addplot[blue, error bars/.cd, y dir = both, y explicit, error bar style={opacity=0.75}]
            table[x=d, y=kl_gt_nondp_mean, y error=kl_gt_nondp_std, col sep=comma] {./data_lnd/power_law_1.5/n_2000_eps_1_different_d.csv};
            \addplot[green, error bars/.cd, y dir = both, y explicit, error bar style={opacity=0.75}]
            table[x=d, y=kl_our_nondp_mean, y error=kl_our_nondp_std, col sep=comma] {./data_lnd/power_law_1.5/n_2000_eps_1_different_d.csv};
            \addplot[violet, error bars/.cd, y dir = both, y explicit, error bar style={opacity=0.75}]
            table[x=d, y=kl_naive_mean, y error=kl_naive_std, col sep=comma] {./data_lnd/power_law_1.5/n_2000_eps_1_different_d.csv};
            \addplot[orange, error bars/.cd, y dir = both, y explicit, error bar style={opacity=0.75}]
            table[x=d, y=kl_our_alg2_mean, y error=kl_our_alg2_std, col sep=comma] {./data_lnd/power_law_1.5/n_2000_eps_1_different_d.csv};
            \end{axis}
        \end{tikzpicture}
        \caption{$n=2000$, $\varepsilon = 1$}
        \label{fig:gain_vs_dimension_sample_twice_power_law_1.5}
    \end{subfigure}
    \hfill
    \begin{subfigure}{0.32\textwidth}
        \centering
        \begin{tikzpicture}
            \begin{axis}[
            width=1\linewidth,
            height=0.8\linewidth,
            xlabel={Specified $\varepsilon$},
            xmode=log,
            only marks,
            mark size=3pt,
            ]
            \addplot[blue, error bars/.cd, y dir = both, y explicit, error bar style={opacity=0.75}]
            table[x=eps, y=kl_gt_nondp_mean, y error=kl_gt_nondp_std, col sep=comma] {./data_lnd/power_law_1.5/n_1000_d_10000_different_eps.csv};
            \addplot[green, error bars/.cd, y dir = both, y explicit, error bar style={opacity=0.75}]
            table[x=eps, y=kl_our_nondp_mean, y error=kl_our_nondp_std, col sep=comma] {./data_lnd/power_law_1.5/n_1000_d_10000_different_eps.csv};
            \addplot[violet, error bars/.cd, y dir = both, y explicit, error bar style={opacity=0.75}]
            table[x=eps, y=kl_naive_mean, y error=kl_naive_std, col sep=comma] {./data_lnd/power_law_1.5/n_1000_d_10000_different_eps.csv};
            \addplot[orange, error bars/.cd, y dir = both, y explicit, error bar style={opacity=0.75}]
            table[x=eps, y=kl_our_alg2_mean, y error=kl_our_alg2_std, col sep=comma] {./data_lnd/power_law_1.5/n_1000_d_10000_different_eps.csv};
            \end{axis}
        \end{tikzpicture}
        \caption{$n=1000$, $d = 10000$}
        \label{fig:gain_vs_eps_sample_twice_power_law_1.5}
    \end{subfigure}

    \caption{(Power law distribution $p_i\propto \frac{1}{i^{1.5}}$) KL error versus dataset size $n$, distribution dimension $d$, and DP guarantee $\varepsilon$ for our methods compared with the simple minimax optimal Add-constant (DP) baseline, and the strongest non-DP baseline of prior (near) instance-optimal Good-Turing estimator.}
    \vspace{-0.5em}
    \label{fig:power_law_1.5}
\end{figure*}


\begin{figure*}[h!]
    \centering
    
    \begin{tikzpicture}
        \begin{axis}[
            width=0.7\linewidth, 
            height=0.12\linewidth, 
            hide axis, 
            xmin=0, xmax=1, ymin=0, ymax=1,
            legend columns=4, 
            legend style={at={(0.5,0)}, anchor=south}, 
            legend cell align=left
        ]
            \addplot[violet, mark=*, mark size=2pt] coordinates {(0,0)};
            \addlegendentry{\small Add-constant (DP)}
            
            \addplot[blue, mark=*, mark size=2pt] coordinates {(0,0)};
            \addlegendentry{\small Good-Turing}

            \addplot[green, mark=*, mark size=2pt] coordinates {(0,0)};
            \addlegendentry{\small Our \cref{alg:nondp_per_instance_upper_alt}}

            \addplot[orange, mark=*, mark size=2pt] coordinates {(0,0)};
            \addlegendentry{\small Our \cref{alg:dp_per_instance_upper_alt} (DP)}
        \end{axis}
    \end{tikzpicture}

    \begin{subfigure}{0.32\textwidth}
        \centering
        \begin{tikzpicture}
            \begin{axis}[
            width=1\linewidth,
            height=0.8\linewidth,
            xlabel={Dataset size $n$},
            ylabel={KL Error},
            xmode=log,
            only marks,
            mark size=3pt,
            ]
            \addplot[green, error bars/.cd, y dir = both, y explicit, error bar style={opacity=0.75}]
            table[x=n, y=kl_our_nondp_mean, y error=kl_our_nondp_std, col sep=comma] {./data_lnd/power_law_2/d_50000_eps_1_different_n.csv};
            \addplot[blue, error bars/.cd, y dir = both, y explicit, error bar style={opacity=0.75}]
            table[x=n, y=kl_gt_nondp_mean, y error=kl_gt_nondp_std, col sep=comma] {./data_lnd/power_law_2/d_50000_eps_1_different_n.csv};
            \addplot[violet, error bars/.cd, y dir = both, y explicit, error bar style={opacity=0.75}]
            table[x=n, y=kl_naive_mean, y error=kl_naive_std, col sep=comma] {./data_lnd/power_law_2/d_50000_eps_1_different_n.csv};
            \addplot[orange, error bars/.cd, y dir = both, y explicit, error bar style={opacity=0.75}]
            table[x=n, y=kl_our_alg2_mean, y error=kl_our_alg2_std, col sep=comma] {./data_lnd/power_law_2/d_50000_eps_1_different_n.csv};
            \end{axis}
        \end{tikzpicture}
        \caption{$d=50000$, $\varepsilon = 1$}
        \label{fig:gain_vs_samples_sample_twice_power_law_2}
    \end{subfigure}
    \hfill
    \begin{subfigure}{0.32\textwidth}
        \centering
        \begin{tikzpicture}
            \begin{axis}[
            width=1\linewidth,
            height=0.8\linewidth,
            xlabel={Data dimension $d$},
            xmode=log,
            only marks,
            mark size=3pt,
            ]
            \addplot[blue, error bars/.cd, y dir = both, y explicit, error bar style={opacity=0.75}]
            table[x=d, y=kl_gt_nondp_mean, y error=kl_gt_nondp_std, col sep=comma] {./data_lnd/power_law_2/n_2000_eps_1_different_d.csv};
            \addplot[green, error bars/.cd, y dir = both, y explicit, error bar style={opacity=0.75}]
            table[x=d, y=kl_our_nondp_mean, y error=kl_our_nondp_std, col sep=comma] {./data_lnd/power_law_2/n_2000_eps_1_different_d.csv};
            \addplot[violet, error bars/.cd, y dir = both, y explicit, error bar style={opacity=0.75}]
            table[x=d, y=kl_naive_mean, y error=kl_naive_std, col sep=comma] {./data_lnd/power_law_2/n_2000_eps_1_different_d.csv};
            \addplot[orange, error bars/.cd, y dir = both, y explicit, error bar style={opacity=0.75}]
            table[x=d, y=kl_our_alg2_mean, y error=kl_our_alg2_std, col sep=comma] {./data_lnd/power_law_2/n_2000_eps_1_different_d.csv};
            \end{axis}
        \end{tikzpicture}
        \caption{$n=2000$, $\varepsilon = 1$}
        \label{fig:gain_vs_dimension_sample_twice_power_law_2}
    \end{subfigure}
    \hfill
    \begin{subfigure}{0.32\textwidth}
        \centering
        \begin{tikzpicture}
            \begin{axis}[
            width=1\linewidth,
            height=0.8\linewidth,
            xlabel={Specified $\varepsilon$},
            xmode=log,
            only marks,
            mark size=3pt,
            ]
            \addplot[blue, error bars/.cd, y dir = both, y explicit, error bar style={opacity=0.75}]
            table[x=eps, y=kl_gt_nondp_mean, y error=kl_gt_nondp_std, col sep=comma] {./data_lnd/power_law_2/n_1000_d_10000_different_eps.csv};
            \addplot[green, error bars/.cd, y dir = both, y explicit, error bar style={opacity=0.75}]
            table[x=eps, y=kl_our_nondp_mean, y error=kl_our_nondp_std, col sep=comma] {./data_lnd/power_law_2/n_1000_d_10000_different_eps.csv};
            \addplot[violet, error bars/.cd, y dir = both, y explicit, error bar style={opacity=0.75}]
            table[x=eps, y=kl_naive_mean, y error=kl_naive_std, col sep=comma] {./data_lnd/power_law_2/n_1000_d_10000_different_eps.csv};
            \addplot[orange, error bars/.cd, y dir = both, y explicit, error bar style={opacity=0.75}]
            table[x=eps, y=kl_our_alg2_mean, y error=kl_our_alg2_std, col sep=comma] {./data_lnd/power_law_2/n_1000_d_10000_different_eps.csv};
            \end{axis}
        \end{tikzpicture}
        \caption{$n=1000$, $d = 10000$}
        \label{fig:gain_vs_eps_sample_twice_power_law_2}
    \end{subfigure}

    \caption{(Power law distribution $p_i\propto \frac{1}{i^2}$) KL error versus dataset size $n$, distribution dimension $d$, and DP guarantee $\varepsilon$ for our methods compared with the simple minimax optimal Add-constant (DP) baseline, and the strongest non-DP baseline of prior (near) instance-optimal Good-Turing estimator.}
    \vspace{-0.5em}
    \label{fig:power_law_2}
\end{figure*}
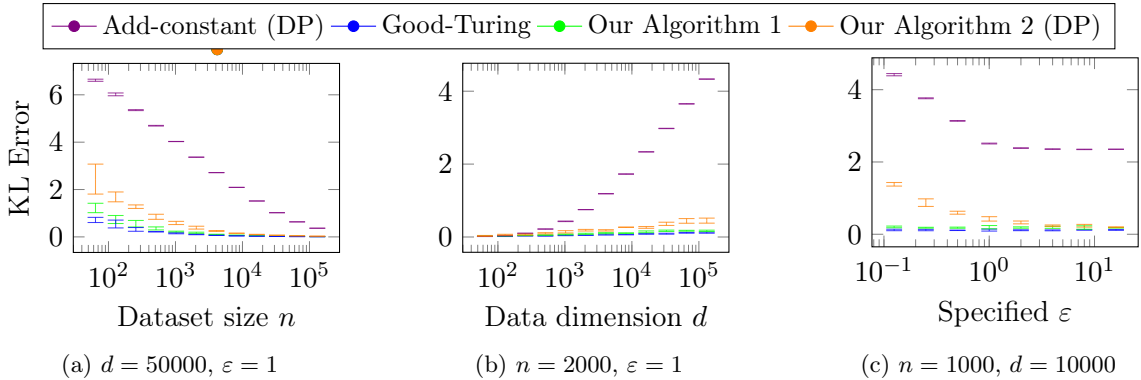

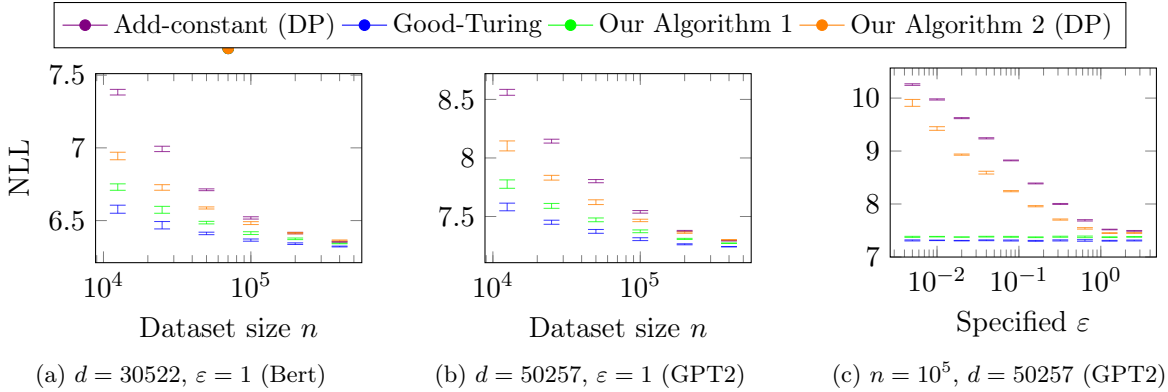
\begin{figure*}[h!]
    \centering
    
    \begin{tikzpicture}
        \begin{axis}[
            width=0.7\linewidth, 
            height=0.12\linewidth, 
            hide axis, 
            xmin=0, xmax=1, ymin=0, ymax=1,
            legend columns=4, 
            legend style={at={(0.5,0)}, anchor=south}, 
            legend cell align=left
        ]
            \addplot[violet, mark=*, mark size=2pt] coordinates {(0,0)};
            \addlegendentry{\small Add-constant (DP)}

            \addplot[blue, mark=*, mark size=2pt] coordinates {(0,0)};
            \addlegendentry{\small Good-Turing}

            \addplot[green, mark=*, mark size=2pt] coordinates {(0,0)};
            \addlegendentry{\small Our \cref{alg:nondp_per_instance_upper_alt}}

            \addplot[orange, mark=*, mark size=2pt] coordinates {(0,0)};
            \addlegendentry{\small Our \cref{alg:dp_per_instance_upper_alt} (DP)}
        \end{axis}
    \end{tikzpicture}

    \begin{subfigure}{0.32\textwidth}
        \centering
        \begin{tikzpicture}
            \begin{axis}[
            width=1\linewidth,
            height=0.8\linewidth,
            xlabel={Dataset size $n$},
            ylabel={NLL},
            xmode=log,
            only marks,
            mark size=3pt,
            ]
            \addplot[blue, error bars/.cd, y dir = both, y explicit, error bar style={opacity=0.75}]
            table[x=n, y=kl_gt_nondp_mean, y error=kl_gt_nondp_std, col sep=comma] {./data_lnd/enron_emails_bert/d_30522_eps_1_different_n.csv};
            \addplot[green, error bars/.cd, y dir = both, y explicit, error bar style={opacity=0.75}]
            table[x=n, y=kl_our_nondp_mean, y error=kl_our_nondp_std, col sep=comma] {./data_lnd/enron_emails_bert/d_30522_eps_1_different_n.csv};
            \addplot[violet, error bars/.cd, y dir = both, y explicit, error bar style={opacity=0.75}]
            table[x=n, y=kl_naive_mean, y error=kl_naive_std, col sep=comma] {./data_lnd/enron_emails_bert/d_30522_eps_1_different_n.csv};
            \addplot[orange, error bars/.cd, y dir = both, y explicit, error bar style={opacity=0.75}]
            table[x=n, y=kl_our_alg2_mean, y error=kl_our_alg2_std, col sep=comma] {./data_lnd/enron_emails_bert/d_30522_eps_1_different_n.csv};
            \end{axis}
        \end{tikzpicture}
        \caption{$d=30522$, $\varepsilon = 1$ (Bert)}
        \label{fig:gain_vs_samples_enron_emails_sample_twice_bert}
    \end{subfigure}
    \hfill
    \begin{subfigure}{0.32\textwidth}
        \centering
        \begin{tikzpicture}
            \begin{axis}[
            width=1\linewidth,
            height=0.8\linewidth,
            xlabel={Dataset size $n$},
            xmode=log,
            only marks,
            mark size=3pt,
            ]
            \addplot[blue, error bars/.cd, y dir = both, y explicit, error bar style={opacity=0.75}]
            table[x=n, y=kl_gt_nondp_mean, y error=kl_gt_nondp_std, col sep=comma] {./data_lnd/enron_emails_gpt2/d_50257_eps_1_different_n.csv};
            \addplot[green, error bars/.cd, y dir = both, y explicit, error bar style={opacity=0.75}]
            table[x=n, y=kl_our_nondp_mean, y error=kl_our_nondp_std, col sep=comma] {./data_lnd/enron_emails_gpt2/d_50257_eps_1_different_n.csv};
            \addplot[violet, error bars/.cd, y dir = both, y explicit, error bar style={opacity=0.75}]
            table[x=n, y=kl_naive_mean, y error=kl_naive_std, col sep=comma] {./data_lnd/enron_emails_gpt2/d_50257_eps_1_different_n.csv};
            \addplot[orange, error bars/.cd, y dir = both, y explicit, error bar style={opacity=0.75}]
            table[x=n, y=kl_our_alg2_mean, y error=kl_our_alg2_std, col sep=comma] {./data_lnd/enron_emails_gpt2/d_50257_eps_1_different_n.csv};
            \end{axis}
        \end{tikzpicture}
        \caption{$d=50257$, $\varepsilon = 1$ (GPT2)}
        \label{fig:gain_vs_enron_emails_sample_twice_gpt}
    \end{subfigure}
    \hfill
    \begin{subfigure}{0.32\textwidth}
        \centering
        \begin{tikzpicture}
            \begin{axis}[
            width=1\linewidth,
            height=0.8\linewidth,
            xlabel={Specified $\varepsilon$},
            xmode=log,
            only marks,
            mark size=3pt,
            ]
            \addplot[blue, error bars/.cd, y dir = both, y explicit, error bar style={opacity=0.75}]
            table[x=eps, y=kl_gt_nondp_mean, y error=kl_gt_nondp_std, col sep=comma] {./data_lnd/enron_emails_gpt2/n_100000_d_50257_different_eps.csv};
            \addplot[green, error bars/.cd, y dir = both, y explicit, error bar style={opacity=0.75}]
            table[x=eps, y=kl_our_nondp_mean, y error=kl_our_nondp_std, col sep=comma] {./data_lnd/enron_emails_gpt2/n_100000_d_50257_different_eps.csv};
            \addplot[violet, error bars/.cd, y dir = both, y explicit, error bar style={opacity=0.75}]
            table[x=eps, y=kl_naive_mean, y error=kl_naive_std, col sep=comma] {./data_lnd/enron_emails_gpt2/n_100000_d_50257_different_eps.csv};
            \addplot[orange, error bars/.cd, y dir = both, y explicit, error bar style={opacity=0.75}]
            table[x=eps, y=kl_our_alg2_mean, y error=kl_our_alg2_std, col sep=comma] {./data_lnd/enron_emails_gpt2/n_100000_d_50257_different_eps.csv};
            \end{axis}
        \end{tikzpicture}
        \caption{$n=10^5$, $d = 50257$ (GPT2)}
        \label{fig:gain_vs_eps_enron_emails_sample_twice}
    \end{subfigure}

    \caption{(Enron-emails Token Distribution Estimation) KL error versus dataset size $n$, distribution dimension $d$, and DP guarantee $\varepsilon$ for our methods compared with the simple minimax optimal Add-constant (DP) baseline, and the strongest non-DP baseline of prior (near) instance-optimal Good-Turing estimator.}
    \label{fig:sampling_twice_enron_emails}
    \vspace{-0.5em}
\end{figure*}

\begin{figure*}[h!]
    \centering
    
    \begin{tikzpicture}
        \begin{axis}[
            width=0.7\linewidth, 
            height=0.12\linewidth, 
            hide axis, 
            xmin=0, xmax=1, ymin=0, ymax=1,
            legend columns=4, 
            legend style={at={(0.5,0)}, anchor=south}, 
            legend cell align=left
        ]
            \addplot[violet, mark=*, mark size=2pt] coordinates {(0,0)};
            \addlegendentry{\small Add-constant (DP)}

            \addplot[blue, mark=*, mark size=2pt] coordinates {(0,0)};
            \addlegendentry{\small Good-Turing}

            \addplot[green, mark=*, mark size=2pt] coordinates {(0,0)};
            \addlegendentry{\small Our \cref{alg:nondp_per_instance_upper_alt}}

            \addplot[orange, mark=*, mark size=2pt] coordinates {(0,0)};
            \addlegendentry{\small Our \cref{alg:dp_per_instance_upper_alt} (DP)}
        \end{axis}
    \end{tikzpicture}

    \begin{subfigure}{0.32\textwidth}
        \centering
        \begin{tikzpicture}
            \begin{axis}[
            width=1\linewidth,
            height=0.8\linewidth,
            xlabel={Dataset size $n$},
            ylabel={NLL},
            xmode=log,
            only marks,
            mark size=3pt,
            ]
            \addplot[blue, error bars/.cd, y dir = both, y explicit, error bar style={opacity=0.75}]
            table[x=n, y=kl_gt_nondp_mean, y error=kl_gt_nondp_std, col sep=comma] {./data_lnd/mmlu_bert/d_30522_eps_1_different_n.csv};
            \addplot[green, error bars/.cd, y dir = both, y explicit, error bar style={opacity=0.75}]
            table[x=n, y=kl_our_nondp_mean, y error=kl_our_nondp_std, col sep=comma] {./data_lnd/mmlu_bert/d_30522_eps_1_different_n.csv};
            \addplot[violet, error bars/.cd, y dir = both, y explicit, error bar style={opacity=0.75}]
            table[x=n, y=kl_naive_mean, y error=kl_naive_std, col sep=comma] {./data_lnd/mmlu_bert/d_30522_eps_1_different_n.csv};
            \addplot[orange, error bars/.cd, y dir = both, y explicit, error bar style={opacity=0.75}]
            table[x=n, y=kl_our_alg2_mean, y error=kl_our_alg2_std, col sep=comma] {./data_lnd/mmlu_bert/d_30522_eps_1_different_n.csv};
            
            \end{axis}
        \end{tikzpicture}
        \caption{$d=30522$, $\varepsilon = 1$ (Bert)}
        \label{fig:gain_vs_samples_mmlu_sample_twice_bert}
    \end{subfigure}
    \hfill
    \begin{subfigure}{0.32\textwidth}
        \centering
        \begin{tikzpicture}
            \begin{axis}[
            width=1\linewidth,
            height=0.8\linewidth,
            xlabel={Dataset size $n$},
            xmode=log,
            only marks,
            mark size=3pt,
            ]
            \addplot[blue, error bars/.cd, y dir = both, y explicit, error bar style={opacity=0.75}]
            table[x=n, y=kl_gt_nondp_mean, y error=kl_gt_nondp_std, col sep=comma] {./data_lnd/mmlu_gpt2/d_50257_eps_1_different_n.csv};
            \addplot[green, error bars/.cd, y dir = both, y explicit, error bar style={opacity=0.75}]
            table[x=n, y=kl_our_nondp_mean, y error=kl_our_nondp_std, col sep=comma] {./data_lnd/mmlu_gpt2/d_50257_eps_1_different_n.csv};
            \addplot[violet, error bars/.cd, y dir = both, y explicit, error bar style={opacity=0.75}]
            table[x=n, y=kl_naive_mean, y error=kl_naive_std, col sep=comma] {./data_lnd/mmlu_gpt2/d_50257_eps_1_different_n.csv};
            \addplot[orange, error bars/.cd, y dir = both, y explicit, error bar style={opacity=0.75}]
            table[x=n, y=kl_our_alg2_mean, y error=kl_our_alg2_std, col sep=comma] {./data_lnd/mmlu_gpt2/d_50257_eps_1_different_n.csv};
            
            \end{axis}
        \end{tikzpicture}
        \caption{$d=50257$, $\varepsilon = 1$ (GPT2)}
        \label{fig:gain_vs_mmlu_sample_twice_gpt}
    \end{subfigure}
    \hfill
    \begin{subfigure}{0.32\textwidth}
        \centering
        \begin{tikzpicture}
            \begin{axis}[
            width=1\linewidth,
            height=0.8\linewidth,
            xlabel={Specified $\varepsilon$},
            xmode=log,
            only marks,
            mark size=3pt,
            ]
            \addplot[blue, error bars/.cd, y dir = both, y explicit, error bar style={opacity=0.75}]
            table[x=eps, y=kl_gt_nondp_mean, y error=kl_gt_nondp_std, col sep=comma] {./data_lnd/mmlu_gpt2/n_100000_d_50257_different_eps.csv};
            \addplot[green, error bars/.cd, y dir = both, y explicit, error bar style={opacity=0.75}]
            table[x=eps, y=kl_our_nondp_mean, y error=kl_our_nondp_std, col sep=comma] {./data_lnd/mmlu_gpt2/n_100000_d_50257_different_eps.csv};
            \addplot[violet, error bars/.cd, y dir = both, y explicit, error bar style={opacity=0.75}]
            table[x=eps, y=kl_naive_mean, y error=kl_naive_std, col sep=comma] {./data_lnd/mmlu_gpt2/n_100000_d_50257_different_eps.csv};
            \addplot[orange, error bars/.cd, y dir = both, y explicit, error bar style={opacity=0.75}]
            table[x=eps, y=kl_our_alg2_mean, y error=kl_our_alg2_std, col sep=comma] {./data_lnd/mmlu_gpt2/n_100000_d_50257_different_eps.csv};
            
            \end{axis}
        \end{tikzpicture}
        \caption{$n=10^5$, $d = 50257$ (GPT2)}
        \label{fig:gain_vs_eps_mmlu_sample_twice}
    \end{subfigure}

    \caption{(MMLU Token Distribution Estimation) KL error versus dataset size $n$, distribution dimension $d$, and DP guarantee $\varepsilon$ for our methods compared with the simple minimax optimal Add-constant (DP) baseline, and the strongest non-DP baseline of prior (near) instance-optimal Good-Turing estimator.}
    \label{fig:sampling_twice_mmlu}
    \vspace{-0.5em}
\end{figure*}
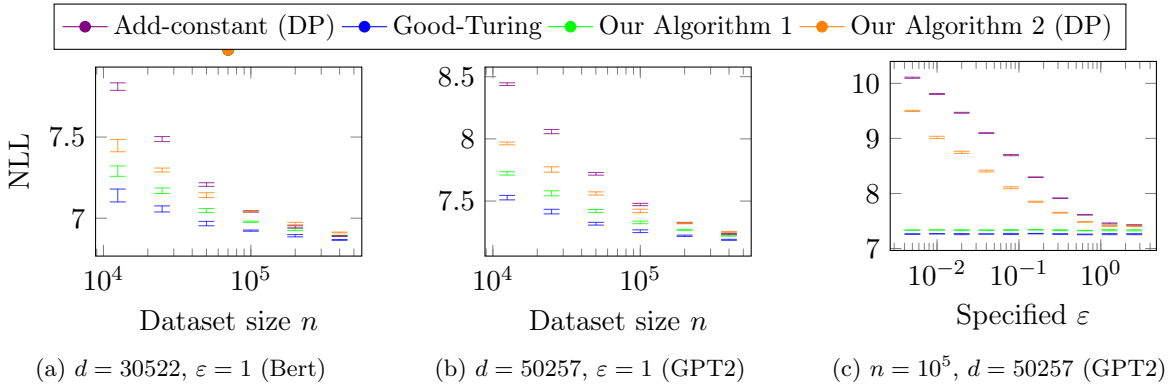

\paragraph{Datasets} We evaluate on both synthetic and real-world data distributions. For synthetic data, we evaluate power law distributions  $p_i\propto \frac{1}{i^\beta}$ over a range of parameters $\beta>0$. We choose power law because of its practical relevance: word frequencies from a text corpus have long been observed to roughly follow power law distributions~\cite{powers1998applications,piantadosi2014zipf,mitzenmacher2004brief,conrad2004power}. For real-word data, we experiment on randomly drawn tokens from  Reddit~\cite{volske-etal-2017-tl}, Enron-email~\cite{enron-email-dataset} and MMLU~\cite{hendryckstest2021,hendrycks2021ethics}, where each token is one (sensitive) record. We chose Reddit and Enron-email datasets because they are user-specific and thus bear a natural notion of privacy risk (compared to e.g., wikipedia), and are standard and widely used text datasets in the private learning literature~\cite{song2017machine,li2021large,lukas2023analyzing}. We additionally evaluate on MMLU to simulate diverse text domains. To vary the distribution dimension (specified by the number of all possible tokens), we use two types of tokenizers (GPT2 and Bert). 

One challenge for evaluating real-word dataset is the unknown ground-truth distribution $p$ (as only empirical samples are given). To address this, we independently sample two equal-size datasets $x$ and $x'$, thus ensuring $\mathbb{E}\left[{x'_i}/{\lVert x'\rVert_1}\right] = p_i$ for any $i\in[d]$. We then compute that
\begin{align}
    & \mathbb{E}\left[KL\left(p, \mathcal{A}(x)\right)\right] =  \mathbb{E}\left[\sum_{i=1}^dp_i\ln\left(\frac{p_i}{\mathcal{A}(x)_i}\right)\right] = \underbrace{\sum_{i=1}^dp_i\ln\left(p_i\right)}_{\text{Negative Entropy of }p} \underbrace{- \mathbb{E}\left[\sum_{i=1}^d\frac{x'_i}{\lVert x' \rVert_1} \cdot \ln\left( \mathcal{A}(x)_i\right)\right]}_{\text{Negative Log Likelihood (NLL)}} \nonumber
\end{align}
Observe that the entropy term is independent of the estimator. Thus in experiments we only report the negative log likelihood ratio term to compare different estimators $\mathcal{A}$.

\paragraph{Hyperparameters} Although our proposed \cref{alg:nondp_per_instance_upper_alt} and \ref{alg:dp_per_instance_upper_alt} are provably instance-optimal, they separately use two disjoint fractions of the dataset. Thus in the extreme scenarios when all symbols are above or below threshold (e.g., when $n, d$ are exceedingly large or small), \cref{alg:nondp_per_instance_upper_alt} and \ref{alg:dp_per_instance_upper_alt} incur twice as much noise compared to the add-one estimator, incurring a suboptimal multiplicative constant of two. To address this limitation, we perform grid search for the optimal hyperparameters over  $\alpha \in \{0.01, 0.1, 0.25, 0.5, 0.75, 0.9, 0.99\}$ and $\tau \in \{0, 0.0625, 0.125, 0.25, 0.5, 1, 2, 4\}\times \ln d$, and use the tuned hyperparameters $\tau = 0, \alpha = 0.5$ for \cref{alg:nondp_per_instance_upper_alt}, and $\tau = \min\left\{\frac{1}{\varepsilon}, 1.0\right\}\times \ln d, \alpha = 0.9$ for \cref{alg:dp_per_instance_upper_alt} in all experiments.

\paragraph{Observations} We show results on power law distributions $p_i\propto\frac{1}{i^{\beta}}$ for $\beta = 1, 1.5, 2$ (\cref{fig:power_law_1}, \ref{fig:power_law_1.5}, and \ref{fig:power_law_2}), Reddit corpus (\cref{fig:sampling_twice_reddit}), Enron-email corpus(\cref{fig:sampling_twice_enron_emails}); and MMLU corpus (\cref{fig:sampling_twice_mmlu}).  In all experiments, our DP instance-optimal \cref{alg:dp_per_instance_upper_alt} consistently outperforms the simple minimax optimal Add-constant (DP) baseline, and our non-DP instance-optimal \cref{alg:nondp_per_instance_upper_alt} is consistently competitive (within constants) to the strongest prior non-DP baseline of (near) instance-optimal Good-Turing~\cite{orlitsky2003always}. The gain of our algorithms are especially significant for real-world datasets, validating the effectiveness of our algorithms.  We also remark that instance optimality means that our algorithm will provably adapt to any input distribution, and no other algorithm can be significantly better (across a neighborhood). 
\section{Conclusion}
\label{sec:conclusion}
We provide tight instance-optimality analysis for private KL distribution estimation, in terms of achieving provably competitive error to the best possible estimator in a small additive local neighborhood of each instance. Furthermore, our constructed neighborhood's size is necessary up to constants for instance-optimality on the \textit{worst-case} instances. Additionally, we proved instance-optimality up to constants, and leave open the question of whether exact instance-optimality is achievable. Such results, if possible, would require improving the constants in per-instance privacy lower bounds and/or designing better estimators.

\section*{Acknowledgements}

The authors thank Audra McMillan and Satchit Sivakumar for insightful discussions on the literature, and Daogao Liu for valuable feedback on earlier drafts. Jiayuan
Ye is supported by the Apple Scholars in AI/ML PhD fellowship.

\bibliography{ref}
\bibliographystyle{alpha}


\newpage

\newpage
\tableofcontents
\appendix

\section{Tools for Lower Bounds}

Below we first define decomposable statistical distance.

\begin{definition}[Decomposable Statistical Distance] Let $d\in\mathbb{N}$ and let $dist: \Delta(d)\times \Delta(d)\rightarrow \mathbb{R}$ be a non-negative function such that $dist(p,q)=0$ if and only if $p=q$. We say $dist$ is decomposable, if for any disjoint sets of symbols $\mathcal{B}_1, \cdots, \mathcal{B}_k\subseteq[d]$, and for any distributions $p, q \in \Delta(d)$, it is the case that
    \begin{align}
        dist(p, q) \geq \sum_{i=1}^k \left(\sum_{j\in\mathcal{B}_i} p_j\right)\cdot dist(p|_i, q|_i)
    \end{align}
    where for any distribution $p\in\Delta(d)$, and any $i=1, \cdots, k$, we have denoted 
    \begin{align}
        p|_i = \begin{cases}
            \frac{p_j}{\sum_{j\in\mathcal{B}_i}p_j} & \text{ if } j\in\mathcal{B}_i\\
            0 & \text{ otherwise}
        \end{cases} \text{ for }i=1, \cdots, k \nonumber
    \end{align}
    \label{def:decomposable_dist}
\end{definition}

\begin{lemma}[KL divergence is decomposable]
    KL divergence is decomposable, that is, for any disjoint sets of symbols $\mathcal{B}_1, \cdots, \mathcal{B}_k$ and any $p, q \in \Delta(d)$, it is the case that
    \begin{align}
        KL(p, q)\geq \sum_{i=1}^k  \left(\sum_{j\in\mathcal{B}_i} p_j\right) \cdot KL\left(p|_i, q|_i\right)
    \end{align}
    \label{lem:KL_is_decomposable}
\end{lemma}
\begin{proof}
    Let $p, q \in \Delta(d)$. Denote $\mathcal{B}^c = [d]\setminus (\cup_{i=1}^k\mathcal{B}_i)$, then by definition we compute
    \begin{align}
        KL(p, q) = & \sum_{i=1}^k \sum_{j\in\mathcal{B}_i} p_j\ln\left(\frac{p_j}{q_j}\right) +  \sum_{j\in\mathcal{B}^c} p_j\ln\left(\frac{p_j}{q_j}\right)\\
        = & \sum_{i=1}^k  \left(\sum_{j\in\mathcal{B}_i} p_j\right) \cdot KL\left(p|_i, q|_i\right) + \sum_{i=1}^k\left(\sum_{j\in\mathcal{B}_i}p_j\right)\cdot \ln\left(\frac{\sum_{j\in\mathcal{B}_i}p_j}{\sum_{j\in\mathcal{B}_i}q_j}\right)\\
        & + \left(\sum_{j\in\mathcal{B}^c} p_j\right) \cdot KL\left(p|_{\mathcal{B}^c}, q|_{\mathcal{B}^c}\right) + \left(\sum_{j\in\mathcal{B}^c}p_j\right)\cdot \ln\left(\frac{\sum_{j\in\mathcal{B}^c}p_j}{\sum_{j\in\mathcal{B}^c}q_j}\right)\\
        \geq & \sum_{i=1}^k  \left(\sum_{j\in\mathcal{B}_i} p_j\right) \cdot KL\left(p|_i, q|_i\right)
    \end{align}
    where the last inequality is by non-negativity of KL divergence $KL\left(p|_{\mathcal{B}^c}, q|_{\mathcal{B}^c}\right)$ and $\sum_{i=1}^k\left(\sum_{j\in\mathcal{B}_i}p_j\right)\cdot \ln\left(\frac{\sum_{j\in\mathcal{B}_i}p_j}{\sum_{j\in\mathcal{B}_i}q_j}\right) + \left(\sum_{j\in\mathcal{B}^c}p_j\right)\cdot \ln\left(\frac{\sum_{j\in\mathcal{B}^c}p_j}{\sum_{j\in\mathcal{B}^c}q_j}\right)$.
\end{proof}

\begin{theorem}[Generalized Assouad's method for decomposable statistical distance]
\label{thm:thm_conditional_packing_nondp} Let $dist$ be a decomposable statistic distance as per \cref{def:decomposable_dist}. Let $\mathcal{B}_1, \cdots, \mathcal{B}_k\subseteq[d]$ be $k$ disjoint sets of symbols.  For each $i=1, \cdots, k$, let $\mathcal{P}_i$ be a set of distributions supported on $\mathcal{B}_i$. For fixed $w_1, \cdots, w_k\geq 0$ such that $\sum_{i=1}^kw_i = 1$, let $\mathcal{P}$ be the following composed packing set:
\begin{align}
    \mathcal{P} = \left\{q\coloneqq \sum_{i=1}^kw_ip^i \Big| p^i\in\mathcal{P}_i\right\}.
\end{align}
If the following two conditions hold:
\begin{enumerate}
    \item There exists a non-negative function $f$ such that
    \begin{align}
        \text{For any }  q\in \Delta(d)\text{, }\quad  \frac{1}{|\mathcal{P}_i|}\sum_{p \in \mathcal{P}_i} \mathbf{1}_{dist(p , q) \geq f(\mathcal{P}_i)} \geq \frac{1}{2}, \label{eqn:packing_assumption_nondp}
    \end{align}
    \item For each $i\in[k]$, there exists $\tau_i \geq 0$ and $\bar{p}^i\in\mathcal{P}_i$, such that for any fixed $p^j\in \mathcal{P}_j, j\neq i$, it holds that 
    \begin{align}
        \int \left(\min_{p^i\in\mathcal{P}_i}\frac{d\mathcal{S}(n, \sum_{j=1}^kw_jp^j)}{d\mathcal{S}(n, w_i\bar{p}^i + \sum_{j\neq i}w_jp^j)}\right) d\mathcal{S}(n, w_i\bar{p}^i + \sum_{j\neq i}w_jp^j) \geq \tau_i\label{eqn:dataset_distance_nondp}
    \end{align}
    where we have denoted $\mathcal{S}(n, p)$ as the distribution of histogram representation of dataset sampled from distribution $p\in\Delta(d)$ with target sample size $n$. 
\end{enumerate}
Then for any fixed $\varepsilon>0$, $\delta\leq \varepsilon$,  
\begin{align}
\min_{\mathcal{A}\text{ is $(\varepsilon, \delta)$-DP}}\ \max_{p\in \mathcal{P}}\  \underset{x\sim\mathcal{S}(n, p)}{\mathbb{E}}\left[dist(p, \mathcal{A}(x))\right] \geq \frac{1}{2}\sum_{i=1}^k w_i \cdot \tau_i \cdot f(\mathcal{P}_i)
\label{eqn:min_max_weighted_nondp}
\end{align}
\end{theorem}
\begin{proof}
    We will reduce the estimation problem over all $d$ symbols to the estimation problem over each bucket $\mathcal{B}_i$. For any distribution $p\in\Delta(d)$, denote its conditional distribution on $\mathcal{B}_i$ as 
    \begin{align}
        p|_i = \begin{cases}
            \frac{p_j}{\sum_{j\in\mathcal{B}_i}p_j} & \text{ if } j\in\mathcal{B}_i\\
            0 & \text{ otherwise}
        \end{cases} \text{ for }i=1, \cdots, k. \label{eqn:def_cond_p}
    \end{align}
    And for any $(\varepsilon, \delta)$-DP estimator $\mathcal{A}$ given dataset $x$ supported on $\mathcal{B}$, similarly denote
    \begin{align}
        \mathcal{A}(x)|_i = \begin{cases}
            \frac{\mathcal{A}(x)_j}{\sum_{j\in\mathcal{B}_i}\mathcal{A}(x)_j} & \text{ if } j\in\mathcal{B}_i\\
            0 & \text{ otherwise}
        \end{cases} \text{ for }i=1, \cdots, k.\label{eqn:def_cond_A}
    \end{align}
    Then by definition, for any fixed $i=1,\cdots, k$, we have 
    \begin{align}
        &\frac{1}{|\mathcal{P}|}\sum_{p\in\mathcal{P}}\underset{x\sim\mathcal{S}(n, p), \mathcal{A}}{\mathbb{E}}\left[dist(p|_i, \mathcal{A}(x)|_i)\right] \geq \frac{f(\mathcal{P}_i)}{|\mathcal{P}|}\sum_{p\in\mathcal{P}} \underset{x\sim\mathcal{S}(n, p), \mathcal{A}}{\mathbb{E}}\left[\mathbf{1}_{dist(p|_i, \mathcal{A}(x)|_i) \geq f(\mathcal{P}_i)}\right] \label{eqn:use_markov_nondp}\\
        = & \frac{f(\mathcal{P}_i)}{\prod_{j=1}^k|\mathcal{P}_j|}\sum_{p^j\in\mathcal{P}_j, j\neq i}\  \sum_{p^i\in\mathcal{P}_i}\underset{x\sim\mathcal{S}\left(n, \sum_{l=1}^dw_lp^l\right)}{\mathbb{E}}\left[\mathbb{E}_{\mathcal{A}}\left[\mathbf{1}_{dist(p^i, \mathcal{A}(x)|_i) \geq f(\mathcal{P}_i)}\right]\right]\\
        \geq & \frac{f(\mathcal{P}_i)}{\prod_{j=1}^k|\mathcal{P}_j|} \sum_{p^j\in\mathcal{P}_j, j\neq i}\  \underset{\bar{x}\sim\mathcal{S}\left(n, w_i\bar{p}^i + \sum_{l\neq i}w_lp^l\right)}{\mathbb{E}}\left[\min_{p^i\in\mathcal{P}_i}\frac{d\mathcal{S}(n, \sum_{j=1}^kw_jp^j)}{d\mathcal{S}(n, w_i\bar{p}^i + \sum_{j\neq i}w_jp^j)}(\bar{x}) \cdot \sum_{p^i\in\mathcal{P}_i}\mathbb{E}_{\mathcal{A}}\left[\mathbf{1}_{dist(p^i, \mathcal{A}(x)|_i) \geq f(\mathcal{P}_i)}\right]\right]\label{eqn:take_sum_packing_inside_nondp}\\
        \geq & \frac{f(\mathcal{P}_i)}{2} \cdot \underset{\bar{x}\sim\mathcal{S}\left(n, w_i\bar{p}^i + \sum_{l\neq i}w_lp^l\right)}{\mathbb{E}}\left[ \min_{p^i\in\mathcal{P}_i}\frac{d\mathcal{S}(n, \sum_{j=1}^kw_jp^j)}{d\mathcal{S}(n, w_i\bar{p}^i + \sum_{j\neq i}w_jp^j)}(\bar{x})\right] \label{eqn:use_packing_assumption_nondp}\\
        \geq & \frac{\tau_i}{2}\cdot  f(\mathcal{P}_i)\label{eqn:use_dataset_assumption_nondp}
    \end{align}
    where \eqref{eqn:use_markov_nondp} is by Markov inequality, \eqref{eqn:take_sum_packing_inside_nondp} is by definition and by moving the sum of $p^i\in\mathcal{P}_i$ inside expectation (as $\bar{x}$ and $\min_{p^i\in\mathcal{P}_i}\frac{d\mathcal{S}(n, \sum_{j=1}^kw_jp^j)}{d\mathcal{S}(n, w_i\bar{p}^i + \sum_{j\neq i}w_jp^j)}(\bar{x})$ are independent of the choice of $p^i$),  \eqref{eqn:use_packing_assumption_nondp} is by the packing assumption \eqref{eqn:use_packing_assumption_nondp}, and \eqref{eqn:use_dataset_assumption_nondp} is by the dataset distance assumption \eqref{eqn:dataset_distance_nondp}.

    We now use \eqref{eqn:use_packing_assumption_nondp} to prove lower bound on the expected KL error for estimating the whole distribution over all possible distributions in the packing set $\mathcal{P}$. By \cref{lem:KL_is_decomposable}, it follows that
    \begin{align}
        \frac{1}{|\mathcal{P}|}\sum_{p\in\mathcal{P}}\underset{x\sim\mathcal{S}(n, p), \mathcal{A}}{\mathbb{E}}\left[dist(p, \mathcal{A}(x))\right] \geq & \frac{1}{|\mathcal{P}|}\sum_{p\in\mathcal{P}}\sum_{i=1}^k w_i \cdot \underset{x\sim\mathcal{S}(n, p), \mathcal{A}}{\mathbb{E}}\left[dist(p|_i, \mathcal{A}(x)|_i)\right]\label{eqn:break_into_conditional_nondp}\\
        \geq & \frac{1}{2}\sum_{i=1}^k w_i \cdot \tau_i \cdot f(\mathcal{P}_i)\label{eqn:use_each_bucket_lower_nondp}
    \end{align}
    where the last inequality \eqref{eqn:use_each_bucket_lower_nondp} is by \eqref{eqn:use_packing_assumption_nondp}. This suffice to prove \eqref{eqn:min_max_weighted_nondp} by observing that average is smaller than maximum.
\end{proof}

\begin{theorem}[Generalized DP Assouad's method for decomposable statistical distance]
\label{thm:thm_conditional_packing} Let $dist$ be a decomposable statistic distance as per \cref{def:decomposable_dist}. Let $\mathcal{B}_1, \cdots, \mathcal{B}_k\subseteq[d]$ be $k$ disjoint sets of symbols.  For each $i=1, \cdots, k$, let $\mathcal{P}_i$ be a set of distributions supported on $\mathcal{B}_i$. For fixed $w_1, \cdots, w_k\geq 0$ such that $\sum_{i=1}^kw_i = 1$, let $\mathcal{P}$ be the following composed packing set:
\begin{align}
    \mathcal{P} = \left\{q\coloneqq \sum_{i=1}^kw_ip^i \Big| p^i\in\mathcal{P}_i\right\}.
\end{align}
For any $p\in \Delta(d)$, denote $\mathcal{S}(n, p)$ as the distribution of histogram representation of dataset sampled from $p$ with target sample size $n$. If the following two conditions hold:
\begin{enumerate}
    \item There exists a non-negative function $f$ such that
    \begin{align}
        \text{For any }  q\in \Delta(d)\text{, }\quad  \frac{1}{|\mathcal{P}_i|}\sum_{p \in \mathcal{P}_i} \mathbf{1}_{dist(p , q) \geq f(\mathcal{P}_i)} \geq \frac{1}{2}, \label{eqn:packing_assumption}
    \end{align}
    \item For each $i\in[k]$, there exists $\tau_i \geq 0$ and $\bar{p}^i\in\mathcal{P}_i$, such that for any fixed $p^j\in \mathcal{P}_j, j=1, \cdots, k$, it holds that
    \begin{align}
        \mathbb{E}_{(x, \bar{x})}\left[ \lVert x - \bar{x}\rVert_1 \right] \leq \tau_i \label{eqn:expected_dataset_distance}
    \end{align}
    for a coupling $(x, \bar{x})$ between distributions $\mathcal{S}(n, \sum_{j=1}^kw_jp^j)$ and $\mathcal{S}(n, w_i\bar{p}^i + \sum_{j\neq i}w_jp^j)$.
\end{enumerate}
Then for any fixed $\varepsilon>0$, $\delta\leq \varepsilon$, 
\begin{align}
\min_{\mathcal{A}\text{ is $(\varepsilon, \delta)$-DP}}\ \max_{p\in \mathcal{P}}\  \underset{x\sim\mathcal{S}(n, p)}{\mathbb{E}}\left[dist(p, \mathcal{A}(x))\right] \geq \sum_{i=1}^k w_i \cdot \left(\frac{1}{10} - 4\varepsilon \cdot \tau_i \right) \cdot f(\mathcal{P}_i)
\label{eqn:min_max_weighted}
\end{align}
\end{theorem}
\begin{proof}
    We will reduce the estimation problem over all $d$ symbols to the estimation problem over each bucket $\mathcal{B}_i$. For any distribution $p\in\Delta(d)$, we denote $p|_i$ as its conditional distribution on $\mathcal{B}_i$ as defined in \eqref{eqn:def_cond_p}. And for any $(\varepsilon, \delta)$-DP estimator $\mathcal{A}$ given dataset $x$ supported on $\mathcal{B}$, we denote $\mathcal{A}(x)|_i$ as the conditional estimate on $\mathcal{B}_i$ as defined in \eqref{eqn:def_cond_A}.
    Then by definition, for any fixed $i=1,\cdots, k$, we have 
    \begin{align}
        &\frac{1}{|\mathcal{P}|}\sum_{p\in\mathcal{P}}\underset{x\sim\mathcal{S}(n, p), \mathcal{A}}{\mathbb{E}}\left[dist(p|_i, \mathcal{A}(x)|_i)\right] \geq \frac{f(\mathcal{P}_i)}{|\mathcal{P}|}\sum_{p\in\mathcal{P}} \underset{x\sim\mathcal{S}(n, p), \mathcal{A}}{\mathbb{E}}\left[\mathbf{1}_{dist(p|_i, \mathcal{A}(x)|_i) \geq f(\mathcal{P}_i)}\right] \label{eqn:use_markov}\\
        = & \frac{f(\mathcal{P}_i)}{\prod_{j=1}^k|\mathcal{P}_j|}\sum_{p^j\in\mathcal{P}_j, j\neq i} \ \sum_{p^i\in\mathcal{P}_i}\underset{x\sim\mathcal{S}\left(n, \sum_{l=1}^dw_lp^l\right)}{\mathbb{E}}\left[\Pr_{\mathcal{A}}\left[dist(p^i, \mathcal{A}(x)|_i) \geq f(\mathcal{P}_i)\right]\right]\\
        \geq & \frac{f(\mathcal{P}_i)}{\prod_{j=1}^k|\mathcal{P}_j|} \sum_{p^j\in\mathcal{P}_j, j\neq i}\ \sum_{p^i\in\mathcal{P}_i}\underset{(x, \bar{x})}{\mathbb{E}}\left[ e^{-\varepsilon\cdot\lVert x - \bar{x}\rVert_1} \cdot \Pr_{\mathcal{A}}\left[dist(p^i, \mathcal{A}(\bar{x})|_i) \geq f(\mathcal{P}_i)\right] - \delta \cdot \lVert x - \bar{x} \rVert_1 \right]\label{eqn:use_group_privacy}\\
        \geq & \frac{f(\mathcal{P}_i)}{\prod_{j=1}^k|\mathcal{P}_j|} \sum_{p^j\in\mathcal{P}_j, j\neq i} \ \sum_{p^i\in\mathcal{P}_i}\underset{(x, \bar{x})}{\mathbb{E}}\left[ e^{-\varepsilon\cdot\tau_i\cdot C} \cdot \Pr_{\mathcal{A}}\left[dist(p^i, \mathcal{A}(\bar{x})|_i) \geq f(\mathcal{P}_i)\right] - \delta \cdot\tau_i\cdot C    - \mathbf{1}_{\lVert x - \bar{x} \rVert_1> \tau_i\cdot C}\right]\\
        = & \frac{f(\mathcal{P}_i)}{\prod_{j=1}^k|\mathcal{P}_j|} \sum_{p^j\in\mathcal{P}_j, j\neq i} \Bigg(e^{-\varepsilon\cdot\tau_i\cdot C} \cdot\underset{ \bar{x}}{\mathbb{E}}\left[ \sum_{p^i\in\mathcal{P}_i} \Pr_{\mathcal{A}}\left[dist(p^i, \mathcal{A}(\bar{x})|_i) \geq f(\mathcal{P}_i)\right]\right] - \delta \cdot\tau_i \cdot C\cdot |\mathcal{P}_i|    \nonumber\\
        & - \sum_{p^i}\Pr_{(x,\bar{x})}\left[ \lVert x - \bar{x} \rVert_1> \tau_i\cdot C\right]\Bigg) \label{eqn:take_sum_packing_inside}\\
        \geq & \left(\frac{e^{-\varepsilon\cdot \tau_i\cdot C}}{2} - \varepsilon \cdot \tau_i\cdot C - \frac{1}{C}\right)\cdot  f(\mathcal{P}_i)\label{eqn:use_packing_assumption}
        \\
        > & \left(\frac{1}{10} - 4\varepsilon\cdot \tau_i\right) \cdot f(\mathcal{P}_i)   \label{eqn:choose_C}
    \end{align}
    where \eqref{eqn:use_markov} is by Markov inequality, \eqref{eqn:use_group_privacy} is by group privacy and holds for arbitrary constant $C>0$, \eqref{eqn:take_sum_packing_inside} is by moving the sum of $p^i\in\mathcal{P}_i$ inside expectation (as $\bar{x}$ is independent of the choice of $p^i$ given coupling $(x, \bar{x})$ between distributions $\mathcal{S}(n, \sum_{j=1}^dw_jp^j)$ and $\mathcal{S}(n, w_i\bar{p}^i + \sum_{j\neq i}w_jp^j)$),  \eqref{eqn:use_packing_assumption} is by the packing assumption \eqref{eqn:packing_assumption} and $\delta\leq \varepsilon$ and the dataset distance assumption \eqref{eqn:expected_dataset_distance}, and \eqref{eqn:choose_C} is by choosing $C = \frac{1}{4\varepsilon\cdot \tau_i}$.

    We now use \eqref{eqn:choose_C} to prove a lower bound on the expected error for estimating the whole distribution over all possible distributions in the packing set $\mathcal{P}$. By \cref{lem:KL_is_decomposable}, it follows that
    \begin{align}
        \frac{1}{|\mathcal{P}|}\sum_{p\in\mathcal{P}}\underset{x\sim\mathcal{S}(n, p), \mathcal{A}}{\mathbb{E}}\left[dist(p, \mathcal{A}(x))\right] \geq & \frac{1}{|\mathcal{P}|}\sum_{p\in\mathcal{P}}\sum_{i=1}^k w_i \cdot \underset{x\sim\mathcal{S}(n, p), \mathcal{A}}{\mathbb{E}}\left[dist(p|_i, \mathcal{A}(x)|_i)\right]\label{eqn:break_into_conditional}\\
        \geq & \sum_{i=1}^k w_i \cdot \left(\frac{1}{10} - 4\varepsilon \cdot \tau_i\right) \cdot f(\mathcal{P}_i)\label{eqn:use_each_bucket_lower}
    \end{align}
    where the last inequality \eqref{eqn:use_each_bucket_lower} is by \eqref{eqn:choose_C}. This suffice to prove \eqref{eqn:min_max_weighted} by observing that average is smaller than maximum.
\end{proof}

Finally, we provide two useful constructions of packing set  that satisfy \eqref{eqn:packing_assumption_nondp} and \eqref{eqn:packing_assumption}.
\begin{lemma}[Packing over Two Symbols]
    Let $\mathcal{B}=\{j_1, j_2\}$ be a set of two symbols. Given $0\leq \Delta\leq a \leq 1$, let $\mathcal{P}\coloneqq\left\{p, p^-\right\}$ be a packing set containing the following two distributions on $\mathcal{B}$. 
    \begin{align}
        p(j) = \begin{cases}
            a & j=j_1\\
            1-a & j=j_2\\
            0 & j\in[d]\setminus\{j_1, j_2\}
        \end{cases} \text{ and } p^-(j) = \begin{cases}
            a - \Delta & j=j_1\\
            1 - a + \Delta & j=j_2\\
            0 & j\in[d]\setminus\{j_1, j_2\}
        \end{cases}\nonumber
    \end{align}
    If $\Delta<\frac{a}{2}$, then 
    \begin{align}
        \text{For any }q\in\Delta(d)\quad  \frac{1}{2}\left(\mathbf{1}_{KL(p, q)\geq \frac{\Delta^2}{8a}} +  \mathbf{1}_{KL(p^-, q)\geq \frac{\Delta^2}{8a}}\right)\geq \frac{1}{2}
    \end{align}
    \label{lem:two_symbol_packing_lower_bound}
\end{lemma}
\begin{proof}
    We will prove this claim by separating the analysis for different $q$.  
    \begin{enumerate}
        \item If $q_{j_1} \leq a - \frac{\Delta}{2}$: by definition, we compute that
        \begin{align}
            KL(p, q) = & a \ln\left(\frac{a}{q_{j_1}}\right) + (1-a) \ln\left(\frac{1-a}{q_{j_2}}\right) \nonumber\\
            \geq & a - q_{j_1} +\frac{1}{2}\cdot \frac{\left(a - q_{j_1}\right)^2}{a} + (1-a) - q_{j_2}\label{eqn:non_dp_lower_use_ln1}\\
            \geq &  \frac{1}{8} \cdot \frac{\Delta^2}{a}
        \end{align}
        where \eqref{eqn:non_dp_lower_use_ln1} is by  $\ln(1 + x)\leq x - \frac{x^2}{2}$ for $x\leq 0$, and by $\ln(1 + x)\leq x$ for $x\geq 0$, and the last inequality is by $q_{j_1} + q_{j_2}\leq 1$ and by using the condition that $q_{j_1} \leq a - \frac{\Delta}{2}$.
        \item If $q_{j_1} > a - \frac{\Delta}{2}$: by  definition, we compute that
        \begin{align}
            KL(p^-, q) = & (a-\Delta) \ln\left(\frac{a-\Delta}{q_{j_1}}\right) + (1-a + \Delta)\ln\left(\frac{1-a+\Delta}{q_{j_2}}\right) \nonumber\\
            \geq & a-\Delta - q_{j_1} +\frac{1}{4}\cdot \frac{\left(a-\Delta - q_{j_1}\right)^2}{a-\Delta} + (1-a + \Delta) - q_{j_2}\label{eqn:non_dp_lower_use_ln2}\\
            \geq & \frac{1}{8}\cdot \frac{\Delta^2}{a}
        \end{align}
        where \eqref{eqn:non_dp_lower_use_ln2} is by  $\ln(1 + x)\leq x - \frac{x^2}{4}$ for $0\leq x<\frac{1}{2}$, and by $\ln(1 + x)\leq x$ for $x\geq 0$, and the last inequality is by $q_{j_1} + q_{j_2}\leq 1$ and by using the condition that $q_{j_1} > a - \frac{\Delta}{2}$.
    \end{enumerate}
\end{proof}

\begin{lemma}[Dirac Distribution Packing over $\kappa$ Symbols]
    Let $\mathcal{B}=\{j_1, \cdots, j_{\kappa}\}$ be a set of $\kappa\geq 2$ symbols. Let $\mathcal{P}\coloneqq\left\{p, p^-\right\}$ be a packing set containing the following dirac distributions on $\mathcal{B}$. 
    \begin{align}
        p^l(j) = \begin{cases}
            1 & j=j_l\\
            0 & j\in[d]\setminus\{j_l\}
        \end{cases} \text{ for }l=1, \cdots, \kappa\nonumber
    \end{align}
    Then it holds that 
    \begin{align}
        \text{For any }q\in\Delta(d)\quad  \frac{1}{\kappa} \sum_{l=1}^\kappa\mathbf{1}_{KL(p^l, q)\geq \ln\left(1 + \frac{\kappa}{4}\right)} \geq \frac{1}{2}
    \end{align}
    \label{lem:dirac_packing_lower_bound}
\end{lemma}
\begin{proof}
    We prove by contradiction, suppose that there exists $q\in\Delta(d)$, such that 
    \begin{align}
        \sum_{l=1}^\kappa\mathbf{1}_{KL(p^l, q)\geq \ln\left(1 + \frac{\kappa}{4}\right)} < \frac{\kappa}{2}
    \end{align}
    Then by definition of KL divergence, we have 
    \begin{align}
        \sum_{l=1}^\kappa \mathbf{1}_{q_{j_l}\leq \frac{1}{1 + \frac{\kappa}{4}}} < \frac{\kappa}{2} \label{eqn:dirac_packing_lower_bound_contradiction}
    \end{align}
    Observe that $\sum_{l=1}^\kappa \mathbf{1}_{q_{j_l}\leq \frac{1}{1 + \frac{\kappa}{4}}}$ is integer, \eqref{eqn:dirac_packing_lower_bound_contradiction} implies that 
    \begin{align}
        \sum_{l=1}^\kappa \mathbf{1}_{q_{j_l}\leq \frac{1}{1 + \frac{\kappa}{4}}} \leq \begin{cases}
        0 & \kappa = 2\\
        \frac{2\kappa}{3} - 1 & \kappa \geq 3
        \end{cases} \label{eqn:dirac_packing_lower_bound_contradiction_new}
    \end{align}
    Thus
    \begin{align}
        \sum_{l=1}^\kappa q_{j_l}\geq &\frac{1}{1 + \frac{\kappa}{4}} \cdot \sum_{l=1}^\kappa \mathbf{1}_{q_{j_l}>\frac{1}{1 + \frac{\kappa}{4}}} = \frac{1}{1 + \frac{\kappa}{4}} \cdot \left(\kappa - \sum_{l=1}^\kappa \mathbf{1}_{q_{j_l}\leq \frac{1}{1 + \frac{\kappa}{4}}}\right) \\
        \geq & \begin{cases}
            \frac{4}{3} & \kappa = 2\\
            \frac{1 + \frac{\kappa}{3}}{1 + \frac{\kappa}{4}} & \kappa \geq 3
        \end{cases} > 1 \label{eqn:dirac_packing_lower_bound_last}
    \end{align}
    where \eqref{eqn:dirac_packing_lower_bound_last} is by \eqref{eqn:dirac_packing_lower_bound_contradiction}. This contradicts $q\in\Delta(d)$.
\end{proof}
\section{Useful Lemmas for Poisson and Laplace Random Variables}

\begin{lemma}[Expectation of inverse Poisson random variable]
     Let $p\in[0,1], m\in\mathbb{N}$, and let $x\sim \text{Poi}(mp)$. Then
     \begin{align}
         \mathbb{E}\left[\frac{1}{x + 1}\right] \leq \frac{1}{mp}
     \end{align}
     \label{lem:poisson_inverse}
\end{lemma}
\begin{proof}
    By definition, we compute that 
    \begin{align}
        \mathbb{E}\left[\frac{1}{x + 1}\right] =& \sum_{t=0}^\infty \frac{1}{t+1} \cdot e^{-mp}\cdot \frac{(mp)^t}{t!} = \frac{1}{mp}\sum_{t=1}^\infty  e^{-mp}\cdot \frac{(mp)^t}{t!} =  \frac{1}{mp}\cdot \left(1 - e^{-mp}\right) \leq \frac{1}{mp}
    \end{align}
\end{proof}

\begin{lemma}[Tail bound for Sum of Poisson and Laplace Random Variables]
    For any $a, b, c>0$. Suppose that $x\sim \text{Poi}(a)$, $z\sim \text{Lap}\left(0, b\right)$, then we have that 
    \begin{align}
        \Pr\left[x + z \leq c\right] \leq  \frac{4}{3}e^{\left(- \frac{a}{3} + \frac{c}{2}\right)\cdot \frac{1}{\max\{b, 1\}}} \quad\text{ and }\quad\Pr\left[x + z \geq c\right] \leq  \frac{4}{3}e^{\frac{a-c}{2}\cdot \frac{1}{\max\{b, 1\}}}.
    \end{align}
    \label{lem:laplace_plus_poisson_tail_prob}
\end{lemma}

\begin{proof}
    We first compute the moment generating function of $x + z$, i.e., the convolution of Poisson random variable and Laplace noise. For any $\theta\in [-\frac{1}{b}, \frac{1}{b}]$, by definition, we have that
    \begin{align}
        M_{x + z}(\theta) = & \mathbb{E}\left[e^{\theta \cdot (x + z)}\right] = \mathbb{E}\left[e^{\theta \cdot x}\right] \cdot \mathbb{E}\left[e^{\theta \cdot Lap\left(0,b\right)}\right] =  e^{a(e^\theta - 1)}\cdot \frac{1}{1 - b^2\theta^2}
    \end{align}
    Then, by using Markov inequality, for $\theta\in [0, \frac{1}{b}\big)$, we have
    \begin{align}
        \Pr[\tilde{x}\leq c] = & \Pr\left[e^{-\theta (x + z)} \geq e^{-\theta c}\right] \leq \frac{M_{x+z}(-\theta)}{e^{-\theta c}} = \frac{e^{a(e^{-\theta} - 1) + \theta c}}{1 - b^2\theta^2}\label{eqn:use_assumption_tau}
    \end{align}
    By choosing $\theta = \frac{1}{2\max\{b, 1\}}$, then by observing $e^{-\theta} - 1\leq - \frac{2}{3}\theta$ for $\theta = \frac{1}{2\max\{b, 1\}}<\frac{1}{2}$, we prove that
    \begin{align}
        \Pr[\tilde{x}\leq c] \leq \frac{4}{3} \cdot e^{\left(-\frac{a}{3} + \frac{c}{2}\right) \cdot \frac{1}{\max\{b, 1\}}}
    \end{align}
    Similarly, we prove a bound for the upper tail by Markov inequality.
    \begin{align}
        \Pr[\tilde{x}\geq c] = & \Pr\left[e^{\theta (x + z)} \geq e^{\theta c}\right] \leq \frac{M_{x+z}(\theta)}{e^{\theta c}} =  \frac{e^{a(e^{\theta} - 1) - \theta c}}{1 -  \theta^2b^2} \leq \frac{4}{3}e^{\frac{a - c}{2}\cdot \frac{1}{\max\{b, 1\}}}\label{eqn:use_assumption_tau_left}
    \end{align}
    where the last inequality is by choosing  $\theta = \frac{1}{2\max\{b, 1\}}$. 
\end{proof}

\begin{lemma}[Bias of Truncated Laplace Random Variable]
    Let $\lambda\geq 0$, $n\in\mathbb{N}$. Let $x\sim\text{Poi}(\lambda)$ and let $Z\sim\text{Lap}(0, b)$. Then the noisy estimator given by $\tilde{x} = \max\left\{x + Z, c\right\}$ satisfies 
    \begin{align}
        0 \leq \mathbb{E}\left[\tilde{x} - \lambda\right] \leq b + c
    \end{align}
    \label{lem:bias_threshold_laplace}
\end{lemma}
\begin{proof}
    We first prove the left inequality in \cref{lem:bias_threshold_laplace}. By definition,
    \begin{align}
        \mathbb{E}\left[\tilde{x} - \lambda\right] = \mathbb{E}\left[\tilde{x} - x\right] = \mathbb{E}\left[Z\cdot \mathbf{1}_{x+Z\geq c} + (c - x)\cdot \mathbf{1}_{x+Z< c}\right] = \mathbb{E}[Z] + \mathbb{E}\left[(c - x - Z)\cdot \mathbf{1}_{x+Z<c}\right]\geq 0\nonumber
    \end{align}
    We then prove the right inequality in \cref{lem:bias_threshold_laplace}. By $\tilde{x} = \max\{x+Z, c\}$, we compute that
    \begin{align}
        \mathbb{E}\left[\tilde{x} - \lambda\right] = \mathbb{E}\left[\tilde{x} - x\right] \leq \mathbb{E}\left[\max\{c, |Z|\}\right]\leq \mathbb{E}[|Z|] + c = b + c
    \end{align}
\end{proof}

\begin{lemma}[Conditional Bias under Thresholding]
    Let $X$ be a random variable over $\mathbb{R}$. Then for any  $c\in\mathbb{R}$, 
    \begin{align}
        \mathbb{E}\left[X| X\geq c \right] \geq \mathbb{E}[X] \quad \text{ and }\mathbb{E}\left[X| X\leq c \right] \leq \mathbb{E}[X]
    \end{align}
    \label{lem:cond_bias}
\end{lemma}

\begin{proof}
    We first prove the first inequality in \cref{lem:cond_bias}. If $c\geq  \mathbb{E}[X]$, then $\mathbb{E}[X|x\geq c]\geq c\geq \mathbb{E}[X]$. If $c<\mathbb{E}[X]$, then by definition, 
    \begin{align}
        \mathbb{E}\left[(X-\mathbb{E}[X]) \cdot \mathbf{1}_{X> c} \right] \geq & \mathbb{E}\left[(X-\mathbb{E}[X]) \right] = 0
    \end{align}
    where the inequality is by $x - \mathbb{E}[X]\leq c - \mathbb{E}[X] \leq 0$ for $x\leq c$. The proof for the second inequality in \cref{lem:cond_bias} is similar. 
\end{proof}

\begin{corollary}[Conditional Bias of Truncated Sum of Poisson and Laplace Random Variable]
    Let $b>0$, $\lambda>0$, $c, d\in\mathbb{R}\cup\{-\infty, +\infty\}$. Let $x\sim\text{Poi}(\lambda)$ and let $Z\sim\text{Lap}(0, b)$. Then the noisy estimator given by $\tilde{x} = \max\left\{x + Z, c\right\}$ satisfies 
    \begin{align}
        \mathbb{E}\left[\tilde{x} - \lambda|x+Z\leq d\right] \leq b + c
        \label{eqn:cond_bias_truncated_poisson_laplace_negative}
    \end{align}
    and
    \begin{align}
        \mathbb{E}\left[\tilde{x} - \lambda|x+Z\geq d\right] \geq 0
        \label{eqn:cond_bias_truncated_poisson_laplace_positive}
    \end{align}
    \label{cor:cond_bias_truncated_poisson_laplace}
\end{corollary}
\begin{proof}
    We first prove \eqref{eqn:cond_bias_truncated_poisson_laplace_negative}. 
    By \cref{lem:cond_bias}, we have that
    \begin{align*}
        \mathbb{E}\left[\tilde{x} - \lambda|x+Z\leq d \right] \leq \mathbb{E}\left[\tilde{x} - \lambda\right] \leq b + c
    \end{align*}
    where the last inequality is by \cref{lem:bias_threshold_laplace}. We then prove \eqref{eqn:cond_bias_truncated_poisson_laplace_positive}. 
    By \cref{lem:cond_bias}, we have that
    \begin{align*}
        \mathbb{E}\left[\tilde{x} - \lambda|x+Z\geq d \right] \geq \mathbb{E}\left[\tilde{x} - \lambda\right] \geq 0
    \end{align*}
    where the last inequality is by \cref{lem:bias_threshold_laplace}.
\end{proof}

\begin{lemma}
    Let $b>0$ and $c\in \mathbb{R}$, and let $Z\sim\text{Lap}(0,b)$. Then 
    \begin{align}
        \mathbb{E}\left[Z|Z\geq c\right] \leq \max\{c, 0\} + b
    \end{align}
    \label{lem:laplace_cond_expectation}
\end{lemma}
\begin{proof}
    We separate our discussion for $c\geq 0$ and $c<0$.
    \begin{enumerate}
        \item If $c\geq 0$, we have that
        \begin{align}
            \Pr[Z=z|Z\geq c] = \begin{cases}
                \frac{\frac{1}{2b}e^{-\frac{z}{b}}}{\frac{1}{2}e^{-\frac{c}{b}}} = \frac{1}{b}e^{-\frac{z-c}{b}} & z\geq c\\
                0 & z<c
            \end{cases}
        \end{align}
        Thus
        \begin{align}
            \mathbb{E}\left[Z|Z\geq c\right] = \int_{c}^{+\infty} z\cdot \frac{1}{b}e^{-\frac{z-c}{b}}dz \leq O\left( c + b\right)
        \end{align}
        \item  If $c<0$, we have that
        \begin{align}
            \Pr[Z=z|Z\geq c] \leq \frac{\frac{1}{2b}e^{-\frac{|z|}{b}}}{\frac{1}{2}} = \frac{1}{b}e^{-\frac{|z|}{b}}
        \end{align}
        Thus
        \begin{align}
            \mathbb{E}\left[Z|Z\geq c\right] \leq \int_{c}^{+\infty} \frac{1}{b}e^{-\frac{|z|}{b}}dz \leq  \int_{-\infty}^{+\infty} \frac{1}{b}e^{-\frac{|z|}{b}}dz \leq O\left( b\right)
        \end{align}
    \end{enumerate}
\end{proof}

\begin{lemma}
    Let $\lambda>0$, $b>0$ and $c\in \mathbb{R}$. Let $X\sim\text{Poi}(\lambda)$ and $Z\sim\text{Lap}(0,b)$ be independent random variables. Then 
    \begin{align}
        \mathbb{E}\left[X + Z|X + Z\geq c\right] \leq \lambda + O\left(b + \max\{c, 0\}\right) 
    \end{align}
    \label{lem:sum_laplace_poisson_cond_expectation}
\end{lemma}
\begin{proof}
    By definition,
    \begin{align}
        \mathbb{E}\left[X+Z|X+Z\geq c\right] = &\sum_{t=0}^\infty\Pr[X=t]\cdot \left(t + \mathbb{E}\left[Z|Z\geq c - t\right]\right)\\
        \leq & \mathbb{E}[X] + \sum_{t=0}^\infty\Pr[X=t]\cdot O\left(\max\{c - t, 0\} + b\right)\label{eqn:use_cond_bias_lap}\\
        \leq & \lambda + O\left(b + \max\{c, 0\}\right) 
    \end{align}
    where \eqref{eqn:use_cond_bias_lap} is by applying \cref{lem:laplace_cond_expectation}

\end{proof}

\begin{lemma}[{KL Divergence between Poisson Distributions~\citep[Theorem 2]{short2013improved}}]
    Let $m, k> 0$ be fixed. Then the KL divergence between two Poisson distributions $\text{Poi}(m)$ and $\text{Poi}(k)$ satisfies
    \begin{align}
        \text{KL}\left(\text{Poi}\left(m\right), \text{Poi}\left(k\right)\right) = m - k + k\cdot \ln\left(\frac{k}{m}\right)
    \end{align}
    \label{lem:kl_poisson}
\end{lemma}

\begin{lemma}[Total Variation Distance between Poisson Distributions]
    Let $\lambda_1, \lambda_2> 0$ be fixed. If $\lambda_1\leq \lambda_2$, Then 
    \begin{align}
        \text{TV}\left(\text{Poi}\left(\lambda_1\right), \text{Poi}\left(\lambda_2\right)\right) \leq \frac{1}{2}\sqrt{\frac{(\lambda_1 - \lambda_2)^2}{\lambda_1}}
    \end{align}
    \label{lem:tv_poisson}
\end{lemma}
\begin{proof}
    By Pinsker's inequality, we have 
    \begin{align}
        \text{TV}\left(\text{Poi}\left(\lambda_1\right), \text{Poi}\left(\lambda_2\right)\right) \leq & \sqrt{\frac{\text{KL}\left(\text{Poi}\left(\lambda_2\right), \text{Poi}\left(\lambda_1\right)\right)}{2}}\\
        \leq & \sqrt{\frac{\lambda_2 - \lambda_1 + \lambda_1\cdot \ln\left(\frac{\lambda_1}{\lambda_2}\right)}{2}}\label{eqn:use_kl_poisson}\\
        \leq & \frac{1}{2}\sqrt{\frac{(\lambda_1 - \lambda_2)^2}{\lambda_1}}
    \end{align}
    where \eqref{eqn:use_kl_poisson} is by \cref{lem:kl_poisson}, and the last inequality is by $\ln(1 + x)\geq x - \frac{x^2}{2}$ for $x\geq0$.
\end{proof}

\begin{lemma}
    Let $p>0$, $m\in \mathbb{N}$, $c_1, c_2\in \mathbb{R}\cup \{+\infty, -\infty\}$, $b\geq 0$, and $c\geq \max\{b, 1\}$. Let $x\sim \text{Poi}(m p)$ and let $Z\sim\text{Lap}\left(0, b\right)$ be independent of $x$. Then the noisy estimator given by $\tilde{x} = \max\{x + Z, c\}$ satisfies 
    \begin{align}
        \mathbb{E}\left[\mathbf{1}_{c_1\leq x+Z\leq c_2} \cdot \left(p \ln\left(\frac{mp}{\tilde{x} } \right) +   \frac{\tilde{x} - mp}{m} \right)\right] \leq O\left(\frac{1}{m} + \frac{b^2+c^2}{m \cdot \max\{c, mp\}}\right)
    \end{align}
    \label{lem:base_lemma_kl}
\end{lemma}
\begin{proof}
    We separate the discussions for $p\leq \frac{c}{m}$ and  $p > \frac{c}{m}$.
    \begin{itemize}
        \item If $p\leq \frac{c}{m}$, by $\ln(1 + t)\leq t$ for any $t>-1$, we have that
        \begin{align}
            & \mathbb{E}\left[ \mathbf{1}_{c_1\leq x+Z\leq c_2} \cdot \left( p \ln\left(\frac{mp}{\tilde{x} } \right) +  \frac{\tilde{x} - mp}{m} \right)\right] \\
            \leq & \mathbb{E}\left[ \mathbf{1}_{c_1\leq x+Z\leq c_2} \cdot \left(\frac{(mp - \tilde{x})^2}{m\tilde{x}} + \frac{mp - \tilde{x}}{m} + \frac{\tilde{x} - mp}{m} \right)\right] \\
            \leq  & \mathbb{E}\left[ \frac{(mp - \tilde{x})^2}{mc} \right] \label{eqn:remove_indicator}\\
            \leq  & \mathbb{E}\left[ \frac{2(mp - x)^2 + 2(x - \tilde{x})^2}{mc} \right] \label{eqn:remove_indicator_follow}\\
            \leq & \mathbb{E}\left[ \frac{2mp}{mc} + \frac{4b^2 + 2c^2}{mc}  \right]\leq O\left(\frac{1}{m} + \frac{b^2 + c^2}{mc}\right)
        \end{align}
        where the  \eqref{eqn:remove_indicator} is by $\tilde{x}\geq c$ and by $(mp - \tilde{x})^2\geq 0$, \eqref{eqn:remove_indicator_follow} is by $(a + b)^2\leq  2a^2 + 2b^2$, and the last inequality is by $\mathbb{E}[(mp-x)^2] = mp$, $\mathbb{E}\left[(x - \tilde{x})^2\right]\leq \mathbb{E}[Z^2 + c^2] = 2b^2 + c^2$, and $p\leq \frac{c}{m}$.
        \item If $p> \frac{c}{m}$: Observe that by $\ln(y)\geq y - 1 - (y - 1)^2$ for any $y\geq \frac{1}{3}$, we have that
        \begin{align}
            & \mathbb{E}\left[\mathbf{1}_{c_1\leq x+Z\leq c_2} \cdot \left(p \ln\left(\frac{mp}{ \tilde{x} } \right) +  \frac{\tilde{x} - mp}{m}\right)  \cdot \mathbf{1}_{x+Z\geq \frac{1}{3}mp}\right] \\
            \leq & \mathbb{E}\left[\mathbf{1}_{c_1\leq x+Z\leq c_2} \cdot \left(  - p\cdot  \frac{\tilde{x} - mp}{mp} + p\cdot \frac{(\tilde{x} - mp)^2}{(mp)^2}  + \frac{\tilde{x} - mp}{m} \right)\cdot \mathbf{1}_{x+Z\geq \frac{1}{3}mp } \right] \label{eqn:second_to_last_low_prob_ln_term}\\
            = & \mathbb{E}\Bigg[  \mathbf{1}_{c_1\leq x+Z\leq c_2} \cdot \frac{(\tilde{x} - mp)^2}{m^2 p}  \cdot \mathbf{1}_{x+Z\geq \frac{1}{3}mp }\Bigg]  \\
            \leq & \frac{2\mathbb{E}\left[ Z^2 + c^2 + (x - mp)^2 \right]}{m^2p} =  O\left(\frac{b^2 + c^2}{m^2 p} + \frac{1}{m}\right) \label{eqn:last_low_prob_ln_term}
        \end{align}
        where  the first inequality in \eqref{eqn:last_low_prob_ln_term} is by $(a + b)^2\leq 2(a^2 + b^2)$ and $(\tilde{x} - x)^2\leq Z^2 + c^2$, and by  $Z^2+ c^ 2 + (x-mp)^2\geq 0$; the last equality in \eqref{eqn:last_low_prob_ln_term} is by $\mathbb{E}\left[Z^2\right] = \frac{1}{\varepsilon^2}$ and $\mathbb{E}\left[(x - mp)^2\right] = mp$.

        On the other hand, by \cref{lem:laplace_plus_poisson_tail_prob}, we have that $\Pr[ x + Z < \frac{1}{3}mp]\leq O\left(e^{-\frac{ mp}{3} \cdot \frac{1}{\max\{b, 1\}} }\right)$. Thus we have that
        \begin{align}
            & \mathbb{E}\left[\mathbf{1}_{c_1\leq x+Z\leq c_2} \cdot \left(p \ln\left(\frac{mp}{ \tilde{x} } \right) +  \frac{\tilde{x} - x}{m}\right) \cdot \mathbf{1}_{x+Z<\frac{1}{3}mp}  \right] \leq \frac{p\ln\left(\frac{mp}{c}\right) + p}{e^{\frac{mp}{3} \cdot \frac{1}{\max\{b, 1\}} }} \\
            = & \frac{\max\{b, 1\}}{m}\cdot \frac{x\ln(x) + x}{e^{x/3}}\Big|_{x=\frac{mp}{\max\{b, 1\}}} + \ln\left(\frac{\max\{b, 1\}}{c}\right) \cdot \frac{\max\{b, 1\}}{n}\cdot \frac{x}{e^{x/3}}\Big|_{x=\frac{mp}{\max\{b, 1\}}} \\
            \leq & O\left(\frac{\max\{b, 1\}^2}{m^2 \cdot p}\right)\leq  O\left( \frac{c^2}{m^2 \cdot p} \right) \label{eqn:base_lemma_second_to_last} 
        \end{align}
        where the inequality in \eqref{eqn:base_lemma_second_to_last} is by $\frac{x\ln(x)}{e^{x/6}}\leq O\left(\frac{1}{x}\right)$ and $\frac{x}{e^{x/6}}\leq O\left(\frac{1}{x}\right)$ for $x\geq 0$, and by $c\geq \max\{b, 1\}$ This suffice to prove the bound in the statement.
    \end{itemize}
\end{proof}
\section{Deferred Minimax Optimlity Results}

\subsection{Recap: Non-DP Minimax Results}
\label{app:nondp_minimax}
The minimax rate for non-DP distribution estimation in KL divergence error is well-studied~\cite{braess2004bernstein,paninski2003estimation,paninski2004variational}, as summarized below in \cref{tab:nondp_kl_hist_min_max}. 

\begin{table}[H]
    \centering
    \begin{tabular}{c|c|c}
    \toprule
        Estimator/Bound & Expected KL Error  & Reference \\\midrule
        Upper Bound & $\ln\left(1 + \frac{d}{n}\right)$ & \citep[Theorem 8]{braess2004bernstein}\\
        (Add-Constant Estimator)& & Recap: \cref{thm:non_dp_min_max_upper}\\\hline
        Lower Bound & $\Omega\left(\ln\left( 1 + \frac{d}{n} \right)\right)$ &  Recap: \cref{thm:lower_non_dp_minimax}\\
        \bottomrule
    \end{tabular}
    \caption{Non-DP Minimax Rates for Distribution Estimation in KL Divergences}
    \label{tab:nondp_kl_hist_min_max}
\end{table}

For completeness, below we offer simple proofs for the minimax upper and lower bounds.

\subsubsection{Recap: Non-DP Minimax KL Error Lower Bound}

The minimax lower bound for distribution estimation in KL divergence error is well-understood to be $\Omega(1 + \frac{d}{n})$, e.g., see the variational lower bounds in \cite{paninski2004variational}. For completeness, below we provide an alternative proof via the tools in this paper, i.e., the generalized DP Assouad's method for decomposable statistical distance (in our case KL divergence) in  \cref{lem:KL_is_decomposable}.

\begin{theorem}[Lower Bound for Non-DP Estimation]
    Let $d$, $n$ be fixed. Then for any estimator $\mathcal{A}$, we have that 
    \begin{align}
        \max_p\underset{x\sim \text{Mult}(n, p)}{\mathbb{E}}\Big[KL(p\lVert \mathcal{A}(x))\Big] \geq \begin{cases}
            \Omega\left(\frac{d - 1}{n}\right) & d -1 \leq 2n\\
            \Omega\left(\ln\left(\frac{d - 1}{n}\right)\right) & d - 1 > 2n
        \end{cases} = \Omega\left(\ln\left( 1 + \frac{d}{n} \right)\right)
    \end{align}
    \label{thm:lower_non_dp_minimax}
\end{theorem}

\paragraph{Proof sketch} The idea is to design $p$ to be uniform  over a random support of symbols with small probability mass, and then reduce the problem to the difficulty of inferring the support given limited samples. 

\begin{proof}
    \begin{itemize}
        \item \textbf{Dense Case $d-1\leq 2n$:} Assume $d\mod 2 = 1$ for convenience. We decompose $d$ symbols into $\frac{d-1}{2}$ buckets $\mathcal{B}_i = \{2i-1, 2i\}$ for ${i=1, \cdots, \frac{d-1}{2}}$. We then construct the packing as follows. For $i=1, \cdots, \frac{d-1}{2}$.
        \begin{align}
            &\text{For }i=1,\cdots,\frac{d-1}{2}\quad \mathcal{P}_i = \{\delta_{2i-1}, \delta_{2i}\}
        \end{align}
        where $\delta_i$ means point distribution on symbol $i$. We then construct a set of distributions.
        \begin{align}
            \mathcal{P} = \left\{p = \sum_{i=1}^{\frac{d-1}{2}}\frac{1}{n}\cdot p^i + \left(1 - \frac{d-1}{2n}\right)\cdot \delta_d: p^i\in \mathcal{P}_i\text{ for any }i=1, \cdots, \frac{d-1}{2}\right\}
        \end{align}
        Then by \cref{lem:KL_is_decomposable}, for any algorithm $\mathcal{A}$ we have that
        \begin{align}
            \frac{1}{|\mathcal{P}|}&\sum_{p\in\mathcal{P}} \underset{x\sim\text{Mult}(n, p)}{\mathbb{E}}\left[KL(p, \mathcal{A}(x))\right]  \geq \frac{1}{n |\mathcal{P}|}\sum_{i=1}^{\frac{d-1}{2}} \sum_{p\in\mathcal{P}} \underset{x\sim\text{Mult}(n, p)}{\mathbb{E}}\left[KL(p|_i, \mathcal{A}(x)|_i)\right]\\
            \geq & \frac{1}{ne|\mathcal{P}|}\sum_{i=1}^{\frac{d-1}{2}}\sum_{p\in\mathcal{P}}  \underset{x\sim\text{Mult}(n, p)}{\mathbb{E}}\left[KL(p|_i, \mathcal{A}(x)|_i)|x_{2i-1} = x_{2i} = 0\right]\\
            \geq & \frac{1}{ne }\sum_{i=1}^{\frac{d-1}{2}}\ln(2) = \frac{(d-1)\ln(2)}{2ne}\\
        \end{align}
        Thus for any $\mathcal{A}$, there must exist one $p\in\mathcal{P}$ such that $\underset{x\sim\text{Mult}(n, p)}{\mathbb{E}}\left[KL(p, \mathcal{A}(x))\right]\geq \Omega(\frac{d-1}{n})$.

        \item \textbf{Sparse Case: $d-1> 2 n$:}
        Denote $\kappa = \lfloor \frac{d-1}{n}\rfloor\geq 2$ and $k = n$ for convenience. We decompose $d$ symbols into $k$ buckets $\mathcal{B}_i = \{\kappa \cdot i- \kappa + 1, \kappa \cdot i\}$ for ${i=1, \cdots, k}$. We then construct the packing as follows.
        \begin{align}
            &\text{For }i=1,\cdots,k\quad \mathcal{P}_i = \{\delta_{\kappa \cdot i - \kappa + 1}, \cdots, \delta_{\kappa \cdot i}\}
        \end{align}
        where $\delta_i$ means point distribution on symbol $i$. 
         We then construct a set of distributions.
        \begin{align}
            \mathcal{P} = \left\{p = \sum_{i=1}^{k}\frac{1}{n}\cdot p^i: p^i\in \mathcal{P}_i\text{ for any }i=1, \cdots, k\right\}
        \end{align}
        Then by \cref{lem:KL_is_decomposable}, for any algorithm $\mathcal{A}$ we have that
        \begin{align}
            \frac{1}{|\mathcal{P}|} & \sum_{p\in\mathcal{P}} \underset{x\sim\text{Mult}(n, p)}{\mathbb{E}}\left[KL(p, \mathcal{A}(x))\right] \geq \frac{1}{n |\mathcal{P}|}\sum_{i=1}^{k} \sum_{p\in\mathcal{P}} \underset{x\sim\text{Mult}(n, p)}{\mathbb{E}}\left[KL(p|_i, \mathcal{A}(x)|_i)\right]\\
            \geq & \frac{1}{ne|\mathcal{P}|}\sum_{i=1}^{k}\sum_{p\in\mathcal{P}}  \underset{x\sim\text{Mult}(n, p)}{\mathbb{E}}\left[KL(p|_i, \mathcal{A}(x)|_i)|x_{\kappa i- \kappa +1} = \cdots = x_{\kappa i} = 0\right]\\
            \geq & \frac{1}{ne }\sum_{i=1}^{k}\ln(\kappa) = \frac{k\ln\kappa}{ne} = \frac{1}{e}\ln\left(\frac{d-1}{n}\right)\\
        \end{align}
        Thus for any $\mathcal{A}$, there must exist one $p\in\mathcal{P}$ such that $\underset{x\sim\text{Mult}(n, p)}{\mathbb{E}}\left[KL(p, \mathcal{A}(x))\right]\geq \Omega(\ln\left(\frac{d-1}{n}\right))$.
        \end{itemize}
\end{proof}


\subsubsection{Recap: Non-DP Minimax KL Error Upper Bound}

The minimax upper bound for distribution estimation is well-studied, and various works~\cite{braess2004bernstein,paninski2003estimation,paninski2004variational} have shown that simple add-constant estimators could achieve the optimal $O\left(\ln\left(1 + \frac{d}{n}\right)\right)$ minimax KL error. Below we offer a simple proof for completeness. 
\begin{theorem}[Non-DP Distribution Estimation KL Upper Bound]
\label{thm:non_dp_min_max_upper}
    There exists estimator $\mathcal{A}$ such that for any $p\in \Delta(d)$, given $n$ empirical data samples $x\sim \text{Mult}(n, p)$, it satisfies that
    \begin{align}
        \underset{x\sim \text{Mult}(n, p)}{\mathbb{E}}\Big[KL(p\lVert \mathcal{A}(x))\Big] \leq \ln\left(1 + \frac{d}{n}\right)
    \end{align}
\end{theorem}

\begin{proof}
    We use that the following simple add-constant estimator.
    \begin{align}
        \mathcal{A}(x)_i = \frac{1}{1 + \frac{d}{n}}\cdot \frac{x_i + 1}{n}  
    \end{align}
    Then by definition
    \begin{align}
        \underset{x\sim \text{Mult}(n, p)}{\mathbb{E}}\Big[KL(p\lVert \mathcal{A}(x))\Big] = & \sum_{i=1}^dp_i\underset{x\sim \text{Mult}(n, p)}{\mathbb{E}}\Big[\ln\left(\frac{np_i}{x_i + 1}\right)\Big] + \ln\left(1 + \frac{d}{n}\right)\\
        \leq & \sum_{i=1}^dp_i\ln\left(np_i\cdot \underset{x\sim \text{Mult}(n, p)}{\mathbb{E}}\Big[ \frac{1}{x_i + 1} \Big] \right)+ \ln\left(1 + \frac{d}{n}\right)\label{eqn:use_concavity_non_dp_min_max}\\
        \leq & \sum_{i=1}^d\ln\left(\frac{np_i}{\underset{x\sim \text{Mult}(n, p)}{\mathbb{E}}\left[x_i\right]}\right) + \ln\left(1 + \frac{d}{n}\right)\label{eqn:use_inverse_bernoulli_nondp_min_max}\\
        = &\ln\left(1 + \frac{d}{n}\right)
    \end{align}
    where \eqref{eqn:use_concavity_non_dp_min_max} is by concavity of the function $\ln(x)$, and \eqref{eqn:use_inverse_bernoulli_nondp_min_max} is by \cref{lem:inverse_sum_bernoulli}.
\end{proof}

\begin{lemma}
    \label{lem:inverse_sum_bernoulli}
        Let $s_1, \cdots, s_K$ be $K$ independent Bernoulli random variables with parameter $p_1, \cdots, p_K$. Let $c\geq 1$ be a positive constant. Then we have that
        \begin{align}
            \mathbb{E}\left[\frac{1}{c + \sum_{k=1}^K s_k}\right] \leq \frac{1}{\sum_{k=1}^Kp_k}
        \end{align}
    \end{lemma}
    \begin{proof}
        Conditioned on any fixed value for $s_3, \cdots, s_K$, denote $c' = c + \sum_{k=3}^Ks_k$ we have that
        \begin{align}
            \mathbb{E}&\left[\frac{1}{c + \sum_{k=1}^K s_k}\Big|s_3, \cdots, s_K\right] = \frac{p_1p_2}{c' + 2} + \frac{p_1(1-p_2) + (1-p_1)p_2}{c' + 1} + \frac{(1-p_1)(1-p_2)}{c'}\\
            = & p_1p_2\left(\frac{1}{c'+2} - \frac{2}{c' + 1} + \frac{1}{c'}\right)  + (p_1 + p_2)\cdot \left(\frac{1}{c' + 1} - \frac{1}{c'}\right) + \frac{1}{c'}\\
            \leq & \left(\frac{p_1 + p_2}{2}\right)^2\left(\frac{1}{c'+2} - \frac{2}{c' + 1} + \frac{1}{c'}\right)  + (p_1 + p_2)\cdot \left(\frac{1}{c' + 1} - \frac{1}{c'}\right) + \frac{1}{c'}
        \end{align}
        where the last inequality is by $\frac{1}{c'+2} - \frac{2}{c' + 1} + \frac{1}{c'}>0$ for $c'>0$. By similar argument, we have that
        \begin{align}
            \mathbb{E}&\left[\frac{1}{c + \sum_{k=1}^K s_k} \right] \leq \mathbb{E}\left[\frac{1}{c + s}\right] \leq \mathbb{E}\left[\frac{1}{1 + s}\right]
        \end{align}
        where $s \sim \text{Bin}(K, \Bar{p})$ where $\bar{p} = \frac{1}{K}\sum_{k=1}^Kp_k$. Then by simple algebraic tricks, we have that
        \begin{align}
            \mathbb{E}\left[\frac{1}{c + \sum_{k=1}^K s_k} \right] \leq & \sum_{j=0}^{K}\binom{K}{j} \bar{p}^{j}(1-\bar{p})^{K-j} \cdot \frac{1}{j+1}\\
            = & \frac{1}{(K + 1)\bar{p}}\sum_{j=0}^{K}\binom{K+1}{j+1} \bar{p}^{j + 1}(1-\bar{p})^{K-j} \\
            < & \frac{1}{K\bar{p}} = \frac{1}{\sum_{k=1}^K p_k}
        \end{align}
    \end{proof}

\subsection{Our DP Minimax Results}

\label{ssec:dp_minmax}
We first summarize our derived DP minimax rates in \cref{tab:dp_kl_hist_min_max} -- observe that  simple variants of Laplace mechanism is minimax optimal. 

\begin{table*}[t]
    \centering
    \begin{tabular}{c|c|c|c}
    \toprule
        Method & KL Error Bound & Conditions & Reference \\\midrule
        Any $(\varepsilon,\delta)$-DP & $\Omega\left(\ln(1 + \frac{d}{n\min\{\varepsilon, 1\}})\right)$ &  $d\geq 4, \delta\leq \varepsilon$ & \cref{thm:lower_dp_min_max}\\\hline
        Add-constant ($(\varepsilon, \delta)$-DP)  & $O\left(\ln\left(1 + \frac{d}{n\min\{\varepsilon, 1\}}\right)\right)$  & $\varepsilon>0$, $\delta\geq 0$ & \cref{thm:dp_min_max_upper_pure}\\
        (Laplace Mechanism) & & &  \\
     \bottomrule
    \end{tabular}
    \caption{DP Minimax KL error bounds}
    \label{tab:dp_kl_hist_min_max}
\end{table*}

We comment that our KL minimax lower bound is stronger than naive conversions of prior TV distance lower bounds to KL lower bounds -- applying Pinsker inequality to the prior $\Omega\left(\frac{d}{\varepsilon n}\right)$ TV lower bounds~\cite{diakonikolas2015differentially,acharya2021differentially} only gives a KL lower bound of $\Omega\left(\min\left\{\frac{d^2}{\varepsilon^2n^2}, 1\right\}\right)$, which is significantly weaker than our bound $\Omega\left(\ln\left(1 + \frac{d}{\varepsilon n}\right)\right)$ in \cref{tab:dp_kl_hist_min_max}, especially for large $d$.

Below we present the proofs for the minimax lower bounds and upper bounds.

\subsubsection{DP Minimax Lower Bound}

\begin{theorem}[Minimax Lower Bound for DP estimation]
\label{thm:lower_dp_min_max} Let $d, n\in\mathbb{N}$, $\varepsilon\geq 0$, and $\delta\in[0,1]$ be fixed. If $d\geq 4$, $\delta\leq \varepsilon$, then 
    \begin{align}
        \max_{p\in\Delta(d)}\underset{x\sim \text{Poi}(n, p)}{\mathbb{E}}\Big[KL(p,  \mathcal{A}(x))\Big] \geq \Omega\left(\ln\left(1 + \frac{d}{\varepsilon n}\right)\right)
    \end{align}
\end{theorem}

\begin{proof}
    Let $\kappa, k\in\mathbb{N}$ be defined as follows.
    \begin{align}
        \kappa = \begin{cases}
            2 & \frac{d}{ 160n\varepsilon} \leq 2 \\
            \lfloor\frac{d}{160n\varepsilon} \rfloor & \frac{d}{160n \varepsilon} > 2 \text{ and }80n\varepsilon\geq 1\\
            \lfloor \frac{d}{2} \rfloor & \frac{d}{160n \varepsilon} > 2 \text{ and }80n\varepsilon<1\\
        \end{cases}
        \text{ and }
        k = \begin{cases}
            \lfloor \frac{d}{4} \rfloor & \frac{d}{160n\varepsilon} \leq 2 \\
            \lfloor 80n\varepsilon \rfloor & \frac{d}{160n \varepsilon} > 2 \text{ and }80n\varepsilon\geq 1 \\
            1 &  \frac{d}{160n \varepsilon} > 2 \text{ and }80n\varepsilon< 1
        \end{cases}.
        \label{eqn:def_kappa_k_dp_min_max}
    \end{align}
    Thus by $d\geq 2$, we have 
    \begin{align}
        \kappa\geq 2\quad\text{and}\quad k\geq 1 \quad \text{and} \quad \kappa\cdot k\leq \frac{d}{2} \quad \text{and} \quad \frac{k}{160n\varepsilon} \leq \frac{1}{2}\label{eqn:combined_mass_kappa_k_dp_minmax}
    \end{align}
    For $i=1, \cdots, k$, we construct a packing set of distributions supported on symbols $\mathcal{B}_i = \{k\cdot i - \kappa + 1, \cdots, \delta_{\kappa \cdot i}\}$ as follows.
    \begin{align}
        \text{For }i=1,\cdots, k\quad \mathcal{P}_i = \{\delta_{\kappa\cdot i- \kappa + 1}, \cdots, \delta_{\kappa\cdot i}\}
    \end{align}
    We construct the following set of distributions that lie in the additive neighborhood of $p$.
    \begin{align}
        \mathcal{P} = \left\{q = \left(1-\frac{k}{160n\varepsilon}\right)\cdot q^c + \sum\limits_{i = 1}^{k}w_i\cdot q^i: w_i = \frac{1}{160n\varepsilon}\text{ and } q^i \in \mathcal{P}_i\text{ for any }i=1, \cdots, k\right\}
    \end{align}
    where 
    \begin{align}
        q_c (j) = \begin{cases}
            0 & j = 1, \cdots, \kappa \cdot k\\
            \frac{1}{d - \kappa\cdot k}  & j = \kappa \cdot k + 1, \cdots, d
        \end{cases}
    \end{align}

    One can verify that the distributions in $\mathcal{P}$ are well-defined (i.e., normalized). Now by applying \cref{lem:dirac_packing_lower_bound} to $\mathcal{P}_i$ for each $i=1, \cdots, k$, we prove that for any $q'\in\Delta(d)$, it holds that 
    \begin{align}
        \frac{1}{|\mathcal{P}_i|}\sum_{q\in \mathcal{P}_i} \mathbf{1}_{KL(q, q')\geq f(\mathcal{P}_i)} & \geq \frac{1}{2} \text{ where } f(\mathcal{P}_i) = \ln\left(1 + \frac{\kappa}{4}\right)\label{eqn:min_max_dp_lower_packing_i}
    \end{align}
    Thus the first condition of \cref{thm:thm_conditional_packing} holds. Below we analyze the second condition of \cref{thm:thm_conditional_packing}. For each $i\in [k]$ and fixed $q^j\in\mathcal{P}_j, j=1, \cdots, k$ and fixed $\bar{q}^i\in\mathcal{P}_i$. Let $x\sim\text{Poi}\left(n, \left(1-\frac{k}{160n\varepsilon}\right)\cdot q^c + \sum\limits_{j = 1}^{k}w_j\cdot q^j\right)$ and $y\sim\text{Poi}\left(n, \left(1-\frac{k}{160n\varepsilon}\right)\cdot q^c + w_i\cdot \bar{q}^i + \sum\limits_{j\neq i} w_j\cdot q^j\right)$ be independent Poisson random variables. Then we could construct the following coupling $(x, \bar{x})$ between distributions $\text{Poi}\left(n, \left(1-\frac{k}{160n\varepsilon}\right)\cdot q^c + \sum\limits_{j = 1}^{k}w_j\cdot q^j\right)$ and $\text{Poi}\left(n, \left(1-\frac{k}{160n\varepsilon}\right)\cdot q^c + w_i\cdot \bar{q}^i + \sum\limits_{j\neq i} w_j\cdot q^j\right)$:
    \begin{align}
        \bar{x}_l = \begin{cases}
            y_l & l = \kappa\cdot i - \kappa + 1, \cdots, \kappa \cdot i\\
            x_l & l \in [d]\setminus \{\kappa\cdot i - \kappa + 1, \cdots, \kappa \cdot i\}
        \end{cases}
    \end{align}
    By definition, we compute that
    \begin{align}
        \mathbb{E}\left[\lVert x - \bar{x}\rVert_1\right] = &\sum_{l = \kappa\cdot i - \kappa + 1}^{\kappa \cdot i}\mathbb{E}\left[|x_l - y_l|\right]\leq \sum_{l = \kappa\cdot i - \kappa + 1}^{\kappa \cdot i}\mathbb{E}\left[x_l + y_l\right] =  \frac{1}{80\varepsilon} \coloneqq \tau \label{eqn:min_max_dp_lower_tau}
    \end{align}
    where the inequality is by triangle inequality for $\ell_1$ distance. By applying \cref{thm:thm_conditional_packing} under our proved conditions \eqref{eqn:min_max_dp_lower_packing_i} and \eqref{eqn:min_max_dp_lower_tau}, we finally prove that 
    \begin{align}
        & \max_{q\in\mathcal{P}} \mathbb{E}\left[KL(q, \mathcal{A}(x))\right] 
        \geq \sum_{i=1}^k w_i \cdot \left(\frac{1}{10} - 4 \varepsilon\cdot \tau \right) \cdot f(\mathcal{P}_i) =  \frac{k}{160n\varepsilon} \cdot \frac{1}{20} \cdot \ln\left(1 + \frac{\kappa}{4}\right)\\
        = & \begin{cases}
            \frac{1}{160\cdot 20n\varepsilon}\lfloor\frac{d}{4}\rfloor\cdot \ln\left(1 + \frac{1}{2}\right) & \frac{d}{160n\varepsilon} \leq 2\\
            \frac{1}{160\cdot 20n\varepsilon}\cdot\lfloor 80n \varepsilon \rfloor\cdot \ln\left(1 + \frac{1}{4}\lfloor\frac{d}{160\cdot n \varepsilon} \rfloor\right) & \frac{d}{160 n\varepsilon } > 2\text{ and }80n\varepsilon\geq 1\\
            \frac{1}{160\cdot 20 n\varepsilon} \cdot \ln\left(1 + \frac{1}{4}\lfloor \frac{d}{2}\rfloor\right) & \frac{d}{160 n\varepsilon } > 2\text{ and }80n\varepsilon< 1\\
        \end{cases}  \label{eqn:use_def_kappa_k_dp_minmax} \\
        \geq & \begin{cases}
            \frac{d}{160\cdot 20n\varepsilon \cdot 8} \ln\left(1 + \frac{1}{2}\right) & \frac{d}{160n\varepsilon} \leq 2\\
            \frac{1}{80}\cdot\lfloor 80n \varepsilon \rfloor\cdot \ln\left(1 +   \frac{d}{8\cdot 160\cdot n \varepsilon}  \right) & \frac{d}{160 n\varepsilon } > 2\text{ and }80n\varepsilon\geq 1\\
            \frac{1}{160\cdot 20 n\varepsilon} \cdot \ln\left(1 + \frac{d}{16}\right) & \frac{d}{160 n\varepsilon } > 2\text{ and }80n\varepsilon< 1\\
        \end{cases}\label{eqn:use_floor_half_dp_minmax}\\
        \geq & \Omega\left(\ln\left(1 + \frac{d}{n\varepsilon}\right)\right)\label{eqn:dp_min_max_lower_stat_second_to_last_inequality}
    \end{align}
    where \eqref{eqn:use_def_kappa_k_dp_minmax} is by using the definitions of $\kappa$ and $k$ in \eqref{eqn:def_kappa_k_dp_min_max}, \eqref{eqn:use_floor_half_dp_minmax} is by $\lfloor x\rfloor\geq \frac{x}{2}$ for any $x\geq 1$, \eqref{eqn:dp_min_max_lower_stat_second_to_last_inequality} is by $\ln(1 + \lambda x)\leq \lambda\cdot \ln\left(1 + x\right)\leq x$ for $0\leq\lambda\leq 1$ and $x>0$, and by $\lambda\ln(1 + x)\geq \ln(1 + \lambda x)$ for $\lambda>1$ and $x>0$.
\end{proof}

\subsubsection{DP Minimax Upper Bound}

\begin{theorem}[Upper Bound - Pure DP]
\label{thm:dp_min_max_upper_pure}
   There exists an $(\varepsilon, \delta)$-DP estimator $\mathcal{A}$ such that for any fixed $d, n\in\mathbb{N}$, 
   \begin{align}
       \underset{x\sim \text{Poi}(n, p)}{\mathbb{E}}\Big[KL(p\lVert \mathcal{A}(x))\Big] \leq  O\left(\ln\left(1 + \frac{d}{n\min\{1, \varepsilon\}}\right)\right)
   \end{align}
\end{theorem}
\begin{proof}
    This can be achieved by a simple estimator that combines an add-constant estimator and Laplace mechanism as follows.
   \begin{align}
       \mathcal{A}(x)_i = \frac{1}{N} \cdot \frac{\tilde{x}_i}{n} \text{ where }\tilde{x}_i = \max\left\{x_i + z_i, \frac{1}{\min\{1, \varepsilon\}}\right\}
   \end{align}
   where $z\sim Lap\left(0, \frac{1}{\varepsilon}\right)^d$, and $N=\sum_i\tilde{x}_i$ is the normalization constant. The $(\varepsilon, 0)$-DP guarantee follows by observing that the $\ell_1$-sensitivity of vector release $(x_1, \cdots, x_d)$ is one, and by applying the DP guarantee for Laplace Mechanism in~\citep[Theorem 1]{dwork2006calibrating}. Below we focus on bounding the KL error. By concavity of $\ln(t)$ on $t>0$, we have
   \begin{align}
        \underset{x\sim\text{Poi}(n, p)}{\mathbb{E}}[KL(p\lVert \mathcal{A}(x))] \leq &  \underset{x\sim\text{Poi}(n, p)}{\mathbb{E}}\left[\sum_i p_i\ln  \left(\frac{p_i\cdot n}{\tilde{x}_i} \right)\right] + \ln\left(1 + \sum_{i}\frac{\mathbb{E}\left[\tilde{x}_i - p_i\right]}{n}\right)\nonumber\\
        \leq & \underset{x\sim\text{Poi}(n, p)}{\mathbb{E}}\left[\sum_{i:p_i>\frac{1}{n\min\{1, \varepsilon\}}} p_i\ln  \left(\frac{p_i\cdot n}{\tilde{x}_i} \right) + \frac{\tilde{x}_i - p_i}{n}\right] + \ln\left(1 + \sum_{i: p_i<\frac{1}{n\min\{1, \varepsilon\}}}\frac{\mathbb{E}\left[\tilde{x}_i - p_i\right]}{n}\right)\label{eqn:pure_dp_min_max_use_lemma}\\
        \leq &  O\left( \frac{\sum\limits_{i:p_i>\frac{1}{n\min\{1, \varepsilon\}}}1}{n\min\{1, \varepsilon\}}\right) + \ln\left(1 + O\left(\frac{\sum\limits_{i:p_i<\frac{1}{n\min\{1, \varepsilon\}}} 1}{n\min\{1, \varepsilon\}}\right)\right) \label{eqn:pure_dp_min_max_second_to_last}\\
        \leq & O\left(1 + \frac{d}{n\min\{1, \varepsilon\}}\right) \label{eqn:pure_dp_min_max_last}
    \end{align}
    where \eqref{eqn:pure_dp_min_max_use_lemma} is by $p_i\ln\left(\frac{p_i\cdot n}{\tilde{x}_i}\right)\leq 0$ for $p_i<\frac{1}{n\min\{1, \varepsilon\}}$, by $ \mathbb{E}[\tilde{x}_i - x_i]\geq 0$ for any $i$ (by \cref{lem:bias_threshold_laplace}), and by $\ln(1 + x + y)\leq x + \ln(1 + y)$ for any $x, y>0$; \eqref{eqn:pure_dp_min_max_second_to_last} is by applying \cref{lem:base_lemma_kl} under setting $m=n$, $c_1=-\infty$, $c_2=+\infty$, $b=\frac{1}{\varepsilon}$ and $c = \frac{1}{\min\{1, \varepsilon\}}$ for $p_i>\frac{1}{n\min\{1, \varepsilon\}}$ and by applying \cref{lem:bias_threshold_laplace} for $p_i<\frac{1}{n\min\{1, \varepsilon\}}$; and \eqref{eqn:pure_dp_min_max_last} is by $a + \ln(1 + b)\leq \ln(1 + 2a + 2b)$ for $0<a<1$ and $b>0$.
\end{proof}

\section{Deferred Proofs for Non-DP Per-Instance Lower Bounds}

\label{app:nondp_per_instance_lower}

For ease of presentation and understanding, we break up the neighborhood $N_{+}(p)$ into two sub-neighborhoods: one $N_{\leq \frac{t}{n}}\subseteq N_{+}(p)$ with the same small perturbation for every symbol, and the other $N_{stat}\subseteq N_{+}(p)$ (based on statistical variance) with larger perturbations for large symbols, defined as follows.
\begin{align}
    &N_{\leq \frac{t}{n}}\left(p\right) = \Bigg\{q: |q_i - p_i|\leq \frac{t}{n}\text{ for any }i\in[d] \text{ and }\sum\limits_{i: p_i\leq \frac{t}{n}}q_i\leq \max\left\{\frac{t}{n}, \sum\limits_{i: p_i\leq \frac{t}{n}}p_i\right\}\Bigg\}  \label{eqn:add_neighborhood_nondp}\\
    & N_{stat}(p) = \Bigg\{q: |q_i-p_i| \leq \min\left\{p_i, \sqrt{\frac{p_i}{n}}\right\} \text{ for any }i\in[d]\Bigg\} \cup \left\{q: \sum_{i=1}^d|q_i - p_i|\leq \frac{1}{n}\right\}  \label{eqn:stat_neighborhood}
\end{align}
We then prove per-instance lower bounds under the sub-neighborhoods $N_{\leq \frac{t}{n}}\left(p\right)$ and $N_{stat}(p)$ respectively. By definition \eqref{eqn:instance_optimality_objective}, their average is a lower bound for the per-instance lower bound under $N_{+}(p)$, i.e., $\text{lower}(p, n, N_{+})\geq \frac{1}{2} \cdot \text{lower}(p, n, N_{\leq \frac{t}{n}} ) + \frac{1}{2} \cdot \text{lower}(p, n, N_{stat})$. (This is because one can construct a distribution over hard instances, choosing the hard instance(s) in $N_{\leq \frac{t}{n}}\left(p\right)$ and $N_{stat}(p)$ with $1/2$ probability respectively.) Our per-instance lower bounds under the two sub-neighborhood are summarized in \cref{tab:nondp_kl_hist_instance_lower}.

\begin{table*}[t]
    \centering
    {\small\begin{tabular}{c|c|c|c}
    \toprule
        Neighborhood $N$ & $\text{lower}(p, n, N)$ \eqref{eqn:instance_optimality_objective} & Conditions & Reference \\\midrule
        $N_{\leq t/n}(p)$ \eqref{eqn:add_neighborhood_nondp} &  $\Omega\Big(\frac{\ln(1 + d_{small}(L'))}{n} + p_{small}(L')  \ln\left(1 + \frac{d_{small}(L')}{np_{small}(L')}\right) \Big)$ & $L'\subseteq[d]$, $t\geq 1$  & \cref{thm:non_dp_lower_per_instance_additive_small_symbols}\\
        & & $d\geq 2$, $n\geq 4$ & \\\hline
        $N_{stat}(p)$ \eqref{eqn:stat_neighborhood} &  $\Omega\Big(\sum\limits_{i}\min\left\{p_i, \frac{1}{n}\right\}\Big)$ & $d\geq 2$ & \cref{thm:non_dp_lower_per_instance_additive}\\
     \bottomrule
    \end{tabular}}
    \caption{Non-DP per-instance lower bounds, where $p_{small}(L') = \sum\limits_{i\in L', p_i\leq \frac{t}{n}}p_i $ and $d_{small}(L') = \sum\limits_{i\in L', p_i\leq \frac{t}{n}} 1$.}
    \label{tab:nondp_kl_hist_instance_lower}
\end{table*}

\subsection{Useful Lemmas}

\begin{lemma}[{Coupling Lemma~\citep[Lemma 3.6]{aldous1983random}}] Let random variables $Z_1, Z_2$ have distributions $\nu_1, \nu_2$. Then
    \begin{align}
        \text{TV}\left(\nu_1, \nu_2\right)\leq \Pr\left[Z_1\neq Z_2\right].
    \end{align}
    Conversely, given probability distributions $\nu_1, \nu_2$, there exists $(Z_1, Z_2)$ such that
    \begin{align}
        \text{TV}\left(\nu_1, \nu_2\right)= \Pr\left[Z_1\neq Z_2\right]
    \end{align}
    where $Z_1, Z_2$ have distributions $\nu_1, \nu_2$ respectively.
    \label{lem:coupling_lemma}
\end{lemma}

\begin{lemma}[Total Variation Inequality between Product Measures]
    \label{lem:tv_product_measures}
    Let $\mu_1, \mu_2$ be probability distributions over $\Omega$ and let $\mu_1', \mu_2'$ be probability distributions over $\Omega'$. Denote $\mu_1\times \mu_1'$ as the product measure of $\mu_1$ and $\mu_1'$ over $\Omega\times \Omega'$, and similarly denote $\mu_2\times \mu_2'$ as the product measure of $\mu_2$ and $\mu_2'$ over $\Omega\times \Omega'$. Then 
    \begin{align}
        \text{TV}(\mu_1\times \mu_1', \mu_2\times \mu_2')\leq \text{TV}(\mu_1, \mu_2) + \text{TV}(\mu_1', \mu_2')
    \end{align}
\end{lemma}
\begin{proof}
    By the coupling lemma \cref{lem:coupling_lemma}, there exists $(Z_1, Z_1')$ such that $Z_1\sim \mu_1$, $Z_1'\sim \mu_1'$ and $\text{TV} = \Pr[Z_1\neq Z_1']$. 
    Similarly, there exists $(Z_2, Z_2')$ such that $Z_2\sim \mu_2$, $Z_2'\sim \mu_2'$ and $\text{TV} = \Pr[Z_2\neq Z_2']$. Let $Z = (Z_1, Z_1')$ and $Z' = (Z_2, Z_2')$. Then $Z\sim \mu_1\times \mu_1'$ and $Z'\sim \mu_2\times \mu_2'$. By again using the coupling lemma \cref{lem:coupling_lemma}, we prove that 
    \begin{align}
        \text{TV}(\mu_1\times \mu_1', \mu_2\times \mu_2')\leq \Pr[Z\neq Z'] \leq \Pr[Z_1\neq Z_1'] + \Pr[Z_2\neq Z_2'] = \text{TV}(\mu_1, \mu_2) + \text{TV}(\mu_1', \mu_2')
    \end{align}
    where the second inequality is by union bound.
\end{proof}

\subsection{Non-DP Per-Instance Lower Bound under $N_{\leq \frac{t}{n}}\left(p\right)$}

We now prove the per-instance lower bound in \cref{tab:nondp_kl_hist_instance_lower} under additive neighborhood $N_{\leq \frac{t}{n}}\left(p\right)$ for low-probability symbols.  

\begin{theorem}\label{thm:non_dp_lower_per_instance_additive_small_symbols} 
    Let $p\in\Delta(d)$ be fixed. For $t>0$, let $N_{\leq \frac{t}{n}}\left(p\right)$ be the below additive local neighborhood of $p$ as defined in \eqref{eqn:add_neighborhood_nondp}.
    \begin{align}
        N_{\leq \frac{t}{n}}\left(p\right) = \left\{q: |q_i - p_i|\leq \frac{t}{n}  \text{ and } \sum\limits_{i: p_i\leq \frac{t}{n}}q_i\leq \max\left\{\frac{t}{n}, \sum\limits_{i: p_i\leq \frac{t}{n}}p_i\right\}\right\}\label{eqn:add_neighborhood_nondp_app}
    \end{align}
    If $t\geq 1$, $d\geq 2$ and $n\geq 4$,  then for any $L'\subseteq [d]$, 
    \begin{align}
        \max_{q\in  N_{\leq \frac{t}{n}}\left(p\right)}\underset{x\sim \text{Poi}(n, q)}{\mathbb{E}}KL(q, \mathcal{A}(x)) \geq  \Omega\left( \frac{\ln(1 + d_{small}(L'))}{n} + p_{small}(L')\ln\left(1 + \frac{d_{small}(L')}{np_{small}(L') }\right)\right)
        \label{eqn:per_instance_non_dp_lower_additive_small_symbols}
    \end{align}
    where $p_{small}(L') = \sum\limits_{i\in L', p_i\leq \frac{t}{n}}p_i$ and $d_{small}(L') = \sum\limits_{i\in L', p_i\leq \frac{t}{n}} 1$.
\end{theorem}

\begin{proof}
    We separate the discussions for different $d_{small}(L')$.
    \begin{enumerate}
        \item If $d_{small}(L')=0$, then  $p_{small}(L') = 0$ and thus \eqref{eqn:per_instance_non_dp_lower_additive_small_symbols} trivially holds.
        \item If $1\leq d_{small}(L') \leq 3$: Without loss of generality, assume that $\{i\in L': p_i\leq \frac{t}{n}\}= \{1, \cdots, d_{small}(L')\}$. 
        \begin{enumerate}
            \item If $\max_{i\in [d]\setminus[d_{small}(L')]}p_i\leq \frac{1}{n}$: Then the neighborhood $N_{stat}$ defined in \eqref{eqn:stat_neighborhood} is a subset of $N_{\leq 
            \frac{t}{n}}$ defined in \eqref{eqn:add_neighborhood_nondp_app}, i.e., $N_{stat}\subseteq N_{\leq 
            \frac{t}{n}}$. Thus \eqref{eqn:per_instance_non_dp_lower_additive_small_symbols} holds by observing that \eqref{eqn:per_instance_non_dp_lower_additive_small_symbols} is dominated by the lower bound in \cref{thm:non_dp_lower_per_instance_additive}. Specifically, by $\ln(x)\leq x - 1$ for any $x>0$ and by $d_{small}(L') \leq 3$, we prove that $\frac{\ln(1 + d_{small}(L'))}{n} + p_{small}(L')\ln\left(1 + \frac{d_{small}(L')}{np_{small}(L') }\right) \leq \frac{2d_{small}(L')}{n} \leq \frac{6}{n} \leq O\left(\sum_{i=1}^d\min\left\{p_i, \frac{1}{n}\right\}\right)$ where the last inequality is by $d\geq 2$.
            
            \item If $\max_{i\in [d]\setminus[d_{small}(L')]}p_i> \frac{1}{n}$, without loss of generality, assume that $p_d>\frac{1}{n}$: Then we construct a packing set of distributions $\mathcal{P} = \{p^+, p^-\}$ that contains the following two distributions.
            \begin{align}
                p^+(j) = \begin{cases}
                    \frac{1}{2n} & j=1\\
                    p_j + \frac{p_1}{d-1} & j =2, \cdots, d-1\\
                    p_d - \frac{1}{2n} + \frac{p_1}{d-1} & j = d
                \end{cases}
            \end{align}
            and
            \begin{align}
                p^-(j) = \begin{cases}
                    \frac{1}{4n} & j=1\\
                    p_j + \frac{p_1}{d-1} & j =2, \cdots, d-1\\
                    p_d - \frac{1}{4n} + \frac{p_1}{d-1} & j = d
                \end{cases}
            \end{align}
            One can validate that the distributions in $\mathcal{P}$ are well-defined and are in the additive neighborhood $N_{\leq \frac{t}{n}}\left(p\right)$ (by observing that $p_1\leq \frac{t}{n}$ and $d\geq 2$, and thus $\frac{p_1}{d-1}\leq \frac{t}{n}$). By applying \cref{lem:two_symbol_packing_lower_bound} (under setting $a=\frac{1}{2n}$ and $\Delta = \frac{1}{4n}$), we further prove that for any $q'\in \Delta(d)$, 
            \begin{align}
                \frac{1}{|\mathcal{P}|}\sum_{q\in \mathcal{P}} \mathbf{1}_{KL(q, q')\geq f(\mathcal{P})} & \geq \frac{1}{2} \text{ where } f(\mathcal{P}) = \frac{\Delta^2}{a} = \frac{1}{8n}\label{eqn:per_instance_non_dp_lower_packing_i_small}
            \end{align}
            Thus the first condition of \cref{thm:thm_conditional_packing_nondp} holds. Below we analyze the second condition of \cref{thm:thm_conditional_packing_nondp}. By definition of $\mathcal{P}$ we compute that
            \begin{align}
                & \int \left(\min_{q\in \mathcal{P}}\frac{d\text{Poi}\left(n, q\right)}{d\text{Poi}\left(n, p^+\right)}\right)  d\text{Poi}\left(n, p^+\right) = \int\min\left\{d\text{Poi}\left(n, p^+\right), d\text{Poi}\left(n, p^-\right)\right\}\label{eqn:use_two_packing_small}\\
                = & 1 - \text{TV}\left(\text{Poi}\left(n, p^+\right), \text{Poi}\left(n, p^-\right)\right)\label{eqn:nondp_lower_use_tv_definition_small}\\
                \geq &1 - \text{TV}\left(\text{Poi}\left(n\cdot \frac{1}{2n}\right), \text{Poi}\left(n\cdot \frac{1}{4n}\right)\right) \nonumber\\
                & - \text{TV}\left(\text{Poi}\left(np_d - \frac{1}{2} + \frac{np_1}{d-1}\right), \text{Poi}\left(np_d - \frac{1}{4} + \frac{np_1}{d-1}\right)\right)\label{eqn:nondp_lower_use_independence_small}\\
                \geq & 1 - \frac{1}{2}\sqrt{\frac{n^2\left(\frac{1}{4n}\right)^2}{n\cdot \frac{1}{4n}}} - \frac{1}{2}\sqrt{\frac{ \left(\frac{1}{4}\right)^2}{ np_d - \frac{1}{2} + \frac{p_1}{d-1}}} \label{eqn:use_tv_poisson_small}\\
                \geq & 1 - \frac{1}{4} - \frac{1}{4\sqrt{2}} > \frac{1}{2} \coloneqq \tau \label{eqn:nondp_lower_stat_last_inequality_small}
            \end{align}
            where \eqref{eqn:use_two_packing_small} is by definition for $\mathcal{P}$ that only contains two distributions, \eqref{eqn:nondp_lower_use_tv_definition_small} is by using the definition of total variation distance, \eqref{eqn:nondp_lower_use_independence_small} is by the total variation inequality for product measures (\cref{lem:tv_product_measures}), \eqref{eqn:use_tv_poisson_small} is by \cref{lem:tv_poisson}. Thus the second condition of \cref{thm:thm_conditional_packing_nondp} holds. By applying \cref{thm:thm_conditional_packing_nondp} under our proved conditions \eqref{eqn:per_instance_non_dp_lower_packing_i_small} and \eqref{eqn:nondp_lower_stat_last_inequality_small}, we finally prove that 
            \begin{align}
                &\max_{q\in\mathcal{P}} \mathbb{E}\left[KL(q, \mathcal{A}(x))\right] 
                \geq \frac{1}{2} \cdot \tau \cdot f(\mathcal{P}) =  \frac{1}{32n} \\
                \geq & \Omega\left(\frac{\ln(1 + d_{small}(L'))}{n} + p_{small}(L')\cdot \ln\left(1 + \frac{d_{small}(L')}{np_{small}(L')}\right)\right)\label{eqn:per_instance_non_dp_lower_stat_second_to_last_inequality_small}
            \end{align}
            where \eqref{eqn:per_instance_non_dp_lower_stat_second_to_last_inequality_small} is by observing that $\frac{\ln(1 + d_{small}(L'))}{n} + p_{small}(L')\ln\left(1 + \frac{d_{small}(L')}{np_{small}(L') }\right) \leq \frac{2d_{small}(L')}{n} \leq \frac{6}{n}$ (due to  $\ln(x)\leq x - 1$ for any $x>0$ and the condition that $d_{small}(L') \leq 3$).
        \end{enumerate}
        \item If $d_{small}(L')\geq 4$: Without loss of generality, assume that $\{i\in L': p_i\leq \frac{t}{n}\} = \{1, \cdots, d_{small}(L')\}$. For brevity, denote $\hat{d} = \lfloor \frac{d_{small}(L')}{2}\rfloor\geq 2$ and $\hat{p} = \sum_{i=1}^{\hat{d}}p_i$. Without loss of generality, also assume that $p_1 \geq \cdots \geq p_{d_{small}(L')}$, then 
        \begin{align}
            \hat{d}\geq \frac{d_{small}(L')}{3}\text{ and }\hat{p}\geq \frac{p_{small}(L')}{3}
            \label{eqn:cond_hat_p_hat_d}
        \end{align}
        Let $\kappa, k\in\mathbb{N}$ be defined as follows.
        \begin{align}
            \kappa = \begin{cases}
                2 & \frac{\hat{d}}{ n\hat{p}} \leq 2\\
                 \hat{d} &  \frac{\hat{d}}{ n\hat{p}} > 2 \text{ and }n\hat{p}< 1\\
                \lfloor\frac{\hat{d}}{n\hat{p} } \rfloor & \frac{\hat{d}}{ n\hat{p} } > 2\text{ and }n\hat{p}\geq 1
            \end{cases}\quad k = \begin{cases}
                \lfloor \frac{\hat{d}}{2} \rfloor & \frac{\hat{d}}{ n\hat{p}} \leq 2\\
                1 &  \frac{\hat{d}}{ n\hat{p}} > 2 \text{ and }n\hat{p}< 1\\
                \lfloor n\hat{p} \rfloor & \frac{\hat{d}}{ n\hat{p}} > 2 \text{ and }n\hat{p}\geq 1
            \end{cases}.\label{eqn:def_kappa_k}
        \end{align}
        Thus by definition, we have 
        \begin{align}
            \kappa\geq 2\quad \text{and}\quad k\geq 1 \quad \text{and} \quad \kappa\cdot k\leq \hat{d} \leq \frac{d_{small}(L')}{2} \quad \text{and} \quad \frac{k}{n} \leq \max\left\{\hat{p}, \frac{1}{n}\right\}\label{eqn:combined_mass_kappa_k}
        \end{align}
        For $i=1, \cdots, k$, we construct a packing set of distributions supported on symbols $\mathcal{B}_i = \{k\cdot i - \kappa + 1, \cdots, \delta_{\kappa \cdot i}\}$ as follows.
        \begin{align}
            \text{For }i=1,\cdots, k\quad \mathcal{P}_i = \{\delta_{\kappa\cdot i- \kappa + 1}, \cdots, \delta_{\kappa\cdot i}\}
        \end{align}
        We construct the following set of distributions that lie in the additive neighborhood of $p$.
        \begin{align}
            \mathcal{P} = \left\{q = \left(1-\frac{k}{n}\right)\cdot q^c + \sum\limits_{i = 1}^{k}w_i\cdot q^i: w_i = \frac{1}{n}\text{ and } q^i \in \mathcal{P}_i\text{ for any }i=1, \cdots, k\right\} \label{eqn:prior_per_instance_additive_small_prob}
        \end{align}
        where 
        \begin{align}
            q_c (j) = \begin{cases}
                0 & j = 1, \cdots, \hat{d}\\
                \frac{1}{1 - \frac{k}{n}} \cdot \left(p_j + \frac{\max\left\{\hat{p}, \frac{1}{n}\right\} - \frac{k}{n}}{d_{small}(L') - \hat{d}} \right)  & j = \hat{d} + 1, \cdots, d_{small}(L')\\
                \frac{p_j}{1 - \frac{k}{n}} \cdot \left(1 + \frac{\hat{p} - \max\left\{\hat{p}, \frac{1}{n}\right\}}{1 - p_{small}(L')}\right) & j = d_{small}(L') + 1, \cdots, d
            \end{cases}
        \end{align}
    
        One can verify that the distributions in $\mathcal{P}$ are well-defined (i.e., normalized) and lie in the neighborhood $N_{\leq \frac{t}{n}}(p)$ defined in \eqref{eqn:add_neighborhood_nondp}. This is by $p_j \leq \frac{t}{n}$ for $j=1, \cdots, \hat{d}$, and by $0\leq \frac{\max\left\{\hat{p}, \frac{1}{n}\right\} - \frac{k}{n}}{d_{small}(L') - \hat{d}} \leq \max\left\{\frac{\hat{p}}{\hat{d}}, \frac{1}{n}\right\} \leq \frac{t}{n}$ under \eqref{eqn:combined_mass_kappa_k} and $\hat{d}<d_{small}(L')$, and by $0 \geq \frac{\hat{p} - \max\left\{\hat{p}, \frac{1}{n}\right\}}{1 - p_{small}(L')}\geq - \frac{1/n}{1 - 3/n} \geq - 1$ under \eqref{eqn:cond_hat_p_hat_d} and $n\geq 4$, and by observing that $0\geq p_j\cdot \frac{\hat{p} - \max\left\{\hat{p}, \frac{1}{n}\right\}}{1 - p_{small}(L')}\geq \hat{p} - \max\left\{\hat{p}, \frac{1}{n}\right\}\geq -\frac{1}{n}$ for any $j=d_{small}(L')+1, \cdots, d$.
    
        Now by applying \cref{lem:dirac_packing_lower_bound} to $\mathcal{P}_i$ for each $i=1, \cdots, k$, we prove that for any $q'\in\Delta(d)$, it holds that
        \begin{align}
            \frac{1}{|\mathcal{P}_i|}\sum_{q\in \mathcal{P}_i} \mathbf{1}_{KL(q, q')\geq f(\mathcal{P}_i)} & \geq \frac{1}{2} \text{ where } f(\mathcal{P}_i) = \ln\left(1 + \frac{\kappa}{4}\right)\label{eqn:per_instance_non_dp_lower_packing_i_small_ln}
        \end{align}
        Thus the first condition of \cref{thm:thm_conditional_packing_nondp} holds. Below we analyze the second condition of \cref{thm:thm_conditional_packing_nondp}. By definition of $\mathcal{P}$, for any fixed $i\in[k]$ and any fixed $\bar{p} \coloneqq \left(1 - \frac{k}{n}\right)\cdot q^c + \sum_{i=1}^k\frac{1}{n}\cdot \bar{q}^i\in\mathcal{P}$, we compute that
        \begin{align}
            & \int \left(\min_{q^i\in \mathcal{P}_i}\frac{d\text{Poi}\left(n, \bar{p} - \frac{1}{n}\bar{q}^i + \frac{1}{n}q^i\right)}{d\text{Poi}\left(n, \bar{p}\right)}\right)  d\text{Poi}\left(n, \bar{p}\right) = \int\min_{q^i\in\mathcal{P}_i} d\text{Poi}\left(n, \bar{p} - \frac{1}{n}\bar{q}^i + \frac{1}{n}q^i\right)\\
            \geq & \min_{q^i\in \mathcal{P}_i}\Pr_{x\sim\text{Poi}(n, \bar{p} - \frac{1}{n}\bar{q}^i + \frac{1}{n}q^i)}\left[x_{\kappa\cdot i - \kappa + 1} = \cdots = x_{\kappa\cdot i}\right] =  \frac{1}{e} \coloneqq \tau \label{eqn:nondp_lower_add_ln_last_inequality}
        \end{align}
        where \eqref{eqn:nondp_lower_add_ln_last_inequality} is by probability mass function of Poisson distributions $\text{Poi}\left(n\cdot 0\right)$ and $\text{Poi}\left(n\cdot \frac{1}{e}\right)$. By applying \cref{thm:thm_conditional_packing_nondp} under our proved conditions \eqref{eqn:per_instance_non_dp_lower_packing_i_small_ln} and \eqref{eqn:nondp_lower_add_ln_last_inequality}, we finally prove that 
        \begin{align}
            & \max_{q\in\mathcal{P}} \mathbb{E}\left[KL(q, \mathcal{A}(x))\right] 
            \geq \frac{1}{2} \cdot \tau \cdot \sum_{i=1}^k w_i \cdot f(\mathcal{P}_i) =  \frac{k}{2en}\ln\left(1 + \frac{\kappa}{4}\right) \\
            = & \begin{cases}
                \frac{1}{2en}\lfloor\frac{\hat{d}}{2}\rfloor\cdot \ln\left(1 + \frac{1}{2}\right) & \frac{\hat{d}}{ n\hat{p}} \leq 2\\
                \frac{1}{2en}\cdot \ln\left(1 + \frac{\hat{d}}{4}\right) &  \frac{\hat{d}}{ n\hat{p}} > 2 \text{ and }n\hat{p}< 1\\
                \frac{1}{2en}\cdot\lfloor n\hat{p}\rfloor\cdot \ln\left(1 + \frac{1}{4}\lfloor\frac{\hat{d}}{n\hat{p} } \rfloor\right) & \frac{\hat{d}}{ n\hat{p} } > 4\text{ and }n\hat{p}\geq 1
            \end{cases} \label{eqn:use_def_kappa_k}\\
            \geq & \begin{cases}
                \Omega\left(\frac{\hat{d}}{n}\right) & \frac{\hat{d}}{ n\hat{p}} \leq 4\\
                \Omega\left(\frac{1}{n}\cdot \ln\left(1 +  \hat{d}\right)\right) &  \frac{\hat{d}}{ n\hat{p}} > 4 \text{ and }n\hat{p}< 1\\
                \Omega\left( \hat{p} \cdot \ln\left(1 +  \frac{\hat{d}}{n\hat{p} }\right)\right) & \frac{\hat{d}}{ n\hat{p} } > 4\text{ and }n\hat{p}\geq 1
            \end{cases}\label{eqn:use_floor_half}\\
            \geq & \Omega\left(\frac{\ln(1 + \hat{d}}{n} + \hat{p}\cdot \ln\left(1 + \frac{\hat{d}}{n\hat{p}}\right)\right)\label{eqn:per_instance_non_dp_lower_stat_second_to_last_inequality_small_ln}\\
            \geq & \Omega\left(\frac{\ln(1 + d_{small}(L'))}{n} + p_{small}(L')\cdot \ln\left(1 + \frac{d_{small}(L')}{n p_{small}(L')}\right)\right)
        \end{align}
        where \eqref{eqn:use_def_kappa_k} is by using the definitions of $\kappa$ and $k$ in \eqref{eqn:def_kappa_k}, \eqref{eqn:use_floor_half} is by $\lfloor x\rfloor\geq \frac{x}{2}$ for any $x\geq 1$, \eqref{eqn:per_instance_non_dp_lower_stat_second_to_last_inequality_small_ln} is by $\lambda\ln(1 + x)\geq  \ln\left(1 + \lambda x\right)$ for $\lambda\geq 1$ and $x>0$, and by $\ln(1 + \lambda x)\leq \lambda\cdot \ln\left(1 + x\right)\leq x$ for $0\leq\lambda\leq 1$ and $x>0$, and the last inequality is by applying \eqref{eqn:cond_hat_p_hat_d}.
    \end{enumerate}
\end{proof}

\subsection{Non-DP Per-Instance Lower Bound under  $N_{stat}(p)$}

We now prove the per-instance lower bound in \cref{tab:nondp_kl_hist_instance_lower} under neighborhood $N_{stat}(p)$ \eqref{eqn:stat_neighborhood}.

\begin{theorem} 
\label{thm:non_dp_lower_per_instance_additive}     
Let $n, d\in\mathbb{N}$ and $p\in\Delta(d)$ be fixed. Let $N_{stat}(p)$ be the following additive local neighborhood of $p$ as defined in \eqref{eqn:stat_neighborhood}.
\begin{align}
    N_{stat}(p) = \left\{q: |q_i-p_i| \leq \min\left\{p_i, \sqrt{\frac{p_i}{n}}\right\}\text{ for any }i\in[d]\right\} \cup \left\{q: \sum_{i=1}^d|q_i - p_i|\leq \frac{1}{n}\right\}  \label{eqn:stat_neighborhood_app}
\end{align}
Then if $d\geq 2$, it holds that 
    \begin{align}
        \max_{q\in  N_{stat}(p)}\underset{x\sim \text{Poi}(n, q)}{\mathbb{E}}KL(q, \mathcal{A}(x)) \geq  \Omega\left(\sum\limits_i\min\left\{p_i, \frac{1}{n}\right\}\right)
        \label{eqn:per_instance_non_dp_lower_additive}
    \end{align}
\end{theorem}

\begin{proof}
    We will apply \cref{thm:thm_conditional_packing_nondp} to prove the lower bound. Below we first construct the packing set. Without loss of generality, assume that $p_1\geq \cdots \geq p_d$. For brevity, denote
    \begin{align}
        w_k = p_{2k-1} + p_{2k}\quad\text{and}\quad\Delta_k =  \frac{1}{2} \min\left\{p_{2k}, \sqrt{\frac{p_{2k}}{n}}\right\} \text{ for }k=1, \cdots, \lfloor \frac{d}{2}\rfloor \label{eqn:def_delta_k}
    \end{align}
    Let $\mathcal{P}$ be the following packing set of distributions.
    \begin{align}
        \mathcal{P} = \left\{q\coloneqq \sum_{k=1}^{\lfloor \frac{d}{2}\rfloor}w_k \cdot q^k: q_k\in \mathcal{P}_k \coloneqq\left\{q_+^k, q_-^k\right\}\right\} \label{eqn:packing_nondp_pair}
    \end{align}
    where 
    \begin{align}
        q^k_- (j) = \begin{cases}
            \frac{1}{w_k}\cdot p_{2k-1} & j=2k-1\\
            \frac{1}{w_k}\cdot p_{2k} &  j=2k\\
            0 & j\in[d]\setminus\{2k-1, 2k\}
        \end{cases} \text{and} \quad q^k_+ (j) = \begin{cases}
            \frac{1}{w_k} \cdot \left(p_{2k-1} + \Delta_k\right) & j=2k-1\\
            \frac{1}{w_k} \cdot \left(p_{2k} - \Delta_k\right) & j=2k\\
            0 & j\in[d]\setminus\{2k-1, 2k\}
        \end{cases}\nonumber
    \end{align}
    One can verify that distributions in the packing set $\mathcal{P}$ are normalized and lie in the statistical neighborhood $N_{stat}(p)$ by observing that $\min\left\{p_{i-1}, \sqrt{\frac{p_{i-1}}{n}}\right\}\geq \min\left\{p_{i}, \sqrt{\frac{p_{i}}{n}}\right\}$ for any $i$. By applying \cref{lem:two_symbol_packing_lower_bound} (under setting $a=\frac{p_{2k-1}}{w_k}$ and $\Delta = \frac{\Delta_k}{w_k}$), we further prove that for any $q\in \Delta(d)$, 
    \begin{align}
        \frac{1}{|\mathcal{P}_k|}\sum_{q^k\in \mathcal{P}_k} \mathbf{1}_{KL(q^k, q)\geq f(\mathcal{P}_k)} & \geq \frac{1}{2} \text{ where } f(\mathcal{P}_k) = \frac{\left(\frac{\Delta_k}{w_k}\right)^2}{8 \cdot \frac{p_{2k}}{w_k}} = \frac{1}{32w_k} \cdot \min\left\{p_{2k}, \frac{1}{n}\right\}\label{eqn:per_instance_non_dp_lower_packing_i}
    \end{align}
    Thus the first condition of \cref{thm:thm_conditional_packing_nondp} holds. We now analyze the second condition of \cref{thm:thm_conditional_packing_nondp}. For any $k$ and any fixed $q^l\in \mathcal{P}_l, l\neq k$, by definition of $\mathcal{P}_k$ we compute that 
    \begin{align}
        & \int \left(\min_{q^k\in \mathcal{P}_k}\frac{d\text{Poi}\left(n, \sum_{l=1}^{\lfloor \frac{d}{2}\rfloor} w_l\cdot q^l\right)}{d\text{Poi}\left(n, w_k\cdot q_+^k + \sum_{l\neq k} w_l\cdot q^l\right)}\right)  d\text{Poi}\left(n, w_k\cdot q_+^k + \sum_{l\neq k} w_l\cdot q^l\right) \\
        = & \int\min\left\{d\text{Poi}\left(n, w_k\cdot q_+^k + \sum_{l\neq k} w_l\cdot q^l\right), d\text{Poi}\left(n, w_k\cdot q_-^k + \sum_{l\neq k} w_l\cdot q^l\right)\right\}\label{eqn:use_two_packing}\\
        = & 1 - \text{TV}\left(\text{Poi}\left(n, w_k\cdot q_+^k + \sum_{l\neq k} w_l\cdot q^l\right), \text{Poi}\left(n, w_k\cdot q_-^k + \sum_{l\neq k} w_l\cdot q^l\right)\right)\label{eqn:nondp_lower_use_tv_definition}\\
        \geq &1 - \text{TV}\left(\text{Poi}\left(np_{2k-1}\right), \text{Poi}\left(np_{2k-1} + n\Delta_k\right)\right) - \text{TV}\left(\text{Poi}\left(np_{2k}\right), \text{Poi}\left(np_{2k} - n\Delta_k\right)\right)\label{eqn:nondp_lower_use_independence}\\
        \geq & 1 - \frac{1}{2}\sqrt{\frac{n^2\Delta_k^2}{np_{2k-1}}} - \frac{1}{2}\sqrt{\frac{n^2\Delta_k^2}{np_{2k} - n\Delta_k}} \label{eqn:use_tv_poisson}\\
        \geq & 1 - \sqrt{\frac{\min\{np_{2k}, 1\}}{2}} \geq \frac{1}{4} \coloneqq \tau_k \label{eqn:nondp_lower_stat_last_inequality}
    \end{align}
    
    where \eqref{eqn:use_two_packing} is because $\mathcal{P}_k$ only contains two distributions, \eqref{eqn:nondp_lower_use_tv_definition} is by using the definition of total variation distance, \eqref{eqn:nondp_lower_use_independence} is by the total variation inequality for product measures (\cref{lem:tv_product_measures}), \eqref{eqn:use_tv_poisson} is by \cref{lem:tv_poisson}, \eqref{eqn:nondp_lower_stat_last_inequality} is by $p_{2k-1}\geq p_{2k}$ and by definition \eqref{eqn:def_delta_k} of $\Delta_k = \frac{1}{2}\min\left\{p_{2k}, \sqrt{\frac{p_{2k}}{n}}\right\}\leq \frac{p_{2k}}{2}$. Thus the second condition of \cref{thm:thm_conditional_packing_nondp} holds. By applying \cref{thm:thm_conditional_packing_nondp} under our proved conditions \eqref{eqn:per_instance_non_dp_lower_packing_i} and \eqref{eqn:nondp_lower_stat_last_inequality}, we finally prove that 
    \begin{align}
        \max_{q\in\mathcal{P}} \mathbb{E}\left[KL(q, \mathcal{A}(x))\right] 
        \geq & \frac{1}{2} \sum_{k=1}^{\lfloor \frac{d}{2}\rfloor} w_k \cdot \tau_k \cdot f(\mathcal{P}_k) =  \frac{1}{128}\sum_{k=1}^{\lfloor \frac{d}{2}\rfloor}\min\left\{p_{2k}, \frac{1}{n}\right\}\\
        \geq & \frac{1}{256}\sum_{i=2}^{d}\min\left\{p_{i}, \frac{1}{n}\right\} \label{eqn:per_instance_non_dp_lower_stat_second_to_last_inequality}
    \end{align}
    where \eqref{eqn:per_instance_non_dp_lower_stat_second_to_last_inequality} is by observing that for $p_1\geq \cdots \geq p_{d}$, it is the case that $\sum_{k=1}^{\lfloor \frac{d}{2}\rfloor}\min\left\{p_{2k}, \frac{1}{n}\right\} \geq \sum_{k=1}^{\lfloor \frac{d}{2}\rfloor}\min\left\{p_{\min\{2k + 1, d\}}, \frac{1}{n}\right\}$. We now separate the discussions for the remaining (at most two) symbols.
    \begin{enumerate}
        \item If $\sum_{i=2}^{d}\min\left\{p_{i}, \frac{1}{n}\right\}\geq \frac{1}{2n}$, then \eqref{eqn:per_instance_non_dp_lower_stat_second_to_last_inequality} suffice to prove the bound \eqref{eqn:per_instance_non_dp_lower_additive} in the statement (by observing that there is only one remaining symbol).
        \item If $\sum_{i=2}^{d}\min\left\{p_{i}, \frac{1}{n}\right\}< \frac{1}{2n}$, then it must be the case that $p_1>1-\frac{1}{2n}$, $p_2<\frac{1}{2n}$, and $\sum_{i=1}^d\min\left\{p_i, \frac{1}{n}\right\} < \frac{3}{2n}$. By repeating the proof for a new packing $\mathcal{P}' = \left\{\hat{p}^-, p\right\}$ where 
        \begin{align}
            \hat{p} (j) = \begin{cases}
                p_1 - \frac{1}{2n}  & j=1\\
                p_2 + \frac{1}{2n} & j=2\\
                p_j & j = 3, \cdots, d
            \end{cases}
        \end{align}
        we similarly prove a new lower bound of $\frac{1}{256}\cdot \frac{1}{n}$. One can also validate that the new packing $\mathcal{P}'$ is also in the neighborhood $N_{stat}$ \eqref{eqn:stat_neighborhood} This suffice to prove  \eqref{eqn:per_instance_non_dp_lower_additive} in the statement.
    \end{enumerate} 
\end{proof}
\section{Deferred Proofs for Non-DP Per-Instance Upper Bounds}

\subsection{Non-DP Per-Instance Upper Bound (Sampling Twice Algorithm)}

We will use the following lemma for bounding the estimation error on zero-count symbols.

\begin{lemma}[Error of \cref{alg:nondp_per_instance_upper_alt} on zero-count Symbols]
    Let $p\in\Delta(d)$ be a discrete distribution over $d$ symbols. Let $n\in\mathbb{N}$. Then \cref{alg:nondp_per_instance_upper_alt} satisfies 
    \begin{align}
        \mathbb{E}\left[\sum_{i\in L}p_i\cdot \ln\left(\frac{\frac{p_i}{\sum_{i\in L}p_i}}{\frac{\tilde{x}_i}{\sum_{i\in L}\tilde{x}_i}}\right)\right] \leq &  \; \mathbb{E}\left[\left(\sum_{i\in L}p_i\right)\ln\left(1 + \frac{2\sum\limits_{i\in L} 1}{\sum_{i\in L}np_i}\right)\right] 
        \label{eqn:small_symbol_error}
    \end{align}
    \label{lem:small_symbol_error}
\end{lemma}
\begin{proof}
    By definition, we have that
    \begin{align}
        \mathbb{E}\left[\sum_{i\in L}p_i\cdot \ln\left(\frac{\frac{p_i}{\sum_{i\in L}p_i}}{\frac{\tilde{x}_i}{\sum_{i\in L}\tilde{x}_i}}\right)\right] = & \underbrace{\mathbb{E}\left[\sum_{i\in L}p_i\cdot \ln\left(\frac{np_i/2}{\tilde{x}_i}\right)\right]}_{\circled{1}} + \underbrace{\mathbb{E}\left[\left(\sum_{i\in L}p_i\right) \ln\left(\frac{\sum_{i\in L}\tilde{x}_i}{\sum_{i\in L}np_i/2}\right)\right]}_{\circled{2}}
    \end{align}
    We first analyze \circled{1}. By concavity of $\ln(t)$ over $t>0$ and by $\tilde{x}_i\leq x'_i + 1$, we have that
    \begin{align}
        \circled{1}\leq \mathbb{E}_L\left[\sum_{i\in L}p_i\cdot \ln\left(\frac{np_i }{2} \cdot  \mathbb{E}\left[\frac{1}{x'_i + 1}\right]\right)\right] \leq 0
    \end{align}
    where the last inequality is by \cref{lem:poisson_inverse} under $m=\frac{n}{2}$.
    We then analyze $\circled{2}$. By concavity of $\ln(t)$ over $t>0$, we have that 
    \begin{align}
        & \circled{2} \leq \mathbb{E}_L\left[\sum_{i\in L}p_i\ln\left(1 + \frac{\sum_{i\in L}\mathbb{E}[\tilde{x}_i - np_i/2]}{\sum_{i\in L}np_i/2}\right)\right] \\
        & \leq \mathbb{E}_L\left[\left(\sum_{i\in L}p_i\right)\ln\left(1 + \frac{2\sum\limits_{i\in L} 1}{\sum_{i\in L}np_i}\right)\right]
    \end{align}
    where the last inequality is by applying \cref{lem:bias_threshold_laplace} with $b = 0$ and $c = 1$.
\end{proof}

We are now ready to prove the Non-DP per-instance upper bound for the non-dp ``sampling twice'' algorithm \cref{alg:nondp_per_instance_upper_alt}.

\begin{theorem}[Per-Instance Upper Bound - Sampling Twice Algorithm]
    Let $\mathcal{A}$ be the estimator given by \cref{alg:nondp_per_instance_upper_alt}. If $d\geq 2$, then for any $n$ and any $p\in \Delta(d)$, 
    \begin{align}
        \underset{x\sim \text{Poi}(n, p)}{\mathbb{E}}\Big[KL(p\lVert \mathcal{A}(x))\Big] \leq &  \;O\left(\mathbb{E}\left[\sum_{i\in L}p_i\cdot \ln\left(1 + \frac{\sum_{j\in L}1}{n\sum_{j\in L}p_j}\right)\right] + \sum_i\min\left\{p_i, \frac{1}{n}\right\} \right)
    \end{align}
    \label{thm:non_dp_per_instance_upper_GT_add_constant_simplified_alt}
\end{theorem}
\begin{proof}
    By definition,
    \begin{align}
        & \underset{x, x'\sim\text{Poi}(np)}{\mathbb{E}}[KL(p\lVert \mathcal{A}(x))] \\
        = & \mathbb{E}\left[\sum\limits_{i\in L} p_i\ln\left(\frac{ p_i/\sum_{j\in L}p_j}{ \tilde{x}_i/ \sum_{j\in L}\tilde{x}_j}\right)\right] + \mathbb{E}\left[\left(\sum_{i\in L} p_i\right)\ln\left(\frac{\sum\limits_{i\in L} p_i}{\frac{\tilde{c}}{n/2}}\right) + \sum\limits_{i\notin L} p_i\ln\left(\frac{p_i}{\frac{\tilde{x}_i}{n/2} }\right) + \ln \left(\frac{\tilde{c} + \sum_{i\notin L}\tilde{x}_i}{n/2}\right)\right]\\
        \leq &  \mathbb{E}\left[\sum_{i\in L}p_i\cdot \ln\left(1 + \frac{2\sum_{j\in L}1}{\sum_{j\in L}np_j}\right)\right]  + \mathbb{E}\left[\left(\sum_{i\in L} p_i\right)\ln\left(\frac{\sum\limits_{i\in L} p_i}{\frac{\tilde{c}}{n/2}}\right) +  \frac{\tilde{c} - \sum_{i\in L}np_i/2}{n/2} \right]\nonumber\\
        & + \mathbb{E}\left[  \sum\limits_{i\notin L} p_i\ln\left(\frac{p_i}{\frac{\tilde{x}_i}{n/2} }\right) + \sum_{i\notin L}\frac{\tilde{x}_i - np_i/2}{n/2} \right]\label{eqn:nondp_samp_twice_use_concavity}\\
        \leq & O\left(\mathbb{E}\left[\sum_{i\in L}p_i\cdot \ln\left(1 + \frac{\sum_{j\in L}1}{n\sum_{j\in L}p_j}\right)\right]\right)  + \frac{1}{n/2}  + \mathbb{E}\left[\sum_{i\notin L}\frac{1}{n/2}\right]\label{eqn:nondp_use_lemmas}\\
        \leq & O\left(\mathbb{E}\left[\sum_{i\in L}p_i\cdot \ln\left(1 + \frac{\sum_{j\in L}1}{n\sum_{j\in L}p_j}\right)\right] + \sum_i\min\left\{p_i, \frac{1}{n}\right\}\right)
    \end{align}
    where \eqref{eqn:nondp_samp_twice_use_concavity} is by $\ln(t)\leq t - 1$ for any $t> 0$ and by applying   \cref{lem:small_symbol_error}, \eqref{eqn:nondp_use_lemmas} is by applying \cref{lem:base_lemma_kl} under $m = n/2$, $c_1 = -\infty$, $c_2=+\infty$, $b = 0$ and $c=1$, and the last inequality is by $\Pr[i\notin L] = 1 - e^{-np_i}\leq \min\{np_i, 1\}$, and by $\sum_{i}\min\{p_i, \frac{1}{n}\}\geq \frac{1}{n}$ for $d\geq 2$.
\end{proof}

\subsection{Poof for Matching Lower and Upper Bound}

\begin{corollary}
    Let $\mathcal{A}$ be the estimator given by \cref{alg:nondp_per_instance_upper_alt}. Let $N_{stat}(p)$ and $N_{\leq\frac{t}{n}}(p)$ be the additive neighborhoods defined in \eqref{eqn:stat_neighborhood} and \eqref{eqn:add_neighborhood_nondp} respectively. Then for any $n$ and any $p\in \Delta(d)$, 
    \begin{align}
        \underset{x\sim \text{Poi}(n, p)}{\mathbb{E}}\Big[KL(p\lVert \mathcal{A}(x))\Big] \leq O\left(\text{lower}(p,n, N_{stat}) + \text{lower}(p, n, N_{\leq \frac{t}{n}}) \right)
    \end{align}
    for any choice of neighborhood size $t\geq 1$ such that $t \cdot e^{-t}\leq \frac{1}{\ln d}$. Specifically, we can always choose $t = \min\left\{1, 2\ln \ln d\right\}$.
    \label{cor:nondp_per_instance_upper_additive}
\end{corollary}
\begin{proof}
    We will use the upper bound given by \cref{thm:non_dp_per_instance_upper_GT_add_constant_simplified_alt}. Observe that by $\frac{\sum_{j: x_j=0}1}{\sum_{j: x_j=0}p_j}\leq \frac{\sum_{j: p_j\leq \frac{t}{n}, x_j=0}1}{\sum_{j: p_j\leq \frac{t}{n}, x_j=0}p_j}$, we have 
    \begin{align}
        \mathbb{E}\left[\sum_{i\in L:p_i\leq \frac{t}{n}}p_i\ln\left(1 + \frac{ \sum_{j\in L}1}{n\sum_{j\in L}p_j}\right)\right] & \leq  \mathbb{E}\left[\sum_{i\in L:p_i\leq \frac{t}{n}}p_i\ln\left(1 + \frac{\sum_{j\in L:p_j\leq \frac{t}{n}}1}{n\sum_{j\in L:p_j\leq \frac{t}{n}}p_j}\right)\right]\\
        & \leq\text{lower}(p, n, N_{\leq \frac{t}{n}})\text{ in \cref{thm:non_dp_lower_per_instance_additive_small_symbols}} \label{eqn:nondp_per_instance_choose_threshold_1}
    \end{align}
    Additionally, by definition, we compute that 
    \begin{align}
        \mathbb{E}\left[\sum_{i\in L:p_i>\frac{t}{n}}p_i\ln\left(1 + \frac{ \sum_{j\in L}1}{n\sum_{j\in L}p_j}\right)\right] \leq &  \;\mathbb{E}\left[\sum_{i\in L:p_i> \frac{t}{n}, x_i=0}p_i\ln\left(1 + \frac{d}{t}\right) \right]\\
        = & \sum_{i:p_i>\frac{t}{n}}p_ie^{-np_i}\ln \left(1 + \frac{d}{t}\right)\label{eqn:nondp_per_instance_choose_threshold_2}\\
        \leq & \;O\left(\sum_{i:p_i>\frac{t}{n}}\frac{1}{n}\right) = \text{lower}(p, n, N_{stat})\text{ in \cref{thm:dp_per_instance_lower_additive_low_prob}} \label{eqn:nondp_per_instance_choose_threshold_3}
    \end{align}
    where \eqref{eqn:nondp_per_instance_choose_threshold_2} is by $L=\{i\in[d]: x_i=0\}$ and by probability mass function of Poisson random variables, and \eqref{eqn:nondp_per_instance_choose_threshold_3} is by choosing $t\geq 1$ such that $t \cdot e^{-t}\leq \frac{1}{\ln d}$. Specifically, we can choose $t = \min\left\{1, 2\ln \ln d\right\}$.
\end{proof}

\section{Deferred Proofs for DP Per-Instance Lower Bounds}

\label{app:dp_per_instance_lower}

For ease of presentation and understanding, we break up the neighborhood $N_{\leq \frac{t}{n\varepsilon}}(p)$ for a given small $t\geq 1$ into two sub-neighborhoods: one $N_{\leq \frac{1}{n}}\subseteq N_{\leq \frac{t}{n\varepsilon}}(p)$ with the same small perturbation for every symbol, and the other $N_{\leq \frac{t}{n\varepsilon}}(p)$ with larger perturbations for small symbols, defined as follows.

We then prove per-instance lower bounds under the sub-neighborhoods $N_{\leq \frac{t}{n\varepsilon}}(p)$ and $N_{\leq \frac{t}{n\varepsilon}}(p)$ respectively. By definition \eqref{eqn:instance_optimality_objective}, their average is a lower bound for the per-instance lower bound under $N_{\leq \frac{t}{n\varepsilon}}(p)$ , i.e., $\text{lower}(p, n, N_{\leq \frac{t}{n\varepsilon}})\geq \frac{1}{2} \cdot \text{lower}(p, n, N_{\leq \frac{1}{n\varepsilon}}) + \frac{1}{2} \cdot \text{lower}(p, n, N_{\leq \frac{t}{n\varepsilon}})$. (This is because one can construct a distribution over hard instances, choosing the hard instance(s) in $N_{\leq \frac{t}{n}}\left(p\right)$ and $N_{stat}(p)$ with $1/2$ probability respectively.)

Our results for  DP per-instance lower bounds under neighborhoods $N_{\leq \frac{t}{n\varepsilon}}, t\geq 1$ are summarized in \cref{tab:dp_instance_lower}. 
\begin{table*}[t]
    \centering
    \begin{tabular}{c|c|c|c}
    \toprule
        Neighborhood $N$ & $\text{lower}_{\varepsilon, \delta}\left(p, n, N\right)$~\eqref{eqn:per_instance_lower_dp} & Conditions & Reference \\\midrule
        $N_{\leq \frac{1}{n\varepsilon}}$ \eqref{eqn:add_neighborhood_dp_small_symbols} & $\Omega\left(\sum\limits_{i: p_i <\frac{1}{n\varepsilon}}p_i + \sum\limits_{i:p_i\geq \frac{1}{n\varepsilon}} \frac{1}{p_i} \cdot \frac{1}{n^2\varepsilon^2} \right)$ & $\delta \leq \varepsilon$, $d\geq 2$ & \cref{thm:dp_per_instance_lower_additive} \\\hline
        $N_{\leq \frac{t}{n\varepsilon}}$ \eqref{eqn:add_neighborhood_dp_small_symbols}  &  $\Omega\Bigg(\frac{\ln\left(1 + d_{small}(L')\right)}{n\varepsilon} + $ & $L'\subseteq[d]$, $t\geq 1$, $d\geq 2$ & \cref{thm:dp_per_instance_lower_additive_low_prob}\\  
        & $p_{small}(L') \cdot \ln\left(1 + \frac{d_{small}(L')}{n\varepsilon \cdot  p_{small}(L') }\right)\Bigg)$ & $\delta\leq \varepsilon, n\varepsilon\geq 1$ & \\
     \bottomrule
    \end{tabular}
    \caption{DP per-instance lower bounds, where $p_{small}(L') = \sum\limits_{i\in L': p_i\leq \frac{t}{n\varepsilon}}p_i$ and $d_{small}(L') = \sum\limits_{i\in L': p_i\leq \frac{t}{n\varepsilon}}1$}
    \label{tab:dp_instance_lower}
\end{table*}

\subsection{DP Lower Bound: $N_{\frac{1}{n\varepsilon}}$ neighborhood}

In this section, we prove the per-instance DP estimation lower bound in \cref{tab:dp_instance_lower} under additive neighborhood. We first prove a lower bound under add-$\frac{1}{n\varepsilon}$ neighborhood.
\begin{theorem}[Lower Bound - $N_{\leq \frac{1}{n\varepsilon}}$ Neighborhood]
    Let $d\in\mathbb{N}$, $\varepsilon\geq 0$ and $0\leq \delta\leq 1$, $p\in\Delta(d)$ be fixed. Let $N_{\leq \frac{1}{n\varepsilon}}(p)$ be the additive neighborhood defined in \eqref{eqn:add_neighborhood_dp_small_symbols} for $t=1$ as follows.
    \begin{align}
         N_{\leq \frac{1}{n\varepsilon}}\left(p\right) = \Bigg\{q: |q_i - p_i|\leq \frac{1}{n\varepsilon} \text{ for any }i\in [d]  \text{ and }\sum\limits_{i: p_i\leq \frac{1}{n\varepsilon}}q_i\leq \max\left\{\frac{1}{n\varepsilon}, \sum\limits_{i: p_i\leq \frac{1}{n\varepsilon}}p_i\right\}\Bigg\}
    \end{align}
    If $\delta\leq \varepsilon$ and $d\geq 2$, then the expected KL error of any $(\varepsilon, \delta)$-DP estimator $\mathcal{A}$ satisfies 
    \begin{align}
       \max_{q \in N_{\leq \frac{1}{n\varepsilon}}(p)}\underset{x\sim \text{Poi}(n, q)}{\mathbb{E}}[KL(q, \mathcal{A}(x))]\geq &  \Omega\left(\sum_{i}\min\left\{p_i, \frac{1}{p_i} \cdot \frac{1}{n^2\varepsilon^2} \right\}\right)\label{eqn:dp_per_instance_lower_additive}
    \end{align}
    \label{thm:dp_per_instance_lower_additive}
\end{theorem}
\begin{proof}
    We will apply \cref{thm:thm_conditional_packing} to prove the lower bound. Below we first construct the packing set. Without loss of generality, assume that that $d\mod 2 =0$. (Otherwise, sort the symbols to satisfy $\min\left\{p_1, \frac{1}{p_1}\cdot \frac{1}{n^2\varepsilon^2}\right\}\geq \cdots \geq \min\left\{p_d, \frac{1}{p_d}\cdot \frac{1}{n^2\varepsilon^2}\right\}$ and ignore the last symbol in the below constructions.) For brevity, denote
    \begin{align}
        w_k = p_{2k-1} + p_{2k}\quad\text{and}\quad\Delta_k =  \frac{1}{160} \min\left\{p_{2k}, \frac{1}{n\varepsilon}\right\} \text{ for }k=1, \cdots, \frac{d}{2} \label{eqn:def_delta_k_dp_1_n_eps}
    \end{align}
    Let $\mathcal{P}$ be the following packing set of distributions.
    \begin{align}
        \mathcal{P} = \left\{q\coloneqq \sum_{k=1}^{\frac{d}{2}}w_k \cdot q^k: q_k\in \mathcal{P}_k \coloneqq\left\{q_+^k, q_-^k\right\}\right\} \label{eqn:packing_pair_dp_1_n_eps}
    \end{align}
    where
    
    \begin{align}
        q^k_- (j) = \begin{cases}
            \frac{1}{w_k}\cdot p_{2k-1} & j=2k-1\\
            \frac{1}{w_k}\cdot p_{2k} &  j=2k\\
            0 & j\in[d]\setminus\{2k-1, 2k\}
        \end{cases} \text{and} \quad q^k_+ (j) = \begin{cases}
            \frac{1}{w_k} \cdot \left(p_{2k-1} + \Delta_k\right) & j=2k-1\\
            \frac{1}{w_k} \cdot \left(p_{2k} - \Delta_k\right) & j=2k\\
            0 & j\in[d]\setminus\{2k-1, 2k\}
        \end{cases}\nonumber
    \end{align}
    One can verify that distributions in the packing set $\mathcal{P}$ are normalized and lie in the additive neighborhood $N_{\leq\frac{1}{n\varepsilon}}(p)$ by observing that $\frac{1}{n\varepsilon}\geq \min\left\{p_{i-1}, \frac{1}{n\varepsilon}\right\}\geq \min\left\{p_{i}, \frac{1}{n\varepsilon}\right\}$ for any $i$. By applying \cref{lem:two_symbol_packing_lower_bound} (under setting $a=\frac{p_{2k-1}}{w_k}$ and $\Delta = \frac{\Delta_k}{w_k}$), we further prove that for any $q\in \Delta(d)$, 
    \begin{align}
        \frac{1}{|\mathcal{P}_k|}\sum_{q^k\in \mathcal{P}_k} \mathbf{1}_{KL(q^k, q)\geq f(\mathcal{P}_k)} & \geq \frac{1}{2} \text{ where } f(\mathcal{P}_k) = \frac{\left(\frac{\Delta_k}{w_k}\right)^2}{8 \cdot \frac{p_{2k}}{w_k}} = \frac{1}{8\cdot 160^2\cdot w_k} \cdot \min\left\{p_{2k}, \frac{1}{p_{2k} \cdot n^2\varepsilon^2}\right\}\label{eqn:per_instance_lower_packing_i_dp_1_n_eps}
    \end{align}
    Thus the first condition of \cref{thm:thm_conditional_packing} holds. We now analyze the second condition of \cref{thm:thm_conditional_packing}. For any $k$ and any fixed $q^l\in \mathcal{P}_l, l=1,\cdots, \frac{d}{2}$ and fixed $\bar{q}_k\in\mathcal{P}_k$. Let $x\sim \text{Poi}\left(n, \sum_{l=1}^{\frac{d}{2}} w_l\cdot q^l\right)$ and $y\sim \text{Poi}\left(n\cdot \Delta_k\right)$ and $y'\sim \text{Poi}\left(n\cdot \Delta_k\right)$ be independent Poisson random variables. Then we could construct the following coupling $(x, \bar{x})$ between distributions $\text{Poi}\left(n, \sum_{l=1}^{\frac{d}{2}} w_l\cdot q^l\right)$ and $\text{Poi}\left(n, w_k\cdot q^k + \sum_{l\neq k} w_l\cdot q^l\right)$:
    \begin{align}
        \bar{x}_l = \begin{cases}
            x_l + y & l=2 k - 1\\
            x_l - y' & l = 2k\\
            x_l & l \in [d] \setminus \{2k-1, 2k\}
        \end{cases}
    \end{align}
    By definition, we compute that
    \begin{align}
        \mathbb{E}\left[\lVert x - \bar{x}\rVert_1\right] =  \mathbb{E}\left[y + y'\right] \leq \frac{1}{80\varepsilon} \coloneqq \tau \label{eqn:dp_1_n_eps_lower_last_inequality}
    \end{align}
    where the inequality is by definition $\Delta_k\leq \frac{1}{160n\varepsilon}$.  Thus the second condition of \cref{thm:thm_conditional_packing} holds. By applying \cref{thm:thm_conditional_packing} under our proved conditions \eqref{eqn:per_instance_lower_packing_i_dp_1_n_eps} and \eqref{eqn:dp_1_n_eps_lower_last_inequality}, we prove that 
    \begin{align}
        \max_{q\in\mathcal{P}} \mathbb{E}\left[KL(q, \mathcal{A}(x))\right] 
        \geq & \sum_{k=1}^{\frac{d}{2}} w_k \cdot \left(\frac{1}{10} - 4 \varepsilon \cdot \tau \right)\cdot f(\mathcal{P}_k) =  \frac{1}{20}\cdot \frac{1}{8\cdot 160^2}\cdot \sum_{k=1}^{\frac{d}{2}}\min\left\{p_{2k}, \frac{1}{p_{2k}\cdot n^2\varepsilon^2}\right\}\label{eqn:dp_per_instance_1_over_n_eps_eq1}
    \end{align}
    By repeating the constructions under $\Delta_k = - \frac{1}{160}\min\left\{p_{2k-1}, \frac{1}{n\varepsilon}\right\}$ for $k=1, \cdots, \frac{d}{2}$, we similarly prove that
    \begin{align}
        \max_{q\in\mathcal{P}} \mathbb{E}\left[KL(q, \mathcal{A}(x))\right] 
        \geq & \frac{1}{20}\cdot \frac{1}{8\cdot 160^2}\cdot \sum_{k=1}^{\frac{d}{2}}\min\left\{p_{2k-1}, \frac{1}{p_{2k-1}\cdot n^2\varepsilon^2}\right\}
        \label{eqn:dp_per_instance_1_over_n_eps_eq2}
    \end{align}
    By combining \eqref{eqn:dp_per_instance_1_over_n_eps_eq1} and \eqref{eqn:dp_per_instance_1_over_n_eps_eq2}, we prove the bound \eqref{eqn:dp_per_instance_lower_additive} in the statement.
\end{proof}

\subsection{DP Lower Bound: $N_{\frac{t}{n\varepsilon}}$ neighborhood}

In this section, we prove the per-instance DP estimation lower bound in \cref{tab:dp_instance_lower} for low probability symbols.
\begin{theorem}[Lower Bound - low probability symbols]
    Let $p\in\Delta(d)$ be fixed. For any $t>0$, let $N_{\frac{t}{n\varepsilon}}$  be the additive neighborhood defined as follows.
    \begin{align}
        N_{\leq \frac{t}{n\varepsilon}}\left(p, n\right) = \Bigg\{q: |q_i - p_i|\leq \frac{t}{n\varepsilon} \text{ for any }i\in [d] \text{ and }\sum\limits_{i: p_i\leq \frac{t}{n\varepsilon}}q_i\leq \max\left\{\frac{t}{n\varepsilon}, \sum\limits_{i: p_i\leq \frac{t}{n\varepsilon}}p_i\right\}\Bigg\}
            \label{eqn:add_neighborhood_dp_app}
    \end{align}
    Then if $\delta \leq \varepsilon$, $n\varepsilon\geq 1$, $t\geq 1$ and $d\geq 2$, then for any $L'\subseteq [d]$, the expected KL error of any $(\varepsilon, \delta)$-DP estimator $\mathcal{A}$ satisfies 
    \begin{align}
        \max_{q\in  N_{\leq \frac{t}{n\varepsilon}}\left(p, n\right)}\underset{x\sim \text{Poi}(n, q)}{\mathbb{E}}KL(q, \mathcal{A}(x)) \geq &  \Omega\left(\frac{\ln\left(1 + d_{small}(L')\right)}{n\varepsilon} + p_{small}(L') \cdot \ln\left(1 + \frac{d_{small}(L')}{n\varepsilon \cdot  p_{small}(L') }\right)\right) \label{eqn:thm:dp_per_instance_lower_additive_low_prob}
    \end{align}
    where
    $p_{small}(L') =  \sum\limits_{i\in L', p_i\leq \frac{t}{n\varepsilon}}p_i$ and $d_{small}(L') = \sum\limits_{i\in L', p_i\leq \frac{t}{n\varepsilon}} 1$.
    \label{thm:dp_per_instance_lower_additive_low_prob}
\end{theorem}
\begin{proof}
    \begin{enumerate}
        \item If $d_{small}(L')=0$, then  $p_{small}(L')=0$ and thus \eqref{eqn:thm:dp_per_instance_lower_additive_low_prob} trivially holds.
        \item If $1\leq d_{small}(L') \leq 3$: Without loss of generality, assume that $\{i\in L': p_i\leq \frac{t}{n\varepsilon}\} = \{1, \cdots, d_{small}(L')\}$. We construct a packing set of distributions $\mathcal{P} = \{p^+, p^-\}$ that contains the following two distributions.
        \begin{align}
            p^+(j) = \begin{cases}
                \frac{1}{80n\varepsilon} & j=1\\
                \left(1-\frac{1}{80n\varepsilon}\right)\cdot \left(p_j + \frac{p_1}{d-1}\right) & j =2, \cdots, d
            \end{cases}
        \end{align}
        and
        \begin{align}
            p^-(j) = \begin{cases}
                \frac{1}{160n\varepsilon} & j=1\\
                \left(1-\frac{1}{80n\varepsilon}\right)\cdot \left(p_j + \frac{p_1}{d-1}\right) & j =2, \cdots, d - 1\\
                \left(1-\frac{1}{80n\varepsilon}\right)\cdot \left(p_d + \frac{p_1}{d-1}\right) + \frac{1}{160n\varepsilon} & j =d
            \end{cases}
        \end{align}
        One can validate that the distributions in $\mathcal{P}$ are well-defined and are in the additive neighborhood $N_{\leq \frac{t}{n\varepsilon}}\left(p\right)$ (by observing that $0\leq p_1\leq \frac{t}{n\varepsilon}$ for $d\geq 2$ and thus $\frac{p_1}{d-1}\leq \frac{t}{n\varepsilon}$). By applying \cref{lem:two_symbol_packing_lower_bound} (under setting $a=\frac{1}{80n\varepsilon}$ and $\Delta = \frac{1}{160n\varepsilon}$), we further prove that for any $q'\in \Delta(d)$, 
        \begin{align}
            \frac{1}{|\mathcal{P}|}\sum_{q\in \mathcal{P}} \mathbf{1}_{KL(q, q')\geq f(\mathcal{P})} & \geq \frac{1}{2} \text{ where } f(\mathcal{P}) = \frac{\Delta^2}{a} = \frac{1}{320n\varepsilon}\label{eqn:per_instance_dp_lower_packing_i_small}
        \end{align}
        Thus the first condition of \cref{thm:thm_conditional_packing} holds. Below we analyze the second condition of \cref{thm:thm_conditional_packing}. Let $\bar{q}=p^+$. For any $q\in\mathcal{P}$, let $y\sim \text{Poi}(n \min\{p^+, p^-\})$ and let $y'\sim \text{Poi}(n  \max\{p^+, p^-\} - n\min\{p^+, p^-\})$. Then we could construct the following coupling $(x, \bar{x})$ between distributions $\text{Poi}\left(n, q\right)$ and $\text{Poi}\left(n, \bar{q}\right)$:
        \begin{align}
            x_l = \begin{cases}
                y_1 + y'_1 & l=1\\
                y_l & l =2, \cdots, d
            \end{cases}\text{ and }\bar{x}_l = \begin{cases}
                y_l & l=1, \cdots, d-1\\
                y_l + y'_l & l = d
            \end{cases}
        \end{align}
        By definition, we compute that
        \begin{align}
            \mathbb{E}\left[\lVert x - \bar{x}\rVert_1\right] = & \mathbb{E}\left[y'_1 + y'_d\right] =  \frac{1}{80\varepsilon} \coloneqq \tau \label{eqn:per_instance_dp_lower_tau}
        \end{align}
        Thus the second condition of \cref{thm:thm_conditional_packing} holds. By applying \cref{thm:thm_conditional_packing} under our proved conditions \eqref{eqn:per_instance_dp_lower_packing_i_small} and \eqref{eqn:nondp_lower_stat_last_inequality_small}, we finally prove that 
        \begin{align}
            \max_{q\in\mathcal{P}} \mathbb{E}\left[KL(q, \mathcal{A}(x))\right] 
            \geq & 1 \cdot \left(\frac{1}{10} - 4 \varepsilon\cdot \tau \right) \cdot f(\mathcal{P}) =  \frac{1}{20\cdot 320 n\varepsilon}\\
            \geq & \Omega\left(\frac{\ln\left(1 + d_{small}(L')\right)}{n\varepsilon} + p_{small}(L')\cdot \ln\left(1 + \frac{d_{small}(L')}{p_{small}(L') \cdot n\varepsilon}\right)\right)\label{eqn:dp_per_instance_lower_stat_second_to_last_inequality}
        \end{align}
        where \eqref{eqn:dp_per_instance_lower_stat_second_to_last_inequality} is by the assumption that $d_{small}(L')\leq 3$ and by $\ln\left(1 + x\right)\leq x$ for $x>0$.
        \item If $d_{small}(L')\geq 4$: Without loss of generality, assume that $\{i\in L': p_i\leq \frac{t}{n\varepsilon}\}$ consists of symbols $1, \cdots, d_{small}(L')$. For brevity, denote $\hat{d} = \lfloor \frac{d_{small}(L')}{2}\rfloor\geq 2$ and $\hat{p} = \sum_{i=1}^{\hat{d}}p_i$. Without loss of generality, assume that $p_1 \geq \cdots \geq p_{d_{small}(L')}$, then it follows that 
        \begin{align}
            \hat{d}\geq \frac{d_{small}(L')}{3}\text{ and }\hat{p}\geq \frac{p_{small}(L')}{3}\label{eqn:hat_d_hat_p_dp}
        \end{align}
        Let $\kappa, k\in\mathbb{N}$ be defined as follows.
        \begin{align}
            \kappa = \begin{cases}
                2 & \frac{\hat{d}}{ 160 n\varepsilon \hat{p}} \leq 2\\
                 \hat{d}  &  \frac{\hat{d}}{160 n\varepsilon \hat{p}} > 2 \text{ and }160n\varepsilon \hat{p}< 1\\
                \lfloor\frac{\hat{d}}{160 n\varepsilon \hat{p} } \rfloor & \frac{\hat{d}}{ 160n\varepsilon \hat{p} } > 2\text{ and }160n\varepsilon \hat{p}\geq 1
            \end{cases} \quad k = \begin{cases}
                \lfloor \frac{\hat{d}}{2} \rfloor & \frac{\hat{d}}{160 n\varepsilon \hat{p}} \leq 2\\
                1 &  \frac{\hat{d}}{ 160n\varepsilon \hat{p}} > 2 \text{ and }160n\varepsilon \hat{p}< 1\\
                \lfloor 160n\varepsilon \hat{p} \rfloor & \frac{\hat{d}}{ 160n\varepsilon \hat{p}} > 2 \text{ and }160n\varepsilon \hat{p}\geq 1
            \end{cases}.\label{eqn:def_kappa_k_dp_per_instance}
        \end{align}
        Thus by definition, 
        \begin{align}
            \kappa\geq 2 \quad \text{and}\quad k\geq 1 \quad \text{and} \quad \kappa\cdot k\leq  \hat{d} \leq \frac{d_{small}(L')}{2}\quad \text{and} \quad \frac{k}{160n\varepsilon } \leq \max\left\{\hat{p}, \frac{1}{160n\varepsilon}\right\}\label{eqn:combined_mass_kappa_k_dp_per_instance}
        \end{align}
        For $i=1, \cdots, k$, we construct a packing set of distributions supported on symbols $\mathcal{B}_i = \{k\cdot i - \kappa + 1, \cdots, \delta_{\kappa \cdot i}\}$ as follows.
        \begin{align}
            \text{For }i=1,\cdots, k\quad \mathcal{P}_i = \{\delta_{\kappa\cdot i- \kappa + 1}, \cdots, \delta_{\kappa\cdot i}\}
        \end{align}
        We construct the following set of distributions that lie in the additive neighborhood of $p$.
        \begin{align}
            \mathcal{P} = \left\{q = \left(1-\frac{k}{160 n\varepsilon}\right)\cdot q^c + \sum\limits_{i = 1}^{k}w_i\cdot q^i: w_i = \frac{1}{160 n\varepsilon}\text{ and } q^i \in \mathcal{P}_i\text{ for any }i=1, \cdots, k\right\} \label{eqn:prior_per_instance_additive_small_prob_dp}
        \end{align}
        where 
        \begin{align}
            q_c (j) = \begin{cases}
                0 & j = 1, \cdots, \hat{d}\\
                \frac{1}{1 - \frac{k}{160n\varepsilon}} \cdot \left(p_j + \frac{\max\left\{\hat{p}, \frac{1}{160n\varepsilon}\right\} - \frac{k}{160n\varepsilon}}{d_{small}(L') - \hat{d}} \right)  & j = \hat{d} + 1, \cdots, d_{small}(L')\\
                \frac{p_j}{1 - \frac{k}{160n\varepsilon}} \cdot \left(1 + \frac{\hat{p} - \max\left\{\hat{p}, \frac{1}{160n\varepsilon}\right\}}{1 - p_{small}(L')}\right) & j = d_{small}(L') + 1, \cdots, d
            \end{cases}
        \end{align}
    
        One can verify that the distributions in $\mathcal{P}$ are well-defined (i.e., normalized) and lie in the neighborhood $N_{\leq \frac{t}{n\varepsilon}}(p)$ defined in \eqref{eqn:add_neighborhood_dp_small_symbols}. This is by observing that $p_j\leq \frac{t}{n\varepsilon}$ for $j=1, \cdots, \hat{d}$, and that $0\leq \frac{\max\left\{\hat{p}, \frac{1}{160n\varepsilon}\right\} - \frac{k}{160\cdot n\varepsilon}}{d_{small}(L') - \hat{d}} \leq \max\left\{\frac{\hat{p}}{\hat{d}}, \frac{1}{160n\varepsilon}\right\}$ due to \eqref{eqn:combined_mass_kappa_k_dp_per_instance} and $\hat{d}<d_{small}(L')$, and by $0\geq \frac{\hat{p} - \max\left\{\hat{p}, \frac{1}{160n\varepsilon}\right\}}{1 - p_{small}(L')} \geq - \frac{\frac{1}{160n\varepsilon}}{1 - \frac{3}{160n\varepsilon}}>-1$ under \eqref{eqn:hat_d_hat_p_dp} and $n\varepsilon\geq 1$, and by $0\geq p_j \cdot  \frac{\hat{p} - \max\left\{\hat{p}, \frac{1}{160n\varepsilon}\right\}}{1 - p_{small}(L')}\geq - \frac{1}{160n\varepsilon}$.
    
        Now by applying \cref{lem:dirac_packing_lower_bound} to $\mathcal{P}_i$ for each $i=1, \cdots, k$, we prove that for any $q'\in\Delta(d)$, 
        \begin{align}
            \frac{1}{|\mathcal{P}_i|}\sum_{q\in \mathcal{P}_i} \mathbf{1}_{KL(q, q')\geq f(\mathcal{P}_i)} & \geq \frac{1}{2} \text{ where } f(\mathcal{P}_i) = \ln\left(1 + \frac{\kappa}{4}\right)\label{eqn:per_instance_dp_lower_packing_i_small_ln}
        \end{align}
        Thus the first condition of \cref{thm:thm_conditional_packing} holds. Below we analyze the second condition of \cref{thm:thm_conditional_packing}. For each $i\in [k]$ and fixed $q^j\in\mathcal{P}_j, j=1, \cdots, k$ and fixed $\bar{q}^i\in\mathcal{P}_i$. Let $x\sim\text{Poi}\left(n, \left(1-\frac{k}{160n\varepsilon}\right)\cdot q^c + \sum\limits_{j = 1}^{k}w_j\cdot q^j\right)$ and $y\sim\text{Poi}\left(n, \left(1-\frac{k}{160n\varepsilon}\right)\cdot q^c + w_i\cdot \bar{q}^i + \sum\limits_{j\neq i} w_j\cdot q^j\right)$ be independent Poisson random variables. Then we could construct the following coupling $(x, \bar{x})$ between distribution $\text{Poi}\left(n, \left(1-\frac{k}{160n\varepsilon}\right)\cdot q^c + \sum\limits_{j = 1}^{k}w_j\cdot q^j\right)$ and distribution $\text{Poi}\left(n, \left(1-\frac{k}{160n\varepsilon}\right)\cdot q^c + w_i\cdot \bar{q}^i + \sum\limits_{j\neq i} w_j\cdot q^j\right)$:
        \begin{align}
            \bar{x}_l = \begin{cases}
                y_l & l = \kappa\cdot i - \kappa + 1, \cdots, \kappa \cdot i\\
                x_l & l \in [d]\setminus \{\kappa\cdot i - \kappa + 1, \cdots, \kappa \cdot i\}
            \end{cases}
        \end{align}
        By definition, we compute that
        \begin{align}
            \mathbb{E}\left[\lVert x - \bar{x}\rVert_1\right] = &\sum_{l = \kappa\cdot i - \kappa + 1}^{\kappa \cdot i}\mathbb{E}\left[|x_l - y_l|\right]\leq \sum_{l = \kappa\cdot i - \kappa + 1}^{\kappa \cdot i}\mathbb{E}\left[x_l + y_l\right] =  \frac{1}{80\varepsilon} \coloneqq \tau \label{eqn:per_instance_dp_lower_tau_ln}
        \end{align}
        where the inequality is by triangle inequality for $\ell_1$ distance.  By applying \cref{thm:thm_conditional_packing} under our proved conditions \eqref{eqn:per_instance_dp_lower_packing_i_small_ln} and \eqref{eqn:per_instance_dp_lower_tau_ln}, we finally prove that 
        \begin{align}
            & \max_{q\in\mathcal{P}} \mathbb{E}\left[KL(q, \mathcal{A}(x))\right] 
            \geq \sum_{i=1}^k w_i \cdot \left(\frac{1}{10} - 4 \varepsilon\cdot \tau \right) \cdot f(\mathcal{P}_i) =  \frac{k}{160\cdot n\varepsilon} \cdot \frac{1}{ 20} \cdot \ln\left(1 + \frac{\kappa}{4}\right)\\
            = & \begin{cases}
                \frac{1}{160\cdot n\varepsilon}\lfloor\frac{\hat{d}}{2}\rfloor\cdot \ln\left(1 + \frac{1}{2}\right) & \frac{\hat{d}}{ 160n\varepsilon \hat{p}} \leq 2\\
                \frac{1}{160\cdot n\varepsilon}\cdot \ln\left(1 + \frac{1}{4}\hat{d}\right) &  \frac{\hat{d}}{160 n\varepsilon \hat{p}} > 2 \text{ and }160n\varepsilon \hat{p}< 1\\
                \frac{1}{160\cdot n\varepsilon}\cdot\lfloor 160n\varepsilon \hat{p}\rfloor\cdot \ln\left(1 + \frac{1}{4}\lfloor\frac{\hat{d}}{ 160n\varepsilon \hat{p} } \rfloor\right) & \frac{\hat{d}}{ 160n\varepsilon \hat{p} } > 2\text{ and }160n\varepsilon \hat{p}\geq 1
            \end{cases} \label{eqn:use_def_kappa_k_dp}\\
            \geq & \begin{cases}
                \Omega\left(\frac{\hat{d}}{n\varepsilon}\right) & \frac{\hat{d}}{160 n\varepsilon \hat{p}} \leq 2\\
                \Omega\left(\frac{1}{n\varepsilon}\cdot \ln\left(1 +  \hat{d}\right)\right) &  \frac{\hat{d}}{160 n\varepsilon \hat{p}} > 2 \text{ and }160n\varepsilon \hat{p}< 1\\
                \Omega\left( \hat{p} \cdot \ln\left(1 +  \frac{\hat{d}}{n\varepsilon \hat{p} }\right)\right) & \frac{\hat{d}}{160 n\varepsilon \hat{p} } > 2\text{ and }160n\varepsilon \hat{p}\geq 1
            \end{cases}\label{eqn:use_floor_half_dp}\\
            \geq & \Omega\left(\frac{\ln(1+\hat{d})}{n\varepsilon} + \hat{p}\cdot \ln\left(1 + \frac{\hat{d}}{n\varepsilon \cdot \hat{p}}\right)\right)\label{eqn:per_instance_dp_lower_second_to_last_inequality_small}\\
            \geq & \Omega\left(\frac{\ln(1+d_{small}(L'))}{n\varepsilon} + p_{small}(L')\cdot \ln\left(1 + \frac{d_{small}(L')}{n\varepsilon \cdot p_{small}(L')}\right)\right)
        \end{align}
        where \eqref{eqn:use_def_kappa_k_dp} is by using the definitions of $\kappa$ and $k$ in \eqref{eqn:def_kappa_k_dp_per_instance}, \eqref{eqn:use_floor_half_dp} is by $\lfloor x\rfloor\geq \frac{x}{2}$ for any $x\geq 1$, \eqref{eqn:per_instance_dp_lower_second_to_last_inequality_small} is by $\lambda\ln(1 + x)\geq  \ln\left(1 + \lambda x\right)$ for $\lambda\geq 1$ and $x>0$, and by $\ln(1 + \lambda x)\leq \lambda\cdot \ln\left(1 + x\right)\leq x$ for $0\leq\lambda\leq 1$ and $x>0$, and the last inequality is by \eqref{eqn:hat_d_hat_p_dp}.
    \end{enumerate}
\end{proof}



\section{Deferred Proofs for DP Per-Instance Upper Bounds}

\subsection{Upper Bound: DP Sampling Twice Algorithm}

For convenience, we first prove the following lemma about the probability of small symbols going above a threshold, and the probability of large symbols going below a threshold.
\begin{lemma}[Probability of False Positive]
    Let $m\in\mathbb{N}$, $p>0$, $\tau>0$ and $\varepsilon>0$. Let $x\sim \text{Poi}(mp)$ and $z\sim \text{Lap}\left(\frac{1}{\varepsilon}\right)$. If $p\leq \frac{1}{m\min\{\varepsilon, 1\}}$, then 
    \begin{align}
        \Pr\left[x+z\geq \frac{\tau}{\min\{\varepsilon, 1\}}\right]\leq O\left(e^{-\frac{\tau}{2}}\right)
    \end{align}
    \label{lem:small_symbols_above_threshold}
\end{lemma}
\begin{proof}
    By applying \cref{lem:laplace_plus_poisson_tail_prob} under setting $a = mp_i$, $b = \frac{1}{\varepsilon}$ and $c = \frac{\tau}{\min\{\varepsilon, 1\}}$, we prove that
    \begin{align}
        \Pr\left[x+z\geq \frac{\tau}{\min\{\varepsilon, 1\}}\right] \leq \frac{4}{3} e^{\frac{mp_i\min\{\varepsilon, 1\}}{2} - \frac{\tau}{2}} \leq O\left(e^{-\frac{\tau}{2}}\right)
    \end{align}
    where the last inequality is by $p\leq \frac{1}{m\min\{\varepsilon, 1\}}$.
    
\end{proof}

\begin{lemma}[Probability of False Negative]
    Let $m\in\mathbb{N}$, $p>0$, $\tau>0$, and $\varepsilon>0$. Let $x\sim \text{Poi}(mp)$ and $z\sim \text{Lap}\left(\frac{1}{\varepsilon}\right)$. If $p\geq \frac{t}{m\min\{\varepsilon, 1\}}$ for $t=6\tau\geq 1$, then for any constant $\gamma\geq 0$, 
    \begin{align}
        \Pr\left[x+z\leq \frac{\tau}{\min\{\varepsilon, 1\}}\right]\leq O\left(e^{-\frac{\tau}{2}} \cdot \frac{1}{m^\gamma\min\{1, \varepsilon\}^\gamma p_i^\gamma}\right)
    \end{align}
    \label{lem:large_symbols_below_threshold}
\end{lemma}
\begin{proof}
    By applying \cref{lem:laplace_plus_poisson_tail_prob} under setting $a = mp_i$, $b = \frac{1}{\varepsilon}$ and $c = \frac{\tau}{\min\{\varepsilon, 1\}}$, we prove that
    \begin{align}
        \Pr\left[x+z\leq \frac{\tau}{\min\{\varepsilon, 1\}}\right] \leq & \frac{4}{3} e^{\frac{\tau}{2} - \frac{mp_i\min\{\varepsilon, 1\}}{3}} \leq O\left(e^{-\frac{\tau}{2} - \frac{mp_i\min\{1, \varepsilon\}}{6}}\right) \label{eqn:large_symbols_below_threshold_prob_second_to_last}\\
        \leq & O\left(e^{-\frac{\tau}{2}}\frac{1}{m^\gamma \min\{1,\varepsilon\}^\gamma p_i^\gamma}\right)
    \end{align}
    where \eqref{eqn:large_symbols_below_threshold_prob_second_to_last} is by $ \frac{\tau}{2}-\frac{mp_i\min\{1, \varepsilon\}}{3} < - \frac{\tau}{2} - \frac{mp_i\min\{1, \varepsilon\}}{6}$ for $p_i>\frac{t}{m\min\{1, \varepsilon\}}$ undersetting $t = 6\tau$, and the last inequality is by $e^{-\frac{x}{6}}\leq O\left(\frac{1}{x^\gamma}\right)$ for $x = mp_i\min\{1, \varepsilon\}\geq t\geq 1$ and $\gamma\geq 0$.
\end{proof}
We then prove two lemmas that bounds KL estimation error on (small) symbols below threshold and (large) symbols above threshold.
\begin{lemma}[Error of \cref{alg:dp_per_instance_upper_alt} on Small Symbols - reused samples]
    \cref{alg:dp_per_instance_upper_alt} satisfies 
    \begin{align}
        \mathbb{E}\left[\sum_{i\in L}p_i\cdot \ln\left(\frac{\frac{p_i}{\sum_{i\in L}p_i}}{\frac{\bar{x}_i}{\sum_{i\in L}\bar{x}_i}}\right)\right] \leq & O\Bigg(\mathbb{E}\left[\left(\sum_{i\in L}p_i\right)\ln\left(1 + \frac{\sum\limits_{i\in L} 1}{\min\{1, \varepsilon\}\sum_{i\in L}\alpha np_i}\right) \right] \\
        & + \sum_{i:p_i>\frac{1}{\alpha n\min\{\varepsilon, 1\}}}\left(\frac{1}{\alpha n} + \frac{1}{\alpha^2 n^2\min\{\varepsilon, 1\}^2p_i}\right)\Bigg)
        \label{eqn:private_gt_bound_below_threshold}
    \end{align}
    where we have denoted $L = \left\{i: x_i + Lap\left(0, \frac{1}{\varepsilon}\right)  \leq \frac{\tau}{\min\{\varepsilon, 1\}} \right\}$ as the randomized instance-dependent subset in \cref{alg:dp_per_instance_upper_alt}. \label{lem:private_gt_bound_below_threshold}
\end{lemma}
\begin{proof}
    By definition, we have that
    \begin{align}
        \mathbb{E}\left[\sum_{i\in L}p_i\cdot \ln\left(\frac{\frac{p_i}{\sum_{i\in L}p_i}}{\frac{\bar{x}_i}{\sum_{i\in L}\bar{x}_i}}\right)\right] = & \underbrace{\mathbb{E}\left[\sum_{i\in L}p_i\cdot \ln\left(\frac{\alpha np_i}{\bar{x}_i}\right)\right]}_{\circled{1}} + \underbrace{\mathbb{E}\left[\left(\sum_{i\in L}p_i\right) \ln\left(\frac{\sum_{i\in L}\bar{x}_i}{\sum_{i\in L}\alpha np_i}\right)\right]}_{\circled{2}}
        \label{eqn:private_gt_bound_below_threshold_breakdown}
    \end{align}
    We first analyze \circled{1}. By $\alpha n p_i\leq \min\{\varepsilon, 1\} \leq \bar{x}_i$ under $p_i\leq \frac{1}{\alpha n \min\{\varepsilon, 1\}}$, we compute that
    \begin{align}
        \circled{1}\leq &\mathbb{E}\left[\sum_{i\in L: p_i>\frac{1}{\alpha n\min\{\varepsilon, 1\}}}p_i\cdot \ln\left(\frac{\alpha np_i}{\bar{x}_i}\right)\right] \label{eqn:private_gt_bound_below_threshold_term1}
    \end{align}
    We then analyze $\circled{2}$. By concavity of $\ln(t)$ over $t>0$, we have that 
    \begin{align}
        & \circled{2} \leq \mathbb{E}_L\left[\sum_{i\in L}p_i\ln\left(1 + \frac{\sum_{i\in L}\mathbb{E}[\bar{x}_i - \alpha np_i|i\in L]}{\sum_{i\in L}\alpha np_i}\right)\right] \\
        & \leq \mathbb{E}_L\left[\left(\sum_{i\in L}p_i\right)\ln\left(1 + \frac{\sum\limits_{i\in L: p_i \leq \frac{1}{\alpha n \min\{\varepsilon, 1\}}} \frac{2}{\min\{1, \varepsilon\}} + \sum\limits_{i\in L: p_i> \frac{1}{\alpha n \min\{\varepsilon, 1\}}}\mathbb{E}\left[\bar{x}_i - \alpha n p_i|i\in L\right]}{\sum_{i\in L}\alpha np_i}\right)\right]\label{eqn:term_2_private_gt_bias_cond}\\
        & \leq O\left(\mathbb{E}\left[\left(\sum_{i\in L}p_i\right)\ln\left(1 + \frac{\sum\limits_{i\in L} 1}{\min\{1, \varepsilon\}\sum_{i\in L}\alpha np_i}\right) + \sum\limits_{i\in L: p_i>\frac{1}{\alpha n \min\{\varepsilon, 1\}}}\frac{\bar{x}_i - \alpha n p_i}{\alpha n}\right]\right) \label{eqn:private_gt_bound_below_threshold_term2}
    \end{align}
    where we have denoted $L = \left\{i: x_i + Lap\left(0, \frac{1}{\varepsilon}\right)  \leq \frac{\tau}{\min\{\varepsilon, 1\}} \right\}$ as the randomized instance-dependent subset in \cref{alg:dp_per_instance_upper_alt}. \eqref{eqn:term_2_private_gt_bias_cond} is by applying \cref{cor:cond_bias_truncated_poisson_laplace}; \eqref{eqn:private_gt_bound_below_threshold_term2} is by $\ln(1 + x + y)\leq \ln(1 + x) + y$ for any $x\geq 0$ and any $y$.
    
    By plugging \eqref{eqn:private_gt_bound_below_threshold_term1} and \eqref{eqn:private_gt_bound_below_threshold_term2} into \eqref{eqn:private_gt_bound_below_threshold_breakdown}, we prove that
    \begin{align}
        & \mathbb{E}\left[\sum_{i\in L}p_i\cdot \ln\left(\frac{\frac{p_i}{\sum_{i\in L}p_i}}{\frac{\bar{x}_i}{\sum_{i\in L}\bar{x}_i}}\right)\right] \\
        \leq & O\left(\mathbb{E}\left[\left(\sum_{i\in L}p_i\right)\ln\left(1 + \frac{\sum\limits_{i\in L} 1}{\min\{1, \varepsilon\}\sum_{i\in L}\alpha np_i}\right) + \sum\limits_{i\in L: p_i>\frac{1}{\alpha n \min\{\varepsilon, 1\}}}\left(p_i\ln\left(\frac{\alpha np_i}{\bar{x}_i}\right) + \frac{\bar{x}_i - \alpha n p_i}{\alpha n}\right)\right]\right)\nonumber\\
        \leq & O\left(\mathbb{E}\left[\left(\sum_{i\in L}p_i\right)\ln\left(1 + \frac{\sum\limits_{i\in L} 1}{\min\{1, \varepsilon\}\sum_{i\in L}\alpha np_i}\right) \right] + \sum_{i:p_i>\frac{1}{\alpha n\min\{\varepsilon, 1\}}}\left(\frac{1}{\alpha n} + \frac{1}{\alpha^2n^2\min\{\varepsilon, 1\}^2p_i}\right)\right)\label{eqn:private_gt_below_threshold_symbosl_use_base_lemma}
    \end{align}

    where  \eqref{eqn:private_gt_below_threshold_symbosl_use_base_lemma} is by applying \cref{lem:base_lemma_kl} under setting $m=\alpha n$, $c_1=-\infty$ and $c_2=\frac{\tau}{\min\{\varepsilon, 1\}}$, $b = \frac{1}{\varepsilon}$ and $c=\frac{1}{\min\{\varepsilon, 1\}}$ for $p_i>\frac{1}{\alpha n\min\{1, \varepsilon\}}$.

\end{proof}

\begin{lemma}[Error of \cref{alg:dp_per_instance_upper_alt} on Large Symbols - reused samples]  \cref{alg:dp_per_instance_upper_alt} satisfies 
    \begin{align}
        &\mathbb{E}\Bigg[\sum_{i\notin L}\Bigg(p_i\ln\left(\frac{\alpha np_i}{\max\left\{\tilde{x}_i, \frac{1}{\min\{\varepsilon, 1\}}\right\}}\right) + \frac{\max\left\{\tilde{x}_i, \frac{1}{\min\{\varepsilon, 1\}}\right\} -  \alpha np_i}{\alpha n}\Bigg) \Bigg] \\
        &\leq O\left(d\tau e^{-\frac{\tau}{2}}\cdot \frac{\mathbf{1}_{\min_ip_i\leq \frac{1}{\alpha n\min\{\varepsilon,1\}}}}{\alpha n\min\{\varepsilon, 1\}} + \sum_{i: p_i\geq \frac{1}{\alpha n\min\{\varepsilon, 1\}}}\left(\frac{1}{\alpha n} + \frac{1}{p_i}\cdot\frac{1}{\alpha^2 n^2\min\{\varepsilon, 1\}^2}\right)\right)\label{eqn:private_gt_bound_large_symbols_x}
    \end{align}
    where we have denoted $L = \left\{i: x_i + Lap\left(0, \frac{1}{\varepsilon}\right)  \leq \frac{\tau}{\min\{\varepsilon, 1\}} \right\}$ as the randomized instance-dependent subset in \cref{alg:dp_per_instance_upper_alt}. \label{lem:private_gt_bound_large_symbols_x}
\end{lemma}
\begin{proof}
    By definition and by $\tau\geq 1$, we compute that
    \begin{align}
        & \mathbb{E}\Bigg[\sum_{i\notin L}\Bigg(p_i\ln\left(\frac{\alpha np_i}{\max\left\{\tilde{x}_i, \frac{1}{\min\{\varepsilon, 1\}}\right\}}\right) + \frac{\max\left\{\tilde{x}_i, \frac{1}{\min\{\varepsilon, 1\}}\right\} -  \alpha np_i}{\alpha n}\Bigg) \Bigg]\\
        = &\underbrace{ \mathbb{E}\left[\sum_{i\notin L: p_i\geq \frac{1}{\alpha n\min\{1, \varepsilon\}}} \Bigg(p_i\ln\left(\frac{\alpha np_i}{ \tilde{x}_i}\right) + \frac{ \tilde{x}_i -  \alpha np_i}{\alpha n}\Bigg)\right]}_{\circled{1}} + \underbrace{\mathbb{E}\left[\sum_{i\notin L: p_i<\frac{1}{\alpha n\min\{1, \varepsilon\}}}\Bigg(p_i\ln\left(\frac{\alpha np_i}{\tilde{x}_i }\right) + \frac{\tilde{x}_i -  \alpha np_i}{\alpha n}\Bigg) \right]}_{\circled{2}}
        \label{eqn:private_gt_error_breakdown}
    \end{align}
    We now analyze $\circled{1}$. By applying \cref{lem:base_lemma_kl} under setting $m=\alpha n$, $c_1=\frac{\tau}{\min\{\varepsilon, 1\}}$, $c_2=+\infty$, $b = \frac{1}{\varepsilon}$ and $c=\frac{1}{\min\{1, \varepsilon\}}$, we prove that
    \begin{align}
        \circled{1}\leq & O\Bigg(\sum_{i: p_i\geq \frac{1}{\alpha n\min\{\varepsilon, 1\}}}\left(\frac{1}{\alpha n} + \frac{1}{p_i}\cdot\frac{1}{\alpha^2 n^2\min\{\varepsilon, 1\}^2}\right) \Bigg) \label{eqn:private_gt_term_3}
    \end{align}
    We finally analyze $\circled{2}$. By $\tilde{x}_i \geq \alpha np_i$ for $i\notin L$  and $p_i<\frac{1}{\alpha n\min\{1, \varepsilon\}}$, we prove that
    \begin{align}
        \circled{2} \leq &   \underbrace{ \sum_{i:p_i< \frac{1}{\alpha n\min\{\varepsilon, 1\}}}  \Pr[i\notin L] \cdot  \frac{\mathbb{E}\left[\tilde{x}_i | i\notin L\right] - \alpha n p_i}{\alpha n}  }_{\text{Error on FP}} \leq O\left( e^{-\frac{\tau}{2}}\cdot \frac{ \frac{\tau}{\min\{\varepsilon, 1\}} + \frac{1}{\varepsilon} }{\alpha n} \cdot d \cdot \mathbf{1}_{\min_ip_i\leq \frac{1}{\alpha n\min\{\varepsilon,1\}}}\right) \label{eqn:bound_large_p_i_large_noisy_count}\\
        = & O\left(d\tau e^{-\frac{\tau}{2}}\cdot \frac{\mathbf{1}_{\min_ip_i\leq \frac{1}{\alpha n\min\{\varepsilon,1\}}}}{\alpha n\min\{\varepsilon, 1\}}\right)\label{eqn:private_gt_term_4}
    \end{align}
    where the second inequality in \eqref{eqn:bound_large_p_i_large_noisy_count} is by applying \cref{lem:small_symbols_above_threshold} under setting $m=\alpha n$,  and by applying \cref{lem:sum_laplace_poisson_cond_expectation} under setting $\lambda = \alpha n p_i$, $b = \frac{1}{\varepsilon}$ and $c =   \frac{\tau}{\min\{\varepsilon, 1\}} $. By plugging  \eqref{eqn:private_gt_term_3} and \eqref{eqn:private_gt_term_4} into \eqref{eqn:private_gt_error_breakdown}, we obtain the bound in the statement.
\end{proof}

\begin{lemma}[Error of \cref{alg:dp_per_instance_upper_alt} on Large Symbols - fresh samples]  \cref{alg:dp_per_instance_upper_alt} satisfies 
    \begin{align}
        &\mathbb{E}\left[\sum_{i\notin L} p_i\ln\left(\frac{(1 - \alpha) np_i}{\max\left\{\tilde{x}'_i, \frac{1}{\min\{\varepsilon, 1\}}\right\}}\right) + \frac{\max\left\{\tilde{x}'_i, \frac{1}{\min\{\varepsilon, 1\}}\right\} -  (1 - \alpha) np_i}{(1-\alpha)n} \right] \label{eqn:private_gt_bound_large_symbols_x_prime_lhs}\\
        &\leq O\left(\sum_{i: p_i\geq \frac{1}{\alpha n\min\{\varepsilon, 1\}}}\left(\frac{1}{(1-\alpha)n} + \frac{1}{p_i}\cdot\frac{1}{(1-\alpha)^2n^2\min\{\varepsilon, 1\}^2}\right) + de^{-\frac{\tau}{2}}\cdot\frac{\mathbf{1}_{\min_ip_i\leq \frac{1}{\alpha n\min\{1, \varepsilon\}}}}{(1-\alpha)n\min\{\varepsilon, 1\}}\right)\label{eqn:private_gt_bound_large_symbols_x_prime}
    \end{align}
    where we have denoted $L = \left\{i: x_i + Lap\left(0, \frac{1}{\varepsilon}\right)  \leq \frac{\tau}{\min\{\varepsilon, 1\}} \right\}$ as the randomized instance-dependent subset in \cref{alg:dp_per_instance_upper_alt}. \label{lem:private_gt_bound_large_symbols_x_prime}
\end{lemma}
\begin{proof}
    By the independence between set $L$ and noisy estimate $\tilde{x}'_i$ (as they are computed on independently sampled datasets $x$ and $x'$ respectively), we compute that
    \begin{align}
        \eqref{eqn:private_gt_bound_large_symbols_x_prime_lhs} = &  \sum_{i} \Pr[i\notin L] \cdot  \mathbb{E}\Bigg[ p_i\ln\left(\frac{(1-\alpha)np_i}{\max\left\{\tilde{x}'_i, \frac{1}{\min\{\varepsilon, 1\}}\right\}}\right) +  \frac{\max\left\{\tilde{x}'_i, \frac{1}{\min\{\varepsilon, 1\}}\right\} - (1-\alpha)np_i}{(1-\alpha)n} \Bigg]\\
        \leq &  O\Bigg(\underbrace{\sum_{i: p_i\geq \frac{1}{\alpha n\min\{\varepsilon, 1\}}}\left(\frac{1}{(1-\alpha)n} + \frac{1}{p_i}\cdot\frac{1}{(1-\alpha)^2n^2\min\{\varepsilon, 1\}^2}\right)}_{\text{Error on TP}} +  \underbrace{ \frac{1}{(1-\alpha)n \min\{1, \varepsilon\}}\sum_{i:p_i< \frac{1}{\alpha n\min\{\varepsilon, 1\}}}e^{-\frac{\tau}{2}} }_{\text{Error on FP}}\Bigg)\label{eqn:bound_large_p_i_large_noisy_count_alt}\\
        \leq & O\left(\sum_{i: p_i\geq \frac{1}{\alpha n\min\{\varepsilon, 1\}}}\left(\frac{1}{(1-\alpha)n} + \frac{1}{p_i}\cdot\frac{1}{(1-\alpha)^2n^2\min\{\varepsilon, 1\}^2}\right) + de^{-\frac{\tau}{2}}\cdot\frac{\mathbf{1}_{\min_ip_i\leq \frac{1}{\alpha n\min\{1, \varepsilon\}}}}{(1-\alpha)n\min\{\varepsilon, 1\}}\right) \label{eqn:dp_per_instance_samp_twice_last}
    \end{align}
    where the first term in \eqref{eqn:bound_large_p_i_large_noisy_count_alt} is by $\Pr[i\notin L]\leq 1$ for $p_i\geq \frac{1}{\alpha n\min\{\varepsilon, 1\}}$ (by definition) and by applying \cref{lem:base_lemma_kl} under $m=(1-\alpha)n$, $b=\frac{1}{\varepsilon}$, $c = \frac{1}{\min\{1, \varepsilon\}}$, $c_1=-\infty$ and $c_2=+\infty$; the second term in \eqref{eqn:bound_large_p_i_large_noisy_count_alt} is by applying \cref{lem:small_symbols_above_threshold} under $m = \alpha n$; \eqref{eqn:dp_per_instance_samp_twice_last} is by definition.
\end{proof}

We are now ready to prove the per-instance upper bound for \cref{alg:dp_per_instance_upper_alt}.

\begin{theorem}[DP ``Sampling Twice'' Algorithm]
\label{thm:dp_instance_LM_noisy_threshold_alt}
    The estimator $\mathcal{A}$ given by \cref{alg:dp_per_instance_upper_alt} is $\varepsilon$-DP and satisfies the following error bound for any fixed $p\in \Delta(d)$.
    \begin{align}
        \underset{x\sim \text{Poi}(n, p)}{\mathbb{E}} \Big[KL(p\lVert \mathcal{A}(x))\Big] \leq &O\Bigg(\mathbb{E}\left[\left(\sum_{i\in L}p_i\right)\ln\left(1 + \frac{\sum\limits_{i\in L} 1}{\min\{1, \varepsilon\}\sum_{i\in L} np_i}\right) + \frac{\mathbf{1}_{L\neq \emptyset}}{n\min\{1, \varepsilon\}} \right] \\
        & +  \sum\limits_{i: p_i\geq \frac{1}{n\min\{1, \varepsilon\}}} \frac{1}{p_i}\cdot \frac{1}{n^2\min\{\varepsilon, 1\}^2} \Bigg)
    \end{align}
    where $L = \left\{i: x_i + Lap\left(0, \frac{1}{\varepsilon}\right)  \leq \frac{\tau}{\min\{\varepsilon, 1\}} \right\}$ is as defined in \cref{alg:dp_per_instance_upper_alt}.
\end{theorem}

\begin{proof}
    By the definition of \cref{alg:dp_per_instance_upper_alt}, we compute that
    \begin{align}
        & \mathbb{E}\left[KL(p, \mathcal{A}(x))\right] \\
        = & 
        \mathbb{E}\Bigg[ \sum_{i\in L} p_i\ln\left(\frac{p_i/\sum_{j\in L}p_j}{ \bar{x}_i/\sum_{j\in L}\bar{x}_j}\right) + \left(\sum_{i\in L}p_i\right)\ln\left(\frac{\sum_{i\in L} p_i}{\frac{\tilde{c}}{(1 - \alpha) n}}\right) + \sum_{i\notin L}p_i\ln\left(\frac{p_i}{\frac{\bar{x}_i}{(1-\alpha)n}}\right) + \ln\left(\frac{\tilde{c} + \sum_{i\notin L}\bar{x}_i}{(1-\alpha)n}\right)\Bigg]\nonumber\\
        \leq & \underbrace{ \mathbb{E}\left[\sum_{i\in L} p_i\ln\left(\frac{p_i/\sum_{j\in L}p_j}{ \bar{x}_i/\sum_{j\in L}\bar{x}_j}\right)\right]}_{\circled{1}} + \underbrace{\mathbb{E}\left[\left(\sum_{i\in L}p_i\right)\ln\left(\frac{\sum_{i\in L}(1-\alpha)np_i}{\tilde{c}}\right) +  \frac{\tilde{c} - \sum_{i\in L}(1-\alpha)np_i}{(1-\alpha)n} \right]}_{\circled{2}} \\
        & + \underbrace{\mathbb{E}\left[\sum_{i\notin L}\left(p_i\ln\left(\frac{(1-\alpha)np_i}{\bar{x}_i}\right) +  \frac{\bar{x}_i - (1-\alpha)np_i}{(1-\alpha)n} \right)\right]}_{\circled{3}}\label{eqn:dp_samp_twice_breakdown}
    \end{align}
    where the last inequality is by $\ln(t)\leq t - 1$ for any $t>0$. We first analyze \circled{1}. By applying \cref{lem:private_gt_bound_below_threshold}, we prove that
    \begin{align}
        \circled{1} \leq & O\left(\mathbb{E}\left[\left(\sum_{i\in L}p_i\right)\ln\left(1 + \frac{\sum\limits_{i\in L} 1}{\min\{1, \varepsilon\}\sum_{i\in L}\alpha np_i}\right) \right] + \sum_{i:p_i>\frac{1}{\alpha n\min\{\varepsilon, 1\}}}\left(\frac{1}{\alpha n} + \frac{1}{\alpha^2n^2\varepsilon^2p_i}\right)\right)\\
        \leq & O\left(\mathbb{E}\left[\left(\sum_{i\in L}p_i\right)\ln\left(1 + \frac{\sum\limits_{i\in L} 1}{\min\{1, \varepsilon\}\sum_{i\in L} np_i}\right) \right] + \sum_{i:p_i>\frac{1}{n\min\{\varepsilon, 1\}}}\left(\frac{1}{n} + \frac{1}{n^2\varepsilon^2p_i}\right)\right)\label{eqn:dp_samp_twice_term_1}
    \end{align}
    where \eqref{eqn:dp_samp_twice_term_1} is by $\alpha=0.5$ in \cref{alg:dp_per_instance_upper_alt}.
    We then analyze \circled{2}. By the independence between $L$ and $\tilde{c}$ in \cref{alg:dp_per_instance_upper_alt} (due to independent sampling of datasets $x$ and $x'$), conditioned on fixed $L$, we apply \cref{lem:base_lemma_kl} under setting $m=(1-\alpha)n$, $p = \sum_{i\in L}p_i$, $c_1=-\infty$, $c_2=+\infty$, $b = \frac{1}{\varepsilon}$ and $c = \frac{1}{\min\{1, \varepsilon\}}$ and prove that
    \begin{align}
        \circled{2} \leq  & O\left(\mathbb{E}_L\left[\frac{\mathbf{1}_{L\neq \emptyset}}{(1-\alpha) n\min\{1, \varepsilon\}}\right]\right) \leq \mathbb{E}\left[\frac{\mathbf{1}_{L\neq \emptyset}}{n\min\{1, \varepsilon\}}\right]\label{eqn:dp_samp_twice_term_2}
    \end{align}
    where \eqref{eqn:dp_samp_twice_term_2} is by setting $\alpha = 0.5$ in \cref{alg:dp_per_instance_upper_alt}. We finally analyze $\circled{3}$. By definition of $\bar{x}_i$ for $i\notin L$, we compute that 
    \begin{align}
        \circled{3} = &  \mathbb{E}\Bigg[\sum_{i\notin L}\Bigg(p_i\ln\left(\frac{np_i}{\max\left\{\tilde{x}_i, \frac{1}{\min\{\varepsilon, 1\}}\right\} + \max\left\{\tilde{x}'_i, \frac{1}{\min\{\varepsilon, 1\}}\right\}}\right) \\
        & +  \frac{\max\left\{\tilde{x}_i, \frac{1}{\min\{\varepsilon, 1\}}\right\} + \max\left\{\tilde{x}'_i, \frac{1}{\min\{\varepsilon, 1\}}\right\} -  np_i}{n} \Bigg)\Bigg]\\
        \leq & \alpha \cdot \mathbb{E}\Bigg[\sum_{i\notin L}\Bigg(p_i\ln\left(\frac{\alpha np_i}{\max\left\{\tilde{x}_i, \frac{1}{\min\{\varepsilon, 1\}}\right\}}\right) + \frac{\max\left\{\tilde{x}_i, \frac{1}{\min\{\varepsilon, 1\}}\right\} -  \alpha np_i}{\alpha n}\Bigg) \Bigg]\nonumber\\
        & + (1-\alpha) \cdot  \mathbb{E}\left[\sum_{i\notin L} p_i\ln\left(\frac{(1 - \alpha) np_i}{\max\left\{\tilde{x}'_i, \frac{1}{\min\{\varepsilon, 1\}}\right\}}\right) + \frac{\max\left\{\tilde{x}'_i, \frac{1}{\min\{\varepsilon, 1\}}\right\} -  (1 - \alpha) np_i}{(1-\alpha)n} \right] \label{eqn:use_joint_convexity_kl}\\
        \leq & O\left(\sum_{i: p_i\geq \frac{1}{n\min\{\varepsilon, 1\}}} \frac{1}{p_i}\cdot\frac{1}{n^2\min\{\varepsilon, 1\}^2}  + \frac{\mathbf{1}_{\min_ip_i\leq \frac{2}{n\min\{1, \varepsilon\}}}}{n\min\{\varepsilon, 1\}}\right) \label{eqn:dp_samp_twice_term_3_second_to_last}\\
        \leq & O\left(\sum_{i: p_i\geq \frac{2}{n\min\{\varepsilon, 1\}}} \frac{1}{p_i}\cdot\frac{1}{n^2\min\{\varepsilon, 1\}^2}  + \frac{\Pr[L\neq \emptyset]}{n\min\{\varepsilon, 1\}}\right)\label{eqn:dp_samp_twice_term_3}
    \end{align}
    where \eqref{eqn:use_joint_convexity_kl} is by the joint convexity of the function $x\ln\left(\frac{x}{y}\right)$ with regard to arguments $x, y\geq 0$; \eqref{eqn:dp_samp_twice_term_3_second_to_last} is by applying \cref{lem:private_gt_bound_large_symbols_x} and \cref{lem:private_gt_bound_large_symbols_x_prime} under setting $\alpha = 0.5$ and $\tau = 4\ln d$ as in \cref{alg:dp_per_instance_upper_alt}; and \eqref{eqn:dp_samp_twice_term_3} is by $\Pr[i\in L] \geq \Omega(1)$ for $p_i<\frac{2}{n\min\{1, \varepsilon\}}$ under $\alpha = 0.5$ and $\tau=4\ln d$ in \cref{alg:dp_per_instance_upper_alt} (by applying \cref{lem:laplace_plus_poisson_tail_prob} under setting $a = \alpha n p_i$, $b = \frac{1}{\varepsilon}$ and $c = \frac{\tau}{n\min\{1, \varepsilon\}}$). By plugging our proved bound \eqref{eqn:dp_samp_twice_term_1}, \eqref{eqn:dp_samp_twice_term_2} and \eqref{eqn:dp_samp_twice_term_3} for \circled{1}, \circled{2}, and \circled{3} into \eqref{eqn:dp_samp_twice_breakdown}, we prove the bound in the statement.
\end{proof}

\subsection{Poof for Matching Lower and Upper Bound}

\begin{corollary}
    Let $\mathcal{A}$ be the estimator given by \cref{alg:dp_per_instance_upper_alt}. Let $N_{stat}$, $N_{\leq \frac{t}{n}}$, $N_{\frac{1}{n\varepsilon}}$, $N_{\leq \frac{t}{n\varepsilon}}$ be the additive neighborhoods defined in \eqref{eqn:stat_neighborhood}, \eqref{eqn:add_neighborhood_nondp},  \eqref{eqn:add_neighborhood_dp_small_symbols} respectively. Then for any $n$ and any $p\in \Delta(d)$, 
    \begin{align}
        \underset{x\sim \text{Poi}(n, p)}{\mathbb{E}}\Big[KL(p\lVert \mathcal{A}(x))\Big] \leq & O\Bigg(\underbrace{\text{lower}\left(p, n, N_{stat}\right) +  \text{lower}\left(p, n, N_{\leq \frac{t}{n}}\right)}_{\text{Non-DP Per-instance Lower Bound}}\nonumber\\
        & + \underbrace{\text{lower}_{\varepsilon, \delta}\left(p, n, N_{\frac{1}{n\varepsilon}}\right) + \text{lower}_{\varepsilon, \delta}\left(p, n, N_{\leq \frac{t}{n\varepsilon}}\right)}_{\text{DP Per-instance Lower Bound}}\Bigg) \label{eqn:instance_optimal_dp_app}
    \end{align}
    under choosing neighborhood size $t = 6\tau$ for $\tau = 4\ln d$.
    \label{cor:dp_per_instance_upper_additive}
\end{corollary}
\begin{proof}
    We will use the upper bound given by \cref{thm:dp_instance_LM_noisy_threshold_alt}. Observe that for any $t>0$, 
    $\frac{\sum\limits_{i\in L} 1}{\min\{1, \varepsilon\}\sum_{i\in L} np_i}\leq \frac{d_{small}(L)}{\min\{1, \varepsilon\} np_{small}(L)}$, where $d_{small}(L) = \sum\limits_{i\in L: p_i\leq \frac{t}{\alpha n\min\{\varepsilon, 1\}}} 1$ and $p_{small}(L) = \sum\limits_{i\in L: p_i\leq \frac{t}{\alpha n\min\{\varepsilon, 1\}}} p_i$. Thus the first term in \cref{thm:dp_instance_LM_noisy_threshold_alt} satisfies
    \begin{align}
        & \mathbb{E}\left[\left(\sum_{i\in L}p_i\right)\ln\left(1 + \frac{\sum\limits_{i\in L} 1}{\min\{1, \varepsilon\}\sum_{i\in L} np_i}\right) + \frac{\mathbf{1}_{L\neq \emptyset}}{n\min\{1, \varepsilon\}}\right] \\
        \leq & \mathbb{E}\left[p_{small}(L)\cdot \ln\left(1 + \frac{d_{small}(L)}{\min\{1, \varepsilon\} np_{small}(L)}\right) \right] + \sum_{i:p_i> \frac{t}{\alpha n\min\{\varepsilon, 1\}}} p_i \cdot \ln\left(1 + \frac{d}{t}\right)\cdot \Pr[i\in L]\label{eqn:partion_below_threshold}\\
        & + \mathbb{E}\left[\frac{\mathbf{1}_{\min_{i\in L}p_i\leq \frac{t}{\alpha n \min\{\varepsilon, 1\}}}}{n\min\{\varepsilon, 1\}}\right] + \sum_{i: p_i\geq \frac{t}{\alpha n \min\{\varepsilon, 1\}}}\frac{\Pr[i\in L]}{n\min\{\varepsilon, 1\}}\label{eqn:partion_below_threshold_second_part}\\
        \leq &  \mathbb{E}\left[p_{small}(L) \ln\left(\frac{ d_{small}(L)}{p_{small}(L)\cdot n\min\{1, \varepsilon\}}\right) + \frac{\ln\left(1 + d_{small}(L)\right)}{n\min\{\varepsilon, 1\}}\right] + O\left(\sum_{i: p_i>\frac{t}{n\min\{\varepsilon, 1\}}} \frac{\ln d e^{-\frac{\tau}{2}} + e^{-\frac{\tau}{2}}}{n^2\min\{1, \varepsilon\}^2p_i}\right) \label{eqn:cor_dp_samp_twice_term_1}\\
        \leq & \max_{L'\subset [d]} \left( p_{small}(L') \ln\left(\frac{ d_{small}(L')}{p_{small}(L')\cdot n\min\{1, \varepsilon\}}\right) + \frac{\ln\left(1 + d_{small}(L')\right)}{n\min\{\varepsilon, 1\}}\right) + O\left(\sum_{i: p_i>\frac{t}{n\min\{\varepsilon, 1\}}} \frac{1}{n^2\min\{1, \varepsilon\}^2p_i}\right)\label{eqn:cor_dp_samp_twice_term_2}\\
        = & O\left(\text{\cref{thm:non_dp_lower_per_instance_additive_small_symbols}} + \text{\cref{thm:dp_per_instance_lower_additive_low_prob}} + \text{\cref{thm:non_dp_lower_per_instance_additive}} + \text{\cref{thm:dp_per_instance_lower_additive}}\right)\label{eqn:dp_per_instance_choose_threshold_1}
    \end{align}
    where \eqref{eqn:partion_below_threshold} is by $\frac{\sum\limits_{i\in L} 1}{\min\{1, \varepsilon\}\sum_{i\in L} np_i}\leq \frac{d_{small}(L)}{\min\{1, \varepsilon\} np_{small}(L)}$, for any $t>0$ and $d_{small}(L) = \sum\limits_{i\in L: p_i\leq \frac{t}{n\min\{\varepsilon, 1\}}} 1$ and $p_{small}(L) = \sum\limits_{i\in L: p_i\leq \frac{t}{n\min\{\varepsilon, 1\}}} p_i$; \eqref{eqn:partion_below_threshold_second_part} is $\Pr[L\neq \emptyset] \leq \Pr\left[\min\limits_{i\in L}p_i\leq \frac{t}{\alpha n \min\{\varepsilon, 1\}}\right] + \sum\limits_{i: p_i\geq \frac{t}{\alpha n \min\{\varepsilon, 1\}}}\Pr[i\in L]$ (ensured by union bound); \eqref{eqn:cor_dp_samp_twice_term_1} is by $\mathbf{1}_{\min_{i\in L}p_i\leq \frac{t}{\alpha n \min\{\varepsilon, 1\}}\}}\leq \ln\left(1 + \sum\limits_{i\in L: p_i\leq \frac{t}{\alpha n \min\{\varepsilon, 1\}}}1\right) = \ln\left(1 + d_{small}(L)\right)$, and by applying \cref{lem:large_symbols_below_threshold} under setting $m = \alpha n$ and choosing $t = 6\tau$; and \eqref{eqn:cor_dp_samp_twice_term_2} is by setting $\tau = 4\ln d$ as defined in \cref{alg:dp_per_instance_upper_alt}, and \eqref{eqn:dp_per_instance_choose_threshold_1} is by definition. Additionally, observe that by definition, the second term in \cref{thm:dp_instance_LM_noisy_threshold_alt} is $\sum\limits_{i: p_i>\frac{1}{n\min\{1, \varepsilon\}}} \frac{1}{p_i}\cdot \frac{1}{n^2\min\{\varepsilon, 1\}^2} \leq  O\left(\text{\cref{thm:non_dp_lower_per_instance_additive}} + \text{\cref{thm:dp_per_instance_lower_additive}}\right)$. Combining this with \eqref{eqn:dp_per_instance_choose_threshold_1} suffice to prove the bound in the statement. 
\end{proof}

\subsection{Discussions on the Neighborhood Size}

\begin{lemma}[Generalized Packing Argument]
    Let $d\geq 16\in\mathbb{N}$ and let $\mathcal{O}$ be an output space. Let $\text{err}: \mathcal{O}\times \mathcal{O}\rightarrow \mathbb{R}$ be an error function. Given $p^1, \cdots, p^d\in \mathcal{O}$. For $i\in[d]$, denote $\mathcal{S}(p^i)$ as the distribution of histogram sampled from distribution $p^i$. Assume that 
    \begin{enumerate}
        \item for any $i, j \in [d]$, 
        \begin{align}
            \mathbb{E}_{(x, x')}\left[\lVert x - x'\rVert_{1}\right] \leq \frac{\ln d}{16\varepsilon} \label{eqn:generalized_packing_cond_1}
        \end{align}
        for a coupling $(x, x')$ between the distributions $\mathcal{S}(p^i)$ and $\mathcal{S}(p^j)$;
        \item for any $q\in\mathcal{O}$ and any $S\subseteq [d]$ such that $|S|\geq d^{1/4}$, it holds that
        \begin{align}
            \frac{1}{S}\sum_{i\in S} \text{err}\left(p^i, q\right) \geq \frac{\ln d}{4} \label{eqn:generalized_packing_cond_2}
        \end{align}
    \end{enumerate}
    Then for any $(\varepsilon, \delta)$-DP algorithm $\mathcal{A}$ with $\delta<\frac{\varepsilon}{d^{1/4}\ln d}$, we have 
    \begin{align}
        \max_{i\in d} \mathbb{E}_{x\sim\mathcal{S}(p^i)}\left[\text{err}\left(p^i, \mathcal{A}(x)\right)\right] \geq \frac{\ln d}{16} \label{eqn:generalized_packing}
    \end{align}
    \label{lem:generalized_packing}
\end{lemma}
\begin{proof}
    We will prove the lemma by contradiction. 
    Consider a bipartite graph $(V, E)$ on $V = [d]\times \mathcal{O}$, where $(i, v)\in E$ if and only if $\text{err}(p^i, q) < \frac{\ln d}{4}$. Then
    \begin{align}
        \text{For any }v\in\mathcal{O}, \quad \text{degree}(v)<d^{1/4}\label{eqn:degree_condition}
    \end{align}
    Otherwise the set of neighbors $\text{Nbr}(v) = \{i\in [d]: (i, v)\in E\}$ violates \eqref{eqn:generalized_packing_cond_2}. 
    
    Suppose that \eqref{eqn:generalized_packing} does not hold, then 
    \begin{align}
        \text{For any }i\in[d], \quad \mathbb{E}_{x\sim\mathcal{S}(p^i)}\left[\text{err}\left(p^i, \mathcal{A}(x)\right)\right] < \frac{\ln d}{16} \label{eqn:generalized_packing_contra}
    \end{align}
    Denote the neighbor set $\text{Nbr}(i) = \{v\in \mathcal{O}: (i, v)\in E\}$. Then by Markov's inequality
    \begin{align}
        \Pr_{x\sim\mathcal{S}(p^i)}\left[\mathcal{A}(x)\in \text{Nbr}(i)\right] = 1-\Pr_{x\sim\mathcal{S}(p^i)}\left[\text{err}\left(p^i, \mathcal{A}(x)\right) \geq \frac{\ln d}{4}\right] \geq 1 -  \frac{\mathbb{E}_{x\sim\mathcal{S}(p^i)}\left[\text{err}\left(p^i, \mathcal{A}(x)\right)\right]}{\frac{\ln d}{4}} \geq \frac{3}{4} \label{eqn:generalized_packing_high_est}
    \end{align}
    where the last equality is by \eqref{eqn:generalized_packing_contra}.

    On the other hand, for any $j\in [d]$, let $(x, x')$ be the coupling between $\mathcal{S}(p^i)$ and $\mathcal{S}(p^j)$ in \eqref{eqn:generalized_packing_cond_1}, we prove that
    \begin{align}
        \Pr_{x\sim\mathcal{S}(p^i)}\left[\mathcal{A}(x)\in \text{Nbr}(i)\right] \leq & \Pr_{(x, x')}\left[ \mathcal{A}(x)\in \text{Nbr}(i)\text{ and }\lVert x - x' \rVert_1\leq \frac{\ln d}{4\varepsilon}\right] + \Pr_{(x, x')}\left[ \lVert x - x' \rVert_1> \frac{\ln d}{4\varepsilon}\right]\label{eqn:generalized_packing_use_union_bound}\\
        \leq & \; e^{\varepsilon\cdot \frac{\ln d}{4\varepsilon}} \cdot \Pr_{(x, x')}\left[ \mathcal{A}(x')\in \text{Nbr}(i)\right] +  \frac{\ln d}{4\varepsilon} \cdot e^{\varepsilon\cdot \frac{\ln d}{4\varepsilon}} \cdot \delta + \frac{\mathbb{E}_{(x,x')}\left[\lVert x - x'\rVert_1\right]}{\frac{\ln d}{4\varepsilon}}\label{eqn:generalized_packing_use_group_privacy}\\
        = & \; d^{1/4}\cdot \Pr_{x\sim\mathcal{S}(p^j)}\left[ \mathcal{A}(x)\in \text{Nbr}(i)\right] + \frac{1}{2}\label{eqn:generalized_packing_use_constants}
    \end{align}
    where \eqref{eqn:generalized_packing_use_union_bound} is by applying union bound; the first term in \eqref{eqn:generalized_packing_use_group_privacy} is by recursive usage of definition of $(\varepsilon, \delta)$-DP to datasets $(x, x')$ with hamming distance bounded by $\frac{\ln d}{4\varepsilon}$; the second term in \eqref{eqn:generalized_packing_use_group_privacy} is by applying Markov's inequality; and \eqref{eqn:generalized_packing_use_constants} is by applying condition \eqref{eqn:generalized_packing_cond_1} and $\delta\leq \frac{\varepsilon}{d^{1/4}\ln d}$. 
    
    By combining \eqref{eqn:generalized_packing_high_est} and \eqref{eqn:generalized_packing_use_constants}, it follows that
    \begin{align}
        \Pr_{x\sim\mathcal{S}(p^j)}\left[ \mathcal{A}(x)\in \text{Nbr}(i)\right]\geq \frac{1}{4d^{1/4}} \label{eqn:generalized_packing_j}
    \end{align}
    By summing \eqref{eqn:generalized_packing_j} over all $i$, we prove that
    \begin{align}
        \sum_{i=1}^d\Pr_{x\sim\mathcal{S}(p^j)}\left[\mathcal{A}(x)\in \text{Nbr}(i)\right]\geq \frac{d^{3/4}}{4} 
    \end{align}
    Thus there exists $v\in \mathcal{O}$, such that
    \begin{align}
        \sum_{i=1}^d \mathbf{1}_{v\in \text{Nbr}(i)}\geq \frac{d^{3/4}}{4} \geq d^{1/4}
    \end{align}
    where the last inequality is by $d\geq 16$. This contradicts with \eqref{eqn:degree_condition}. 
\end{proof}

\begin{theorem}
    \label{thm:necessity_neighborhood_size}
    Let $n, d \geq 4\in\mathbb{N}$, $\varepsilon = \frac{\ln(d)}{16n}$, and $0\leq \delta\leq \frac{\varepsilon}{d^{1/4}\ln d}$. For $\gamma\leq \frac{1}{32}$, let $N_{\gamma}(p)$ be the local neighborhood as defined in \eqref{eqn:add_neighborhood_dp_small_symbols} for $t = \gamma\ln d$.
    \begin{align}
        N_{\gamma} (p) = \Bigg\{q: |q_i - p_i|\leq \frac{\gamma\ln d}{n\varepsilon} \text{ for any }i\in [d] \text{ and }\sum\limits_{i: p_i\leq \frac{\gamma\ln d}{n\varepsilon}}q_i\leq \max\left\{\frac{\gamma\ln d}{n\varepsilon}, \sum\limits_{i: p_i\leq \frac{\gamma \ln d}{n\varepsilon}}p_i\right\}\Bigg\}
    \end{align}
    Then there exists a set $\mathcal{P}$ of distribution instances on $\Delta(d)$, and a per-neighborhood estimator $\mathcal{A}_{N_\gamma(p)}$ under neighborhood size $\gamma$, such that
    \begin{align}
        \max_{p\in\mathcal{P}} \max_{q\in N_\gamma(p)}\underset{x\sim\text{Poi}(nq)}{\mathbb{E}}\left[KL\left(q, \mathcal{A}_{N_\gamma(q)}(x)\right)\right] \leq 48\gamma\ln d\label{eqn:per_neighborhood_est_upper}
    \end{align}
    while for any $(\varepsilon, \delta)$-DP estimator $\mathcal{A}$, we have 
    \begin{align}
        \max_{p\in\mathcal{P}}\underset{x\sim\text{Poi}(np)}{\mathbb{E}}\left[KL\left(p, \mathcal{A}(x)\right)\right] \geq \frac{\ln d}{16}\label{eqn:dp_selection_lower}
    \end{align}
    Thus if $\gamma\leq o(1)$, then no $(\varepsilon, \delta)$-DP estimator $\mathcal{A}$ could satisfy \eqref{eqn:instance_optimality_objective} (otherwise it contradicts \eqref{eqn:per_neighborhood_est_upper} and \eqref{eqn:dp_selection_lower}).
\end{theorem}

\begin{proof}
    Consider the following construction of $\mathcal{P}$.
    \begin{align}
        \mathcal{P} = \Big\{p^i\coloneqq \delta_i\text{ for }i\in[d] \Big\}
    \end{align}
    Then 
    \begin{align}
        N_\gamma(p^i) = \left\{q: q_i \geq 1 - \gamma\cdot \frac{\ln d}{n\varepsilon}, q_j\leq \gamma\cdot \frac{\ln d}{n\varepsilon}\text{ for any }j\neq i\right\}
    \end{align}
    Thus the following construction of per-neighborhood estimator satisfies 
    \eqref{eqn:per_neighborhood_est_upper}.
    \begin{align}
        \mathcal{A}_{N_\gamma(p^i)}(x)_j = \begin{cases}
            1 - \gamma \cdot \frac{\ln d}{n\varepsilon} & j = i\\
            \gamma\cdot \frac{\ln d}{n\varepsilon\cdot (d - 1)} & j\neq i
        \end{cases} 
    \end{align}
    This is because
    \begin{align}
        \max_{q\in N_\gamma(p^i)}\underset{x\sim\text{Poi}(nq)}{\mathbb{E}}\left[KL(q, \mathcal{A}_{N_\gamma(p^i)})\right] \leq &\ln\left(\frac{1}{1 - \gamma\cdot \frac{\ln d}{n\varepsilon}}\right) + \gamma\cdot \frac{\ln d}{n\varepsilon} \cdot \ln(d-1)\\
        \leq & 2\gamma\cdot \frac{\ln d}{n\varepsilon} + \gamma\cdot \frac{(\ln d)^2}{n\varepsilon}
        \leq 48\gamma\cdot \ln d\label{eqn:per_neighborhood_est_upper_second_to_last}
    \end{align}
    where \eqref{eqn:per_neighborhood_est_upper_second_to_last} is by $\gamma \cdot \frac{\ln d}{n\varepsilon}\leq \frac{1}{2}$ under $\gamma\leq \frac{1}{32}$ and by $\varepsilon = \frac{\ln d}{2n}$.
    
    Below we focus on proving \eqref{eqn:dp_selection_lower} by applying \cref{lem:generalized_packing}. We only need to validate that the two conditions of \cref{lem:generalized_packing} hold.
    \begin{enumerate}
        \item The first condition \eqref{eqn:generalized_packing_cond_1} holds by $\varepsilon = \frac{\ln d}{16n}$.
        \item The second condition of \eqref{eqn:generalized_packing_cond_2} holds by convexity of the function $\ln(\frac{1}{t})$ on $t>0$, which ensures that for any $S\subseteq [d]$ with $|S|\geq d^{1/4}$ and any $q\in\Delta(d)$, we have 
    \begin{align}
        \frac{1}{|S|}\sum_{i\in S}KL(p^i, q) = \frac{1}{|S|}\sum_{i\in S}\ln\left(\frac{1}{q_i}\right) \geq \ln\left(\frac{1}{\frac{1}{|S|}\sum_{i\in S}q_i}\right) \geq \ln(|S|) \geq \frac{1}{4}\ln d
    \end{align}
    where the second-to-last inequality is by $\sum_{i\in S}q_i\leq 1$, and the last inequality is by $|S|\geq d^{1/4}$.
    \end{enumerate}
    
\end{proof}


\end{document}